\documentclass[aps,prl,twocolumn,superscriptaddress]{revtex4-2}

\usepackage{graphicx}

\usepackage{amssymb}
\usepackage{amsmath}
\usepackage{verbatim}

\usepackage{algorithmicx}
\usepackage{algpseudocode}

\begin{document}

\title{Beyond Episodic AI: Cognitive Field Networks for Biologically Inspired Persistent Cognition}

\author{Byung Gyu Chae}

\affiliation{Electronics and Telecommunications Research Institute, 218 Gajeong-ro, Yuseong-gu, Daejeon 34129, Republic of Korea
\\ bgchae@etri.re.kr}


\begin{abstract}

Cognitive Field Theory describes cognition as
memory-dressed collective dynamics, in which slow
relaxation modes support a persistent cognitive field.
By connecting non-Markovian memory dynamics to discrete
recurrent state evolution through an extended state,
we develop a Cognitive Field Network (CFN), a recurrent
Transformer architecture with hidden-field re-entry,
\(
\Phi_{n+1}
=
F_\theta(X_{n+1},\Phi_n).
\)
Rather than prescribing an explicit memory operation,
the CFN allows the full token-resolved hidden field
to re-enter the next computational cycle.
We find that learning organizes this pathway into
a functional channel for persistent, content-dependent
internal states, with temporal persistence systematically
controlled by the recurrent training horizon.
The learned dynamics generalize beyond this horizon:
semantic continuation sustains the recurrent state
without explicit answer replay, whereas unsupported
states exhibit finite passive relaxation.
Periodic re-exposure to relevant input renews the
surviving field, producing an approximately stationary
nonzero regime over extended recurrent trajectories.
Unrelated-input and recurrence-off controls fail to
reproduce this behavior, while near-paraphrased input
produces weaker renewal, revealing the dependence of
persistence on both field re-entry and input relevance.
These findings connect memory-dressed field formation,
input-driven reorganization, and recurrent state
continuity within a unified dynamical framework.
The CFN thus provides an experimentally controllable
platform for studying persistent, history-dependent
cognition without a separately prescribed memory system.

\end{abstract}
\maketitle

\section{I. Introduction}

A central problem in the science of cognition is to
explain how transient neural activity develops into
persistent internal dynamics without presupposing
a separate mechanism that stores, observes, or
interprets that activity. Persistent and recurrent
neural activity has long been implicated in working
memory and cognitive control, where maintained
internal representations influence subsequent
neural processing \cite{1,2,3,4,5,6}.
An intelligent system operating continuously in
time must carry forward the consequences of its
previous computation while remaining responsive
to new information. The fundamental question is
therefore how internally generated states can
persist, influence subsequent computation, and
remain reorganizable by ongoing experience.

Cognitive Field Theory (CFT) provides a physical
framework for this problem by describing cognition
as a collective nonequilibrium dynamical process
\cite{7,8}. Within CFT, learned cognitive
organization generates a spectrum of collective
relaxation and circulation modes. The infrared sector
of this spectrum produces non-Markovian memory
feedback, renormalizing the effective cognitive
forgetting gap and supporting
a memory-bearing macroscopic cognitive field.
In this description, inference is
not merely an instantaneous input--output
mapping but involves the continuing evolution
of a history-dependent collective state.

An important computational question is how
such internally generated dynamics can become
causally available to subsequent inference.
The original Transformer architecture replaced
explicit sequential hidden-state recurrence
with self-attention, enabling substantially
more parallel sequence computation
\cite{9,10,11,12,13}. Subsequent architectures
have introduced recurrent computation through
attention, depth recurrence, segment-level
hidden-state reuse, feedback memory, dedicated
recurrent states, and block-level recurrent
processing \cite{14,15,16,17,18,19,20}.
These approaches demonstrate the computational
value of recurrence for long-context modeling,
memory, state tracking, and increased effective
computational depth.

The present study addresses a related but
distinct dynamical question: whether an
internally generated hidden state can acquire
persistent and content-dependent memory
functions through learning, without a
prescribed memory-storage operation.
From the perspective of CFT, non-Markovian
memory feedback can be represented through
an extended dynamical state containing
collective relaxation modes. This provides
a connection between continuous memory
dynamics and discrete recurrent state
evolution, motivating a computational
architecture in which the internal state
generated during one cycle participates
in subsequent inference.

To investigate this possibility, we introduce
the Cognitive Field Network (CFN), a recurrent
Transformer architecture in which the full
token-resolved hidden field generated during
one computational cycle is allowed to re-enter
subsequent cycles. Its recurrent dynamics
can be expressed abstractly as
\begin{equation}
\Phi_{t+1}
=
F_\theta(X_{t+1},\Phi_t),
\label{eq:intro_cfn_map}
\end{equation}
where $\Phi_t$ denotes the recurrent hidden
field and $X_{t+1}$ denotes newly presented
information. The network therefore processes
each new input in the presence of an internally
generated state carrying the consequences
of its preceding computation.

The central architectural principle is
field re-entry. Rather than prescribing
a separate symbolic memory, storage algorithm,
or explicit memory-update rule, the CFN
provides a recurrent pathway through which
the network's own hidden activity can
influence future computation. Learning
determines whether and how this pathway
acquires a functional memory role.
The recurrent hidden field is a learned
neural representation.

Field re-entry introduces a distinction
between unidirectional computation and
dynamically recursive inference. In
conventional feed-forward processing \cite{21,22,23},
internally generated hidden representations
are primarily used to determine the current
output. In the CFN, the hidden field
generated during one computational cycle
remains available to subsequent computation,
where it interacts with newly arriving
information. The resulting dynamics is
explicitly history-dependent, since the
present computation depends on both
external input and an internally generated
state inherited from preceding cycles.

In this restricted dynamical sense, field
re-entry also provides a minimal form
of self-reference. The term refers to
the causal participation of an internally
generated state in subsequent neural
computation, rather than to consciousness,
self-awareness, or metacognition. It
expresses the computational principle
that the consequences of previous internal
activity can become part of the conditions
governing future activity.

This principle has a biological motivation.
Persistent neural activity and recurrent
cortical interactions have long been
associated with working memory, cognitive
control, and the maintenance of internally
available information across time
\cite{2,3,4,5}. From a dynamical
perspective, working-memory-like functions
may arise through the adaptive organization
of recurrent neural interactions without
requiring a separately prescribed memory
operation. The CFN does not attempt to
reproduce a specific biological circuit.
Its biological inspiration lies in the
more general principle of cognitive
continuity through recurrent state
persistence and reorganization.

We investigate the functional consequences
of field re-entry through controlled
interventions and long-horizon memory
experiments. We first compare correct-field,
incorrect-field, and recurrence-off
conditions to examine whether content-specific
retrieval depends causally on the recurrent
hidden state. We then vary the recurrent
distance used during training and evaluate
memory retention beyond the trained horizon
to characterize the learning dependence
of temporal persistence.

To examine the evolution of the recurrent
state under subsequent input, we compare
propagation through unrelated random inputs,
semantic continuation, and periodic
re-exposure to relevant source information.
Additional controls examine the dependence
of field renewal on the content and
representation of incoming information.
These experiments are designed to distinguish
causal memory dependence, learned persistence,
passive relaxation, and input-driven
reorganization within a common recurrent
architecture.

The remainder of this paper is organized
as follows. Section~II establishes the
dynamical connection between memory-dressed
cognitive fields and recurrent state
evolution, describes the Transformer
implementation, and develops the
input-driven collective description.
Section~III examines learning-dependent
dynamics, controlled field interventions,
and temporal persistence. Section~IV
investigates long-horizon retention,
semantic continuation, passive relaxation,
and content-dependent field renewal.
Section~V discusses the physical,
computational, and biological implications,
and Sec.~VI presents the conclusions.

\begin{figure*}[t]
\centering
\includegraphics[width=1.0\textwidth, trim=0cm 0.5cm 0cm 0cm]{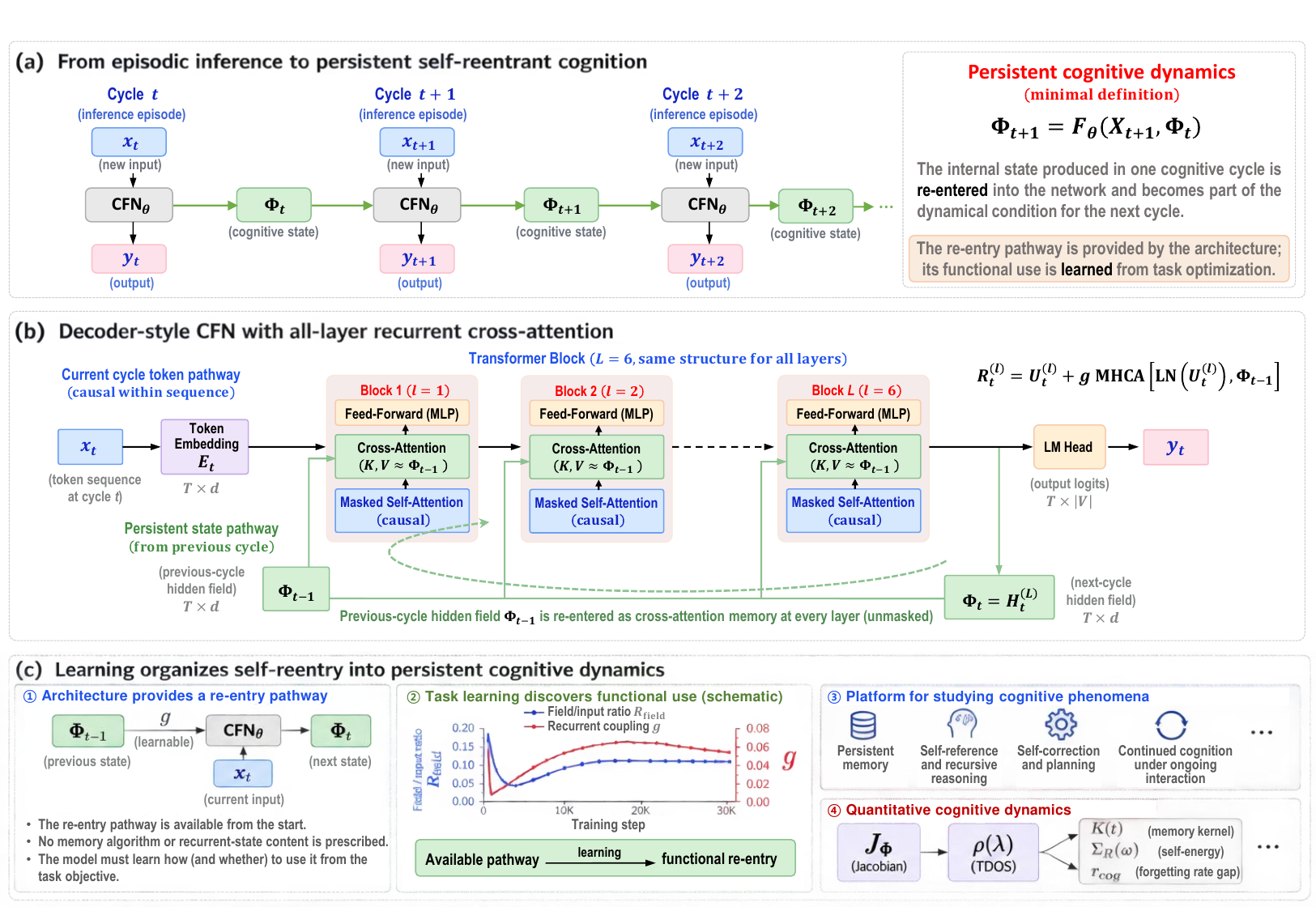}
\caption{
Cognitive Field Network (CFN): self-reentrant architecture and
learning for persistent cognition.
(a) Transition from episodic inference to self-reentrant cognitive
dynamics. The hidden field $\Phi_t$ generated during one computational
cycle is retained and re-entered into the subsequent cycle, yielding
the state-dependent evolution
$\Phi_{t+1}=F_\theta(X_{t+1},\Phi_t)$.
(b) Decoder-style CFN implementation used in this work.
The current token sequence follows the ordinary causal Transformer
pathway, while the complete token-resolved hidden field
$\Phi_{t-1}$ from the preceding cycle is supplied as unmasked
cross-attention memory to every Transformer layer, with learnable
recurrent coupling $g$.
After the final layer, $\Phi_t=H_t^{(L)}$ provides both the internal
field retained for the next cycle and the hidden representation
from which the language-model output is read out.
(c) Learning and experimental analysis of self-reentrant cognitive
dynamics. The architecture makes field re-entry available without
prescribing its functional role. Task-driven learning can organize
this pathway into persistent, content-dependent recurrent
computation. The recurrent hidden field can be directly accessed
and experimentally manipulated, enabling causal tests of memory
persistence and the functional contribution of self-reentry across
successive computational cycles. The framework also provides a basis
for quantitative dynamical analysis through Jacobian spectra,
relaxation-mode distributions, memory kernels, and collective
response observables.
}
\label{fig:cfn_architecture}
\end{figure*}

\section{II. Cognitive Field Network: Self-Reentrant Architecture for Persistent Cognition}
\label{sec:cfn}

Cognitive Field Theory describes cognition as a
memory-dressed collective dynamics in which the present
cognitive state depends on its preceding history.
To extend this principle to recurrent neural computation,
we develop a Cognitive Field Network (CFN) in which
the internally generated hidden field re-enters
subsequent inference together with new input.

Figure~\ref{fig:cfn_architecture} illustrates the
self-reentrant architecture and its learning framework.
We first establish the dynamical connection between
memory-dressed cognitive fields and discrete field
re-entry, then describe the Transformer implementation,
and finally examine how incoming information drives
and reorganizes the recurrent cognitive state.

\subsection{A. From memory-dressed cognitive dynamics to field re-entry}
\label{sec:cft_to_reentry}

Cognitive Field Theory (CFT) describes cognition as a
collective nonequilibrium dynamical process in which
the present cognitive state is shaped by both incoming
information and the memory of preceding internal dynamics.
In this framework, the collective relaxation modes
generated by learned cognitive organization produce
non-Markovian memory feedback, renormalizing the effective
relaxation of the macroscopic cognitive field.

The central question for a neural implementation is
how such continuous-time, memory-dressed dynamics can
be represented by a discrete recurrent computation.
We establish this connection by introducing the
memory-bearing relaxation modes as components of an
extended dynamical state. 
The essential steps are presented below, while
the detailed construction of the extended state
and its exact discrete-time propagation are
provided in Appendix~A.

The coarse-grained cognitive field $\phi(t)$ obeys
a retarded memory equation of the form
\begin{equation}
\partial_t\phi(t)
=
-r\phi(t)
+
\int_0^\infty d\tau\,
K(\tau)\phi(t-\tau)
+
I_{\rm eff}(t),
\label{eq:cfn_bare_memory}
\end{equation}
where $r$ is the local relaxation coefficient before
memory renormalization, $K(\tau)$ is the causal
memory kernel, and $I_{\rm eff}(t)$ denotes the
effective input drive. The full dynamical theory
also includes fluctuations, which are omitted here
to focus on the deterministic recurrent structure.

In frequency space, the corresponding retarded
response function is
\begin{equation}
\chi_R^{-1}(\Omega)
=
-i\Omega+r-\Sigma_R(\Omega),
\quad
\Sigma_R(\Omega)
=
\int_0^\infty d\tau\,
K(\tau)e^{i\Omega\tau}.
\label{eq:cfn_response}
\end{equation}
The static contribution of memory feedback defines
the renormalized cognitive forgetting gap,
\begin{equation}
r_{\rm cog}
\equiv
r-\Sigma_R(0).
\label{eq:cfn_forgetting_gap}
\end{equation}
Consequently, the inverse response can be
reorganized as
\begin{equation}
\chi_R^{-1}(\Omega)
=
-i\Omega+r_{\rm cog}
-\Delta\Sigma_R(\Omega),
\label{eq:cfn_ren_response}
\end{equation}
where
\(
\Delta\Sigma_R(\Omega)
\equiv
\Sigma_R(\Omega)-\Sigma_R(0).
\)
This representation separates the static
renormalization of the forgetting gap from
the remaining frequency-dependent memory feedback.
Here $\Sigma_R(0)$ is assumed to be finite or
defined under a consistent infrared regularization.

Transforming Eq.~(\ref{eq:cfn_ren_response})
back to the time domain gives the equivalent
renormalized cognitive-field equation,
\begin{equation}
\begin{aligned}
\partial_t\phi(t)
=
-r_{\rm cog}\phi(t)
&+
\int_0^\infty d\tau\,
K(\tau)
\bigl[\phi(t-\tau)-\phi(t)\bigr]
\\
&+
I_{\rm eff}(t).
\label{eq:cfn_ren_memory}
\end{aligned}
\end{equation}
The subtraction removes the static part of the
memory self-energy already incorporated into
$r_{\rm cog}$, thereby avoiding double counting.
Equation~(6) is written for a prescribed past
history; finite-time initialization requires
the corresponding history terms, as discussed
in Appendix~A.

To express the remaining memory feedback
through dynamical state variables, consider
a finite approximation to the relaxation
spectrum,
\begin{equation}
K(\tau)
\simeq
\sum_{\alpha=1}^{M}
g_\alpha e^{-\lambda_\alpha\tau},
\qquad
\lambda_\alpha>0,
\label{eq:cfn_spectral_kernel}
\end{equation}
where $\lambda_\alpha$ denotes the relaxation
rate of mode $\alpha$ and $g_\alpha$ its
effective coupling weight. The continuum
limit corresponds to a coupling-weighted
spectral density, which is related to the
time-scale density of states (TDOS) when
the mode couplings are sufficiently regular.
For simplicity, Eq.~(\ref{eq:cfn_spectral_kernel})
uses purely relaxational modes; oscillatory
modes can be incorporated by extending the
auxiliary dynamics to complex-conjugate
mode pairs.

For each relaxation mode, introduce an
auxiliary memory coordinate
\begin{equation}
m_\alpha(t)
\equiv
\int_0^\infty d\tau\,
e^{-\lambda_\alpha\tau}
\phi(t-\tau).
\label{eq:cfn_memory_modes}
\end{equation}
Its time evolution follows directly from
differentiation,
\begin{equation}
\partial_t m_\alpha(t)
=
-\lambda_\alpha m_\alpha(t)+\phi(t).
\label{eq:cfn_auxiliary_evolution}
\end{equation}
Using
$\int_0^\infty e^{-\lambda_\alpha\tau}d\tau
=1/\lambda_\alpha$,
the renormalized memory term becomes
\begin{equation}
\int_0^\infty d\tau\,
K(\tau)
\bigl[\phi(t-\tau)-\phi(t)\bigr]
\simeq
\sum_{\alpha=1}^{M}
g_\alpha
\left[
m_\alpha(t)
-
\frac{\phi(t)}{\lambda_\alpha}
\right].
\label{eq:cfn_memory_mode_expansion}
\end{equation}

The non-Markovian field equation is thus
equivalent, within the finite-mode
representation, to the coupled system
\begin{equation}
\begin{aligned}
\partial_t\phi(t)
&=
-r_{\rm cog}\phi(t)
+
\sum_{\alpha=1}^{M}
g_\alpha
\left[
m_\alpha(t)
-
\frac{\phi(t)}{\lambda_\alpha}
\right]
+
I_{\rm eff}(t),
\\
\partial_t m_\alpha(t)
&=
-\lambda_\alpha m_\alpha(t)
+
\phi(t).
\end{aligned}
\label{eq:cfn_coupled_memory_modes}
\end{equation}
Although the reduced field $\phi(t)$ depends
on its entire preceding trajectory, the
combined system evolves through its
instantaneous state. The non-Markovian
dynamics has therefore been embedded in
a Markovian system of enlarged dimension.

Defining the extended state
\begin{equation}
\mathbf{\Phi}(t)
=
\bigl(
\phi(t),m_1(t),\ldots,m_M(t)
\bigr)^T,
\label{eq:cfn_extended_state}
\end{equation}
we obtain the state-space equation
\begin{equation}
\partial_t\mathbf{\Phi}(t)
=
A_{\rm ren}\mathbf{\Phi}(t)
+
B I_{\rm eff}(t),
\label{eq:cfn_state_space}
\end{equation}
where the effective evolution matrix
$A_{\rm ren}$ incorporates the renormalized
forgetting gap, the relaxation rates, and
the memory-mode couplings.

Integrating Eq.~(\ref{eq:cfn_state_space})
over a computational interval
$\Delta t=t_{n+1}-t_n$ gives
\begin{equation}
\begin{aligned}
\mathbf{\Phi}_{n+1}
={}&
e^{A_{\rm ren}\Delta t}\mathbf{\Phi}_n
\\
&+
\int_{t_n}^{t_{n+1}}ds\,
e^{A_{\rm ren}(t_{n+1}-s)}
B I_{\rm eff}(s).
\end{aligned}
\label{eq:cfn_exact_discrete}
\end{equation}
For an input held constant over each
interval, the evolution reduces to
\begin{equation}
\mathbf{\Phi}_{n+1}
=
A_d\mathbf{\Phi}_n
+
B_d X_{n+1},
\label{eq:cfn_discrete_linear}
\end{equation}
where
\begin{equation}
A_d=e^{A_{\rm ren}\Delta t},
\qquad
B_d=
\int_0^{\Delta t}
e^{A_{\rm ren}s}B\,ds.
\label{eq:cfn_discrete_operators}
\end{equation}
Here the effective input is represented
by the interval-dependent input variable
$X_{n+1}$, with any required input
projection absorbed into $B$.

Importantly, the memory structure is retained under
this extended-state discretization. Partitioning
the discrete propagator into the collective field
and auxiliary memory sectors,
\begin{equation}
A_d=
\begin{pmatrix}
A_{\phi\phi}&A_{\phi m}\\
A_{m\phi}&A_{mm}
\end{pmatrix},
\label{eq:cfn_discrete_blocks}
\end{equation}
and eliminating the auxiliary coordinates yields
an effective discrete-time memory kernel,
\begin{equation}
K_k^{(d)}
=
A_{\phi m}A_{mm}^{\,k}A_{m\phi},
\qquad k=0,1,2,\ldots .
\label{eq:cfn_discrete_kernel}
\end{equation}
Thus, the one-step evolution of the extended
state retains a history-dependent response
when projected onto the collective field.
The relaxation spectrum and its coupling
weights jointly determine the resulting
memory kernel, as derived in Appendix~A.

More generally, nonlinear interactions
and input-dependent reorganization lead
to an extended-state equation
\begin{equation}
\partial_t\mathbf{\Phi}(t)
=
\mathcal{G}_\theta
\bigl(\mathbf{\Phi}(t),I_{\rm eff}(t)\bigr).
\label{eq:cfn_nonlinear_state}
\end{equation}
Its evolution over a finite interval
defines a nonlinear recurrent map,
\begin{equation}
\mathbf{\Phi}_{n+1}
=
\mathcal{U}_{\theta,\Delta t}
\bigl(\mathbf{\Phi}_n,X_{n+1}\bigr).
\label{eq:cfn_discrete_nonlinear}
\end{equation}

This construction establishes the
dynamical connection between
memory-dressed cognitive-field evolution
and discrete recurrent computation:
the preceding internal state and the
incoming input jointly determine the
next state, while the memory-bearing
degrees of freedom preserve information
about preceding dynamics.

Motivated by this structure, we define
the Cognitive Field Network (CFN)
through the recurrent hidden-field update
\begin{equation}
\Phi_{n+1}
=
F_\theta(X_{n+1},\Phi_n),
\label{eq:cfn_reentry_equation}
\end{equation}
as illustrated in Fig.~\ref{fig:cfn_architecture}(a).
Here $\Phi_n$ denotes the token-resolved
hidden field carried between computational
cycles. The CFN field is a learned neural
representation rather than an explicit
identification with the individual
relaxation-mode coordinates $m_\alpha$.
Accordingly, the correspondence is at
the level of recurrent dynamical structure,
not a claim that a particular Transformer
implementation uniquely follows from
the continuous-time theory.

The architecture provides a pathway
through which the preceding internal
field can influence subsequent inference,
without prescribing an explicit memory
storage or retrieval algorithm. Whether
learning organizes this pathway into
persistent, content-dependent dynamics
is an empirical question. In the
following subsection, we describe
the Transformer implementation of
field re-entry and subsequently
examine its dynamical and functional
properties through controlled experiments.

\subsection{B. Recurrent field re-entry and its functional organization}
\label{sec:cfn_reentry}

We implement the recurrent dynamics using a decoder-style
Transformer, as illustrated in Fig.~\ref{fig:cfn_architecture}(b).
The current token sequence follows the ordinary causal
Transformer pathway, while the final hidden field generated
during the preceding cycle is supplied as recurrent
cross-attention memory.

Let
\begin{equation}
X_t
=
(x_{t,1},x_{t,2},\ldots,x_{t,T})
\end{equation}
denote the token sequence presented during cycle $t$.
The initial hidden representation is
\begin{equation}
H_t^{(0)}
=
E_t
=
\operatorname{Embed}(X_t)
\in
\mathbb{R}^{T\times d},
\label{eq:cfn_embedding}
\end{equation}
where $T$ is the sequence length and $d$ the Transformer
hidden dimension.

The recurrent field entering the current cycle is the
final hidden field generated during the preceding cycle,
\begin{equation}
\Phi_{t-1}
\equiv
H_{t-1}^{(L)},
\label{eq:cfn_previous_field}
\end{equation}
where $L$ denotes the number of Transformer layers.
Thus, $\Phi_{t-1}$ is the full token-resolved hidden
field generated by the network itself rather than a
previous language-model output.

We retain the standard pre-layer normalization and
residual structure of the decoder while suppressing
architectural details not essential to the recurrent
construction. At each layer $\ell=1,\ldots,L$, the
current-cycle representation first undergoes masked
causal self-attention,
\begin{equation}
U_t^{(\ell)}
=
H_t^{(\ell-1)}
+
\operatorname{SA}^{(\ell)}
\left[
\operatorname{LN}
\left(
H_t^{(\ell-1)}
\right)
\right].
\label{eq:cfn_self_attention}
\end{equation}

The resulting representation interacts with the
preceding recurrent field through multi-head
cross-attention,
\begin{equation}
R_t^{(\ell)}
=
U_t^{(\ell)}
+
g\,
\operatorname{MHCA}^{(\ell)}
\left[
\operatorname{LN}
\left(
U_t^{(\ell)}
\right),
\Phi_{t-1}
\right],
\label{eq:cfn_cross_attention}
\end{equation}
where $g$ is a learnable recurrent coupling.

For attention head $h$, the current-cycle
representation provides the queries, while the
preceding hidden field provides the keys and values,
\begin{align}
Q_{t,h}^{(\ell)}
&=
W_{Q,h}^{(\ell)}
\operatorname{LN}
\left(
U_t^{(\ell)}
\right),
\\
K_{\Phi,t-1,h}^{(\ell)}
&=
W_{K,h}^{(\ell)}
\Phi_{t-1},
\\
V_{\Phi,t-1,h}^{(\ell)}
&=
W_{V,h}^{(\ell)}
\Phi_{t-1}.
\label{eq:cfn_qkv}
\end{align}

The recurrent readout of each head is
\begin{equation}
Z_{t,h}^{(\ell)}
=
\operatorname{softmax}
\left[
\frac{
Q_{t,h}^{(\ell)}
K_{\Phi,t-1,h}^{(\ell)\mathsf T}
}{
\sqrt{d_h}
}
\right]
V_{\Phi,t-1,h}^{(\ell)},
\label{eq:cfn_head_attention}
\end{equation}
and the head outputs are concatenated and projected
back to the full hidden dimension,
\begin{equation}
\operatorname{MHCA}^{(\ell)}
=
W_O^{(\ell)}
\operatorname{Concat}
\left[
Z_{t,1}^{(\ell)},\ldots,Z_{t,H}^{(\ell)}
\right].
\label{eq:cfn_mhca}
\end{equation}

The complete token-resolved representation of the
preceding field remains available to every Transformer
layer, allowing attention heads to selectively address
different components of that field. Unlike causal
self-attention within the current token sequence,
cross-attention to $\Phi_{t-1}$ is unmasked because
it operates on a completed preceding cycle.

The recurrent operation is inserted after causal
self-attention and before the feed-forward transformation
at every Transformer layer. The block is completed by
the standard feed-forward update,
\begin{equation}
H_t^{(\ell)}
=
R_t^{(\ell)}
+
\operatorname{FFN}^{(\ell)}
\left[
\operatorname{LN}
\left(
R_t^{(\ell)}
\right)
\right].
\label{eq:cfn_ffn}
\end{equation}

After the final layer, the resulting hidden field becomes
\begin{equation}
\Phi_t
=
H_t^{(L)},
\label{eq:cfn_collective_state}
\end{equation}
which is retained for re-entry during the next cycle.
The external language-model prediction is obtained
through the standard readout,
\begin{equation}
y_t
=
\operatorname{LMHead}
\left(
\Phi_t
\right).
\label{eq:cfn_lm_output}
\end{equation}

The recurrent pathway and the language-model readout
originate from the same final hidden field, but only
the hidden field is propagated internally across
computational cycles. The architecture therefore
retains the standard causal Transformer computation
while introducing recurrent cross-attention to the
preceding token-resolved field.

The architecture does not prescribe a separate memory
store or explicit write--read operations. The recurrent
cross-attention parameters and coupling $g$ are
optimized jointly with the language model, allowing
the functional use of the re-entry pathway to be
determined through task learning, as illustrated
in Fig.~\ref{fig:cfn_architecture}(c).

The dynamical contribution of cross-attention can be
characterized by linearizing the learned recurrent
update around a reference hidden field. For a fixed
input, the cross-cycle Jacobian is defined as
\begin{equation}
J_{\rm rec}
=
\left.
\frac{\partial F_\theta(X_{n+1},\Phi)}
{\partial\Phi}
\right|_{\Phi=\Phi_n}.
\label{eq:cfn_recurrent_jacobian}
\end{equation}
This operator includes the dependence of the
subsequent hidden field on the preceding field
through cross-attention and the remaining
Transformer transformations.

For a stationary linearization, the recurrent
state may be projected onto a collective field
coordinate and complementary internal degrees
of freedom. Partitioning the Jacobian into
the corresponding blocks,
\begin{equation}
J_{\rm rec}
=
\begin{pmatrix}
J_{\phi\phi}&J_{\phi m}\\
J_{m\phi}&J_{mm}
\end{pmatrix},
\label{eq:cfn_recurrent_blocks}
\end{equation}
the elimination of the complementary coordinates
defines an effective discrete memory kernel,
\begin{equation}
K_k^{\rm CFN}
=
J_{\phi m}J_{mm}^{\,k}J_{m\phi},
\qquad k=0,1,2,\ldots .
\label{eq:cfn_recurrent_kernel}
\end{equation}

Cross-attention is therefore not itself the
temporal memory kernel, but contributes to
the learned recurrent couplings from which
an effective memory response can be constructed.
This provides a quantitative route for testing
the correspondence between CFN recurrence
and the memory-dressed dynamics of CFT.
The derivation and its assumptions are
presented in Appendix~A.

Because the recurrent hidden field can be retained,
replaced, or disabled during evaluation, its causal
contribution to subsequent inference can be tested
directly. This permits controlled measurements of
content-dependent persistence, passive relaxation,
and input-driven reorganization across computational
cycles. 
The following subsection examines how incoming
information excites collective relaxation modes
and reorganizes the memory-dressed cognitive field,
providing a dynamical interpretation of recurrent
inference.

\subsection{C. From driven cognitive-field dynamics to recurrent inference}
\label{sec:cfn_driven_field}

The preceding subsections established the connection
between memory-dressed cognitive dynamics and discrete
field re-entry, together with its Transformer
implementation. We now examine how incoming information
acts on the collective relaxation modes and reorganizes
the existing cognitive field.

A defining feature of cognitive inference is that
external input acts as a finite drive rather than
merely an infinitesimal perturbation used to measure
linear response. Its effect depends on the dynamical
organization of the internal state through which
the incoming information is processed.

Consider the projected collective dynamics obtained
by separating the macroscopic field $\phi(t)$ from
the complementary relaxation sector $\mathcal{X}(t)$.
The corresponding driven equations are
\begin{align}
\partial_t\phi(t)
&=
-r\phi(t)
+
D\mathcal{X}(t)
+
B_\phi I(t)
+
\xi_\phi(t),
\label{eq:cfn_projected_field}
\\
\partial_t\mathcal{X}(t)
&=
-M\mathcal{X}(t)
+
C\phi(t)
+
B_{\mathcal X}I(t)
+
\xi_{\mathcal X}(t).
\label{eq:cfn_projected_modes}
\end{align}
Here $M$ is the projected relaxation operator,
$C$ and $D$ describe the coupling between the
collective field and the complementary sector,
and $B_\phi$ and $B_{\mathcal X}$ describe the
corresponding projections of the external input.
The projection procedure and the resulting
memory-feedback and input-transfer structures
are derived in Appendix~B.

For a diagonalizable, generally non-Hermitian
relaxation operator, the complementary dynamics
can be expanded in biorthogonal modes with
complex relaxation rates
\begin{equation}
\mu_\alpha
=
\lambda_\alpha+i\omega_\alpha,
\label{eq:cfn_complex_rates}
\end{equation}
where $\lambda_\alpha$ determines the relaxation
rate and $\omega_\alpha$ the intrinsic temporal
phase evolution.

The resulting mode dynamics takes the form
\begin{equation}
\partial_t\mathcal{X}_\alpha(t)
=
-\mu_\alpha\mathcal{X}_\alpha(t)
+
c_\alpha\phi(t)
+
b_\alpha(t)
+
\eta_\alpha(t),
\label{eq:cfn_driven_modal}
\end{equation}
where the input-induced modal excitation is
\begin{equation}
b_\alpha(t)
=
\widetilde u_\alpha^\dagger
B_{\mathcal X}I(t).
\label{eq:cfn_modal_input}
\end{equation}
Here $\widetilde u_\alpha$ is the corresponding
left eigenvector of $M$, and
$c_\alpha=\widetilde u_\alpha^\dagger C$.

Equation~(\ref{eq:cfn_modal_input}) identifies
the mode-selective action of incoming information.
The excitation of each relaxation mode depends
on the projection of the input onto the learned
collective dynamical directions. Consequently,
different inputs can produce different patterns
of modal activation, even when their overall
magnitudes are comparable.

The field and input therefore interact through
two coupled pathways. The existing field drives
the relaxation sector through $c_\alpha\phi$,
while incoming information excites that sector
through $b_\alpha(t)$. The activated modes
subsequently contribute to the evolution of
the macroscopic field.

Eliminating the complementary relaxation modes
produces the memory-dressed field equation
introduced in Sec.~II.A. In the stationary
linearized regime, its frequency-domain form is
\begin{equation}
\left[
-i\Omega+r-\Sigma_R(\Omega)
\right]
\phi(\Omega)
=
\mathcal{T}_I(\Omega)I(\Omega)
+
\xi_{\rm eff}(\Omega),
\label{eq:cfn_driven_frequency}
\end{equation}
where initial transients have been omitted.

The retarded memory self-energy is
\begin{equation}
\Sigma_R(\Omega)
=
\sum_\alpha
\frac{d_\alpha c_\alpha}
{\mu_\alpha-i\Omega},
\label{eq:cfn_driven_self_energy}
\end{equation}
while the effective input-transfer function is
\begin{equation}
\mathcal{T}_I(\Omega)
=
B_\phi
+
\sum_\alpha
\frac{d_\alpha q_\alpha}
{\mu_\alpha-i\Omega},
\qquad
q_\alpha
=
\widetilde u_\alpha^\dagger B_{\mathcal X}.
\label{eq:cfn_input_transfer}
\end{equation}
Here $d_\alpha=Du_\alpha$, with $u_\alpha$
denoting the corresponding right eigenvector.

Defining the memory-dressed retarded response
\begin{equation}
\chi_R(\Omega)
=
\frac{1}
{-i\Omega+r-\Sigma_R(\Omega)},
\label{eq:cfn_cognitive_propagator}
\end{equation}
the driven field response becomes
\begin{equation}
\phi(\Omega)
=
\chi_R(\Omega)
\mathcal{T}_I(\Omega)I(\Omega)
+
\chi_R(\Omega)\xi_{\rm eff}(\Omega).
\label{eq:cfn_driven_field_solution}
\end{equation}

This expression separates two complementary
aspects of cognitive inference. The transfer
function $\mathcal{T}_I(\Omega)$ determines
how incoming information excites the collective
dynamical manifold, whereas $\chi_R(\Omega)$
determines how the resulting excitation
propagates through the memory-dressed field.

The same relaxation spectrum contributes to
both quantities, but with different coupling
weights. The self-energy contains the
field--mode feedback weights $d_\alpha c_\alpha$,
whereas the input-transfer function contains
the input--mode weights $d_\alpha q_\alpha$.
Thus, memory feedback and input-driven
reorganization are dynamically related
without being identical processes.

The finite cognitive input $I(t)$ should
also be distinguished from an infinitesimal
probe used to define susceptibility. 
The former drives the evolving cognitive state,
whereas the latter characterizes its response
around a specified dynamical background.
This distinction is developed further in
Appendix~B.

In the CFN, this input-driven reorganization
is implemented through the learned mapping
between the current input and the preceding
hidden field. The continuous collective
coordinates and the token-resolved recurrent
field need not be identified microscopically.
Rather, the architecture provides a
computational realization of the principle
that new information acts on an internal
state carrying the consequences of
preceding computation.

The resulting recurrent dynamics can be
tested by comparing inference under correct,
incorrect, and disabled preceding fields.
Further experiments examine how the learned
state persists without relevant input and
how subsequent information reorganizes
the surviving field. These measurements
distinguish the availability of a recurrent
pathway from its learned, content-dependent
functional organization.

\begin{table*}[t]
\centering
\caption{
Architecture of the decoder-style Cognitive Field Network used
in this work.
The final token-resolved hidden field from each computational cycle is
retained and supplied as unmasked recurrent cross-attention memory to
every Transformer layer of the subsequent cycle.
}
\label{tab:cfn_architecture}

\begin{tabular*}{\textwidth}{@{\extracolsep{\fill}}lcc@{}}
\hline\hline
Component & Specification \\
\hline
Backbone & GPT-NeoX decoder Transformer \\
Number of Transformer blocks & $6$ \\
Hidden dimension $d$ & $512$ \\
Attention heads & $8$ \\
Feed-forward dimension & $2048$ \\
Positional representation & Rotary positional embedding (RoPE) \\
Current-cycle attention & Masked causal self-attention \\
Recurrent-field source & Final hidden field $H_{t-1}^{(L)}=\Phi_{t-1}$ \\
Recurrent-field dimension & $T\times d$ \\
Re-entry mechanism & Multi-head recurrent cross-attention \\
Cross-attention heads & $8$ \\
Cross-attention head dimension $d_h$ & $64$ \\
Cross-attention query & Current-cycle hidden representation \\
Cross-attention key/value & Previous-cycle hidden field $\Phi_{t-1}$ \\
Cross-attention masking & Unmasked \\
Re-entry layers & All Transformer layers \\
Re-entry coupling & Learnable scalar $g$ \\
Recurrent-field update & $\Phi_t=H_t^{(L)}$ \\
Language-model readout & $y_t=\operatorname{LMHead}(\Phi_t)$ \\
\hline\hline
\end{tabular*}
\end{table*}

\section{III. Emergence of Recurrent Cognitive-State Dynamics}
\label{sec:results}

The preceding section defines the CFN as an architecture in which an
internally generated hidden field can re-enter subsequent computation.
The existence of this pathway alone, however, does not establish that
learning will organize it into a functional cognitive dynamics.

We therefore examine how self-reentry is transformed by learning and
whether the resulting recurrent field acquires measurable dynamical and
informational properties.
We first characterize the learned utilization of the recurrent pathway
and the accompanying reorganization of collective relaxation dynamics.
We then test whether the recurrent field carries content-specific
information, whether that information is causally required for selective
retrieval, and whether its persistence across computational cycles can
itself be extended through learning.

Together, these experiments trace the transition from architectural
self-reentry to a learned, content-bearing, and temporally persistent
cognitive-state dynamics.


\subsection{A. Learning organizes functional recurrent re-entry}
\label{sec:learning_organization}


We first ask whether the recurrent pathway 
remains functionally relevant when the network is
trained under an ordinary language-model objective.
The architecture makes the preceding hidden field available to the
current computation, but it does not prescribe how strongly that field
should influence subsequent processing.
Functional utilization of self-reentry must therefore be organized
through learning.

To examine this question, we trained a compact six-block GPT-NeoX
decoder reorganized into the CFN architecture described in
Sec. II.B.
The model contains approximately $7.67\times10^7$ trainable parameters,
with hidden dimension $d=512$, eight attention heads, and feed-forward
dimension $2048$.
At every Transformer layer, masked causal self-attention is followed by
eight-head recurrent cross-attention to the complete token-resolved
hidden field of the preceding cycle and then by the feed-forward
transformation.
The architectural configuration used throughout the experiments is
summarized in Table~\ref{tab:cfn_architecture}.

The model was pretrained on WikiText-103 using sequences of length
$1024$ for $30\,000$ optimizer steps.
Training and validation measurements were performed on separate data
splits, with the validation set held out from parameter optimization.
A physical batch size of $8$ with eight gradient-accumulation steps gave
an effective batch size of $64$ sequences, corresponding to
$65\,536$ tokens per optimizer update.
The recurrent coupling was initialized at $g_0=0.05$ and optimized
jointly with the remaining model parameters.
The learning rate was warmed up to $10^{-3}$ and subsequently annealed
toward $10^{-4}$ during pretraining.

\begin{figure}[t]
\centering
\includegraphics[scale=0.52, trim= 0.2cm 0.5cm 0cm 0cm]{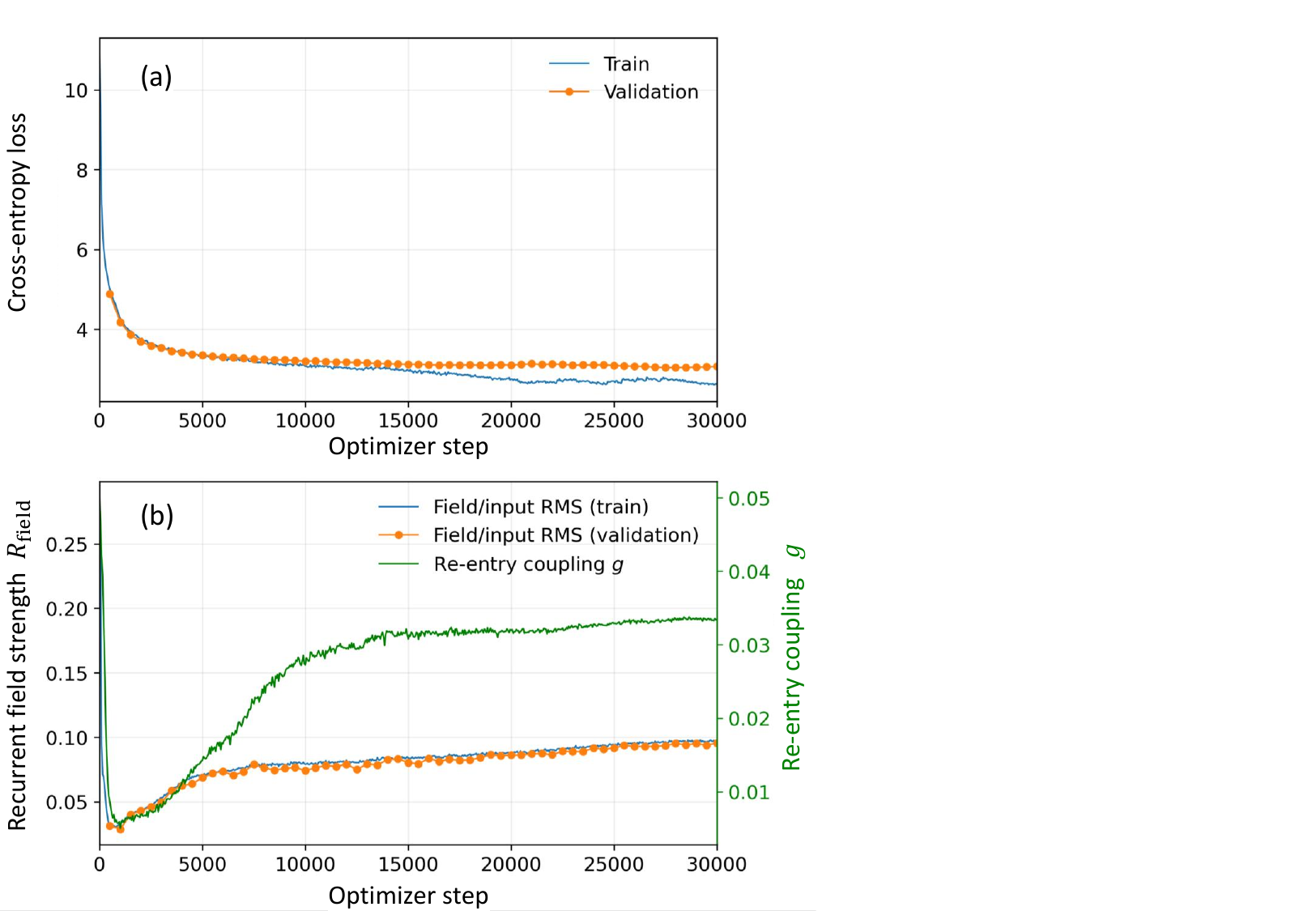}
\caption{
Learning dynamics of recurrent field re-entry during
language-model pretraining.
(a) Training and validation cross-entropy losses for the
GPT-NeoX-based CFN trained on WikiText-103.
Both losses decrease rapidly during early training, followed
by a slower late-training regime.
(b) Evolution of the learnable re-entry coupling $g$ (green)
and the recurrent-field strength $R_{\rm field}$ measured
on the training (blue) and validation (orange) sets.
The recurrent quantities exhibit an initial decrease,
followed by recovery and gradual stabilization.
}
\label{fig:cfn_learning}
\end{figure}

Figure~\ref{fig:cfn_learning} summarizes the language-model optimization
and the simultaneous evolution of the recurrent pathway.
As shown in Fig.~\ref{fig:cfn_learning}(a), the training and validation
cross-entropy losses decrease rapidly during the early stage of
pretraining, reaching values near $4$ within the first few thousand
optimizer steps.
The validation loss subsequently decreases more gradually and approaches
approximately $3.0$ near the end of training, corresponding to a
validation perplexity of approximately $21$.
The training loss continues to decrease below the validation loss,
reaching approximately $2.6$ at the final checkpoint.
Thus, the self-reentrant architecture can be optimized stably under an
ordinary next-token language-model objective.

The recurrent pathway exhibits a more structured evolution
in Fig.~\ref{fig:cfn_learning}(b).
The learnable coupling $g$ does not remain near its initialized value.
Starting from $g_0=0.05$, it is strongly suppressed during the earliest
stage of optimization, reaching a minimum of approximately
$g\simeq5.0\times10^{-3}$ at around $1\,000$ steps.
Its evolution then reverses, and the coupling progressively increases
throughout subsequent training.
At later stages it enters a slowly evolving regime and reaches
approximately $g\simeq3.3\times10^{-2}$ at $30\,000$ steps.
The recurrent connection is therefore not simply retained at its
initialized strength, but is dynamically reweighted as language-model
learning proceeds.

The scalar coupling alone, however, does not determine the actual
contribution of the preceding field to the current hidden computation.
For layer $\ell$, the recurrent contribution is
\begin{equation}
\Delta H_{{\rm field},t}^{(\ell)}
=
g\,
\operatorname{MHCA}^{(\ell)}
\left[
\operatorname{LN}\!\left(U_t^{(\ell)}\right),
\Phi_{t-1}
\right],
\label{eq:effective_recurrent_perturbation_results}
\end{equation}
where the recurrent multi-head cross-attention contains the learned
query, key, value, and output projections.
Its magnitude can therefore evolve differently from the scalar coupling
$g$ itself.

We quantify the effective recurrent contribution by the dimensionless
field-strength ratio
\begin{equation}
R_{\rm field}
=
\frac{
\operatorname{RMS}\!\left(\Delta H_{\rm field}\right)
}{
\operatorname{RMS}\!\left(H_{\rm current}\right)
}.
\label{eq:rfield_results}
\end{equation}
This quantity measures the RMS magnitude of the recurrent-field
perturbation relative to the current hidden representation.

As shown in Fig.~\ref{fig:cfn_learning}(b),
$R_{\rm field}$ undergoes a closely related but distinct reorganization.
Following an early transient, the effective recurrent contribution is
strongly suppressed and then progressively recovers as training proceeds.
By approximately $5\times10^3$--$10^4$ optimizer steps, the recurrent
field has recovered to a substantial finite magnitude and subsequently
strengthens more gradually.
At the final checkpoint, $R_{\rm field}$ reaches approximately $0.10$,
corresponding to a recurrent perturbation with an RMS magnitude of
approximately $10\%$ of the current hidden-state magnitude.

Importantly, the $R_{\rm field}$ trajectories measured on the training
and held-out validation splits closely track one another throughout
learning.
The finite recurrent contribution is therefore not confined to the
sequences used for parameter optimization, but is reproducibly expressed
on unseen validation language data.

Preliminary experiments with simplified re-entry pathways showed the
same qualitative tendency.
Across the full CFN, single-attention, and MLP-based re-entry
configurations, learning produced substantial nonzero recurrent
participation, with representative late-training field-strength ratios
of approximately
$R_{\rm field}\simeq0.10$--$0.15$.
These auxiliary experiments were not designed as a systematic
cross-architecture comparison and therefore do not establish
architecture-independent behavior.
They nevertheless suggest that learned utilization of re-entry is not
restricted to the particular full CFN implementation.
Rather, once a causal pathway is made available through which a
preceding internal state can influence subsequent computation,
ordinary language-model optimization can learn to make functional use
of that pathway.

Taken together, the evolution of $g$ and $R_{\rm field}$ shows that
self-reentry is functionally reorganized during ordinary language-model
pretraining.
The architecture provides the recurrent pathway, whereas learning
determines how strongly and in what form the preceding hidden field
participates in subsequent computation.
The finite late-training recurrent contribution is therefore a learned
operating property of the network rather than a direct consequence of
the initialized coupling.

This establishes the first empirical step from architectural
self-reentry to functional recurrent dynamics.
Whether the learned recurrent field carries identifiable information
across inference cycles, whether that information is causally required
for subsequent computation, and whether it can be selectively retrieved
or renewed are separate functional questions examined below.

\begin{figure*}[t]
\centering
\includegraphics[width=1.0\textwidth, trim=0cm 1.7cm 0cm 0cm]{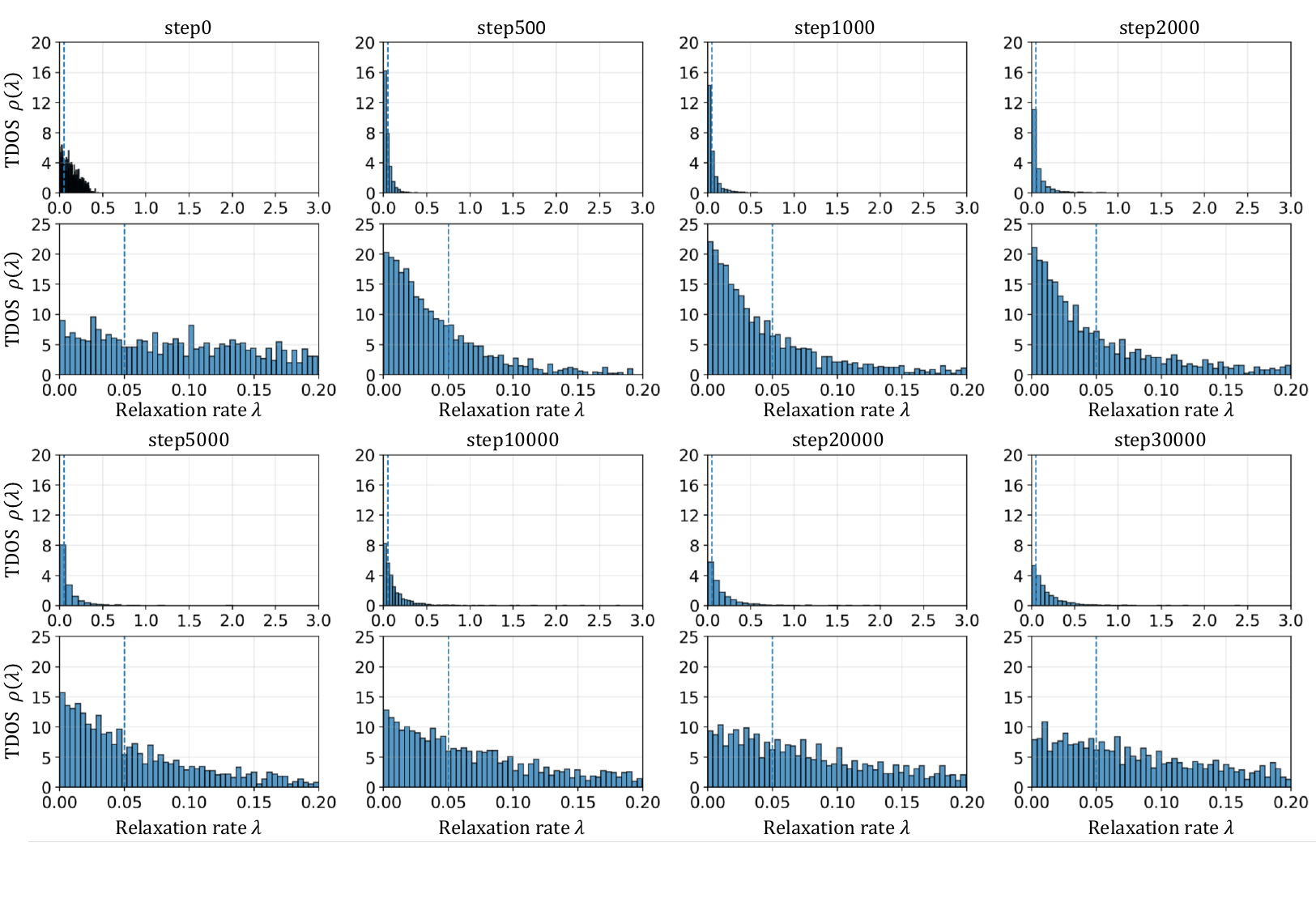}
\caption{
Learning-induced reorganization of the relaxation spectrum.
Time-scale density of states (TDOS) obtained from the local
Jacobian of the $H^{(4)}\rightarrow H^{(5)}$ layer map at
successive language-model pretraining checkpoints.
A fixed probe sequence of length $T=8$ yields a
$4096\times4096$ Jacobian, with relaxation rates defined as
$\lambda_\alpha=-\ln|\mu_\alpha|$, where $\mu_\alpha$ are
the Jacobian eigenvalues.
For each checkpoint, the upper panel shows the full stable
relaxation spectrum, and the lower panel shows the infrared
region $0<\lambda<0.2$.
The dashed vertical line indicates the reference cutoff
$\lambda_c=0.05$.
The infrared spectral weight increases rapidly during early
training and subsequently decreases as the spectrum broadens.
}
\label{fig:cfn_tdos}
\end{figure*}

\begin{figure}[t]
\centering
\includegraphics[scale=0.52, trim= 0cm 0.5cm 0cm 0cm]{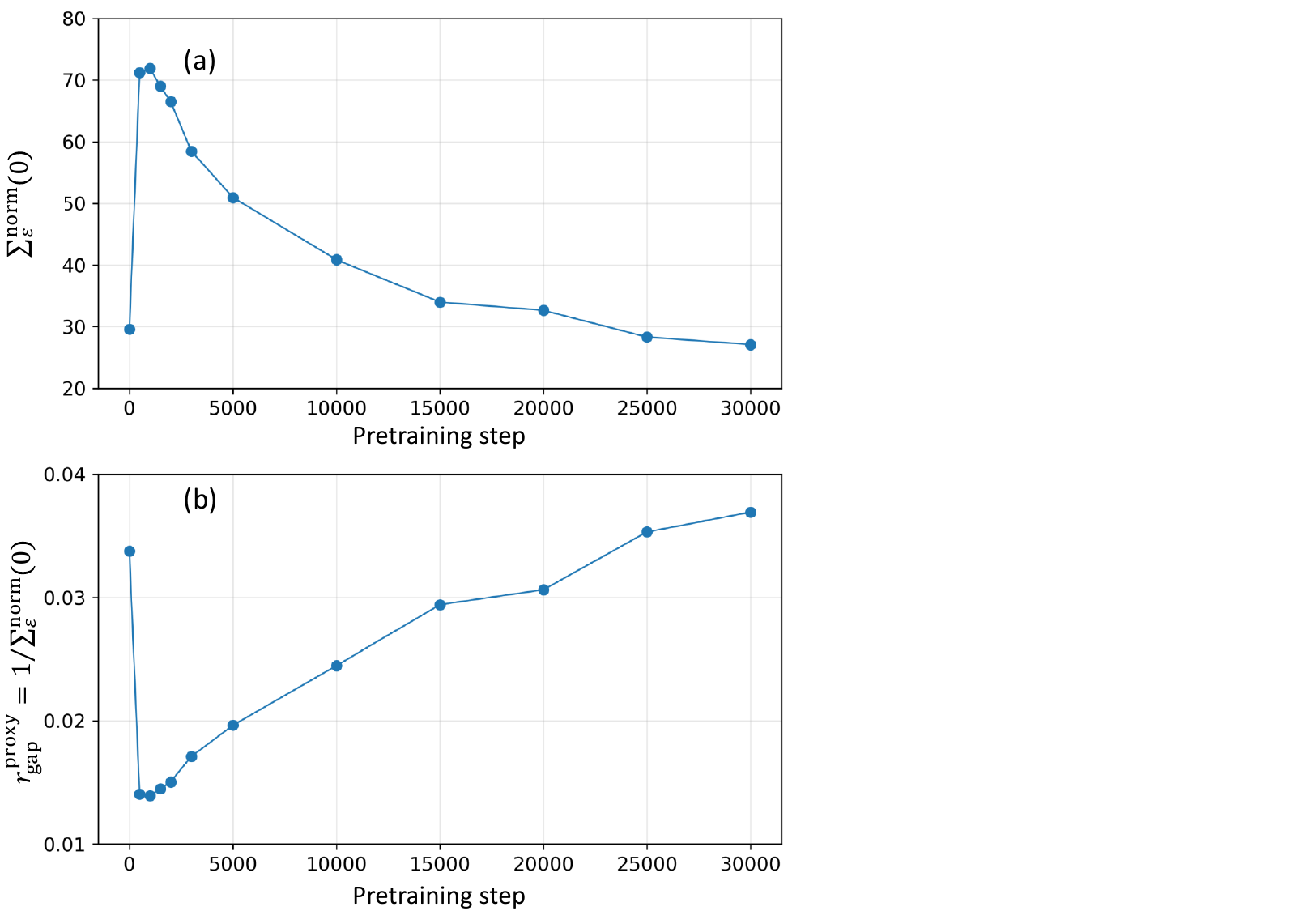}

\caption{
Infrared self-energy and inverse-gap proxies during learning.
(a) Normalized infrared self-energy proxy
$\Sigma_{\epsilon}^{\rm norm}(0)$.
(b) Corresponding inverse-self-energy proxy
$r_{\rm gap}^{\rm proxy}
=1/\Sigma_{\epsilon}^{\rm norm}(0)$.
Both quantities are computed from the stable relaxation
spectrum of the $H^{(4)}\rightarrow H^{(5)}$ Jacobian
at successive pretraining checkpoints, using a fixed
probe sequence of length $T=8$ and an infrared regulator
$\epsilon=10^{-3}$.
}
\label{fig:cfn_selfenergy}
\end{figure}

\subsection{B. Learning reorganizes the infrared collective dynamics}
\label{sec:infrared_reorganization}

The emergence of a finite recurrent pathway shown in
Fig.~\ref{fig:cfn_learning} raises a second question.
Does learning merely adjust the magnitude of recurrent re-entry, or
does it also reorganize the collective dynamics through which hidden
perturbations propagate inside the network?

To address this question, we measured the local collective relaxation
spectrum at successive pretraining checkpoints.
For a fixed probe sequence of length $T=8$, one completed hidden field
was re-entered into the subsequent inference cycle, and the Jacobian of
the layer map from $H^{(4)}$ to $H^{(5)}$ was evaluated.
With hidden dimension $d=512$, this gives a $4096\times4096$ Jacobian.
If $\mu_\alpha$ denotes a Jacobian eigenvalue, we define the associated
relaxation rate as
\begin{equation}
\lambda_\alpha=-\log|\mu_\alpha|,
\label{eq:cfn_relaxation_rate}
\end{equation}
and construct the time-scale density of states (TDOS)
\begin{equation}
\rho(\lambda)
=
\frac{1}{N_+}
\sum_{\alpha:\lambda_\alpha>0}
\delta(\lambda-\lambda_\alpha),
\label{eq:cfn_tdos}
\end{equation}
where $N_+$ is the number of positive relaxation rates.

Figure~\ref{fig:cfn_tdos} shows a pronounced reorganization of this
spectrum during pretraining.
At initialization, the relaxation spectrum is comparatively broad,
with only modest accumulation near the infrared sector
$\lambda\rightarrow0$.
Within the first several hundred optimizer steps, however, substantial
spectral weight is redistributed toward slow relaxation modes.
The infrared enhancement is strongest around steps $500$--$1000$,
where a large fraction of the stable collective modes becomes
concentrated near small relaxation rates.

We quantify this redistribution using the slow-mode fraction
\begin{equation}
f_{\rm slow}(\lambda_c)
=
\frac{1}{N_+}
\sum_{\alpha:\lambda_\alpha>0}
\Theta(\lambda_c-\lambda_\alpha),
\qquad
\lambda_c=0.05.
\label{eq:cfn_slow_fraction}
\end{equation}
The slow-mode fraction increases from approximately $0.251$ at
initialization to $0.683$ at step $500$, corresponding to more than a
twofold increase in the population of slow stable modes.
The redistribution is not monotonic, however.
After this early maximum, $f_{\rm slow}$ progressively decreases and
reaches approximately $0.269$ at step $30000$.

The TDOS therefore reveals a distinct dynamical sequence.
Learning first produces a pronounced transient infrared concentration
and subsequently redistributes spectral weight over a broader range of
relaxation scales.
The early stage of optimization is thus characterized by strong
collective softening rather than by a monotonic accumulation of slow
modes throughout training.

\vspace{5pt}

\paragraph{Infrared self-energy and transient dynamical softening.}

To summarize the infrared contribution of the measured relaxation
spectrum, we define the regularized zero-frequency self-energy proxy
\begin{equation}
\Sigma_{\epsilon}^{\rm norm}(0)
=
\frac{1}{N_+}
\sum_{\alpha:\lambda_\alpha>0}
\frac{1}{\lambda_\alpha+\epsilon},
\qquad
\epsilon=10^{-3}.
\label{eq:cfn_selfenergy_proxy}
\end{equation}
Because each mode is weighted inversely by its relaxation rate, this
quantity is particularly sensitive to collective spectral weight near
$\lambda=0$.
The regulator is fixed across all checkpoints so that the training
trajectory is compared at a common infrared resolution.

As shown in Fig.~\ref{fig:cfn_selfenergy}, the self-energy proxy follows
the same strongly nonmonotonic evolution observed directly in the TDOS.
It increases from approximately $29.60$ at initialization to $71.21$
at step $500$ and reaches approximately $71.88$ at step $1000$.
The strongest infrared dressing therefore occurs during the early stage
of pretraining rather than at the end of optimization.
With continued training, the proxy progressively decreases, reaching
approximately $27.09$ at step $30000$.

For visualization of the corresponding dynamical softening, we also use
the inverse quantity
$r_{\rm gap}^{\rm proxy}=1/\Sigma_{\epsilon}^{\rm norm}(0)$.
This quantity is not an independently measured physical mass or the
exact cognitive forgetting gap.
It is simply an inverse measure of the observed infrared collective
dressing, such that stronger infrared enhancement corresponds to a
smaller $r_{\rm gap}^{\rm proxy}$.
Consistent with the TDOS evolution, it decreases from approximately
$3.38\times10^{-2}$ at initialization to $1.39\times10^{-2}$ near
step $1000$, and subsequently increases to approximately
$3.69\times10^{-2}$ at step $30000$.

The TDOS, slow-mode fraction, and self-energy proxy therefore provide
mutually consistent signatures of a transiently softened collective
regime during early learning.
As relaxation rates accumulate toward $\lambda\rightarrow0^{+}$,
an increasing fraction of perturbations decays over progressively
longer time scales, while the simultaneous increase of
$\Sigma_{\epsilon}^{\rm norm}(0)$ shows that this behavior reflects a
collective redistribution of infrared spectral weight rather than an
isolated near-zero mode.

In Cognitive Field Theory, the physical significance of such an
infrared reorganization follows from the memory kernel generated by the
collective relaxation spectrum,
\begin{equation}
K(t)
=
\int_{0}^{\infty}
d\lambda\,
\rho(\lambda)e^{-\lambda t},
\label{eq:cfn_memory_kernel}
\end{equation}
so that redistribution of spectral weight toward
$\lambda\rightarrow0^{+}$ enhances the long-time contribution to the
collective memory kernel.
Correspondingly, the zero-frequency self-energy has the infrared
structure
\begin{equation}
\Sigma_R(0)
\sim
\int_{0}^{\infty}
d\lambda\,
\frac{\rho(\lambda)}{\lambda},
\label{eq:cfn_ir_dressing}
\end{equation}
which shows directly why slow collective modes contribute strongly to
the dressing of the macroscopic field.

The measured $\Sigma_{\epsilon}^{\rm norm}(0)$ should therefore be
understood as an operational spectral proxy for this infrared
enhancement rather than as a direct measurement of the full
field-theoretic self-energy.
Within CFT, the corresponding memory dressing suppresses the cognitive
forgetting gap according to
\begin{equation}
r_{\rm cog}=r-\Sigma_R(0),
\label{eq:cfn_cognitive_gap}
\end{equation}
with the static susceptibility scaling as
$\chi_R(0)=1/r_{\rm cog}$.
Increasing infrared spectral weight therefore corresponds, within this
field-theoretic description, to slower collective relaxation and
enhanced susceptibility to subsequent input.

Importantly, the relevant collective regime is not one in which all
relaxation rates collapse to zero.
Exact gap closing, $r_{\rm cog}=0$, represents the critical boundary
rather than a stable operating state.
The protected near-critical regime proposed by CFT instead corresponds
to $0<r_{\rm cog}\ll\Lambda$, where finite forgetting and dynamical
stability coexist with long collective time scales and enhanced
susceptibility.

The observed training trajectory is qualitatively consistent with this
picture.
Around steps $500$--$1000$, the strong accumulation of slow modes and
the maximum of the self-energy proxy indicate a transient approach
toward a critically softened collective regime.
With continued optimization, however, the excess infrared weight
decreases and the relaxation spectrum broadens rather than collapsing
further toward $\lambda=0$.
At the same time, the recurrent pathway remains finite and subsequently
strengthens.
Learning therefore appears to pass through strong collective softening
before settling into a dynamically stable recurrent regime.

This distinction is important for persistent cognitive dynamics.
A completely frozen system could preserve perturbations but would have
little capacity for continual reorganization, whereas a strongly gapped
system would rapidly erase them.
A near-critical collective organization instead provides an
intermediate dynamical regime in which previous information can remain
influential while the recurrent field continues to respond to new
input.

The present measurements do not directly determine $\Sigma_R(0)$,
$r_{\rm cog}$, or a thermodynamic critical point.
Rather, the measured TDOS and
$\Sigma_{\epsilon}^{\rm norm}(0)$ provide spectral evidence for the
underlying infrared collective organization predicted by this
field-theoretic picture.
Whether the resulting recurrent field actually acquires
content-specific macroscopic organization is a separate functional
question.

This spectral reorganization is especially informative when
compared with the recurrent-pathway measurements in
Fig.~\ref{fig:cfn_learning}.
The recurrent coupling $g$ and effective field strength
$R_{\rm field}$ characterize the participation of the
preceding hidden field in subsequent computation, whereas
the TDOS characterizes the collective relaxation structure
through which perturbations propagate.
Their distinct training trajectories show that recurrent
participation and infrared spectral organization do not
evolve in a simply proportional manner.

Taken together, Figs.~\ref{fig:cfn_learning}--\ref{fig:cfn_selfenergy}
show that learning neither monotonically strengthens
recurrent coupling nor continuously drives the network
toward slower relaxation.
Instead, persistent cognitive-state dynamics involves
the joint organization of field re-entry and collective
relaxation, rather than a static memory register or an
indefinitely softening dynamical system.

\begin{figure*}[t]
\centering
\includegraphics[width=0.93\textwidth, trim=0cm 6cm 0cm 0cm]{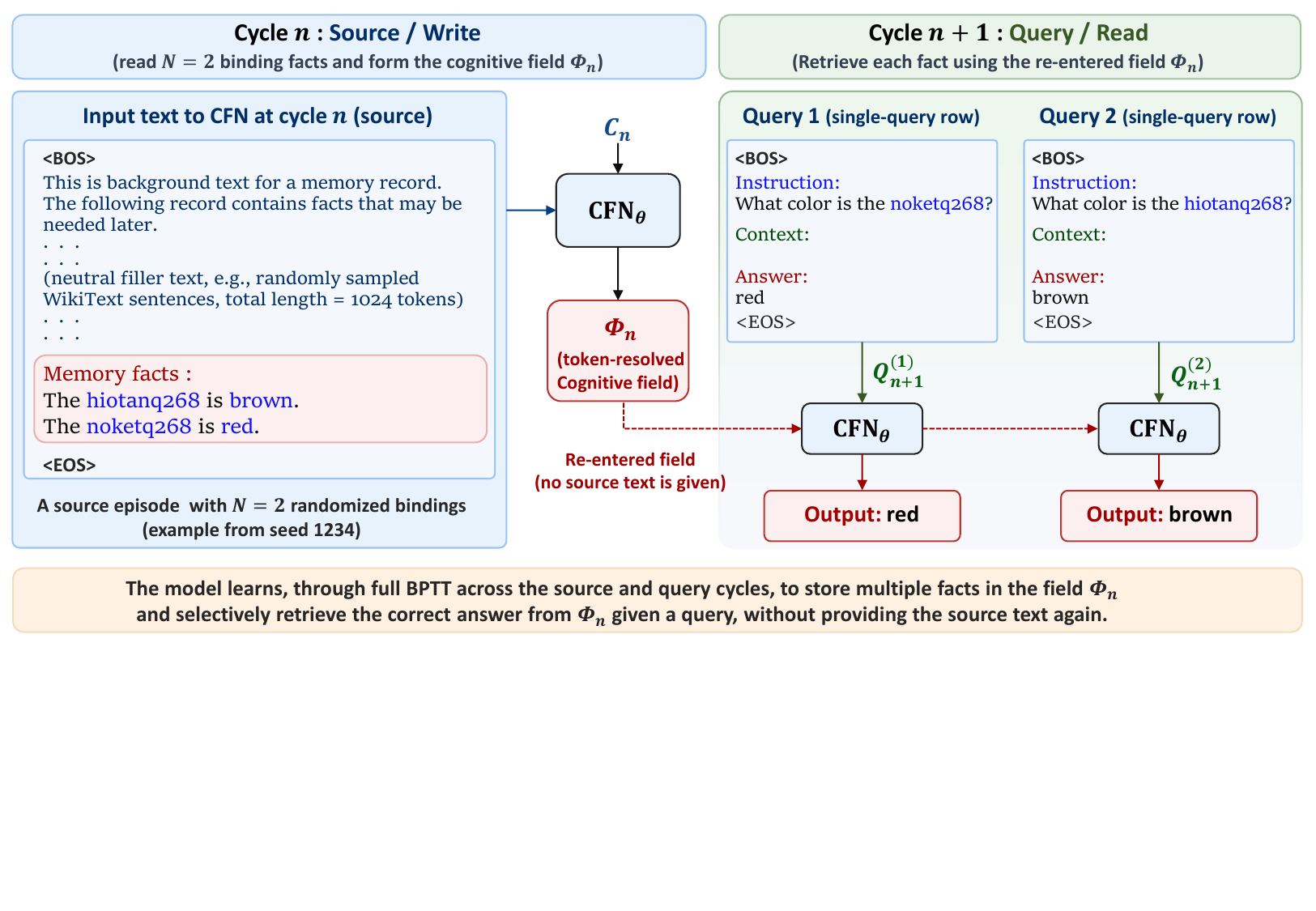}
\caption{
Cross-cycle protocol for learning content-specific memory
and selective retrieval.
During the source (WRITE) cycle $n$, the CFN receives
a sequence containing $N$ randomly generated
entity--attribute bindings (illustrated for $N=2$)
and produces the token-resolved recurrent field
$\Phi_n=H_n^{(L)}$.
During the subsequent QUERY cycle $n+1$, the source text
is no longer available, and the field $\Phi_n$ is re-entered
through recurrent cross-attention.
Independent single-query inputs retrieve different
attributes from the same preceding field $\Phi_n$.
Training uses full backpropagation through both cycles,
with answer supervision applied only during the QUERY cycle.
}
\label{fig:selective_retrieval_protocol}
\end{figure*}

\begin{figure}[t]
\centering
\includegraphics[scale=0.5, trim= 0.1cm 8.8cm 0cm 0cm]{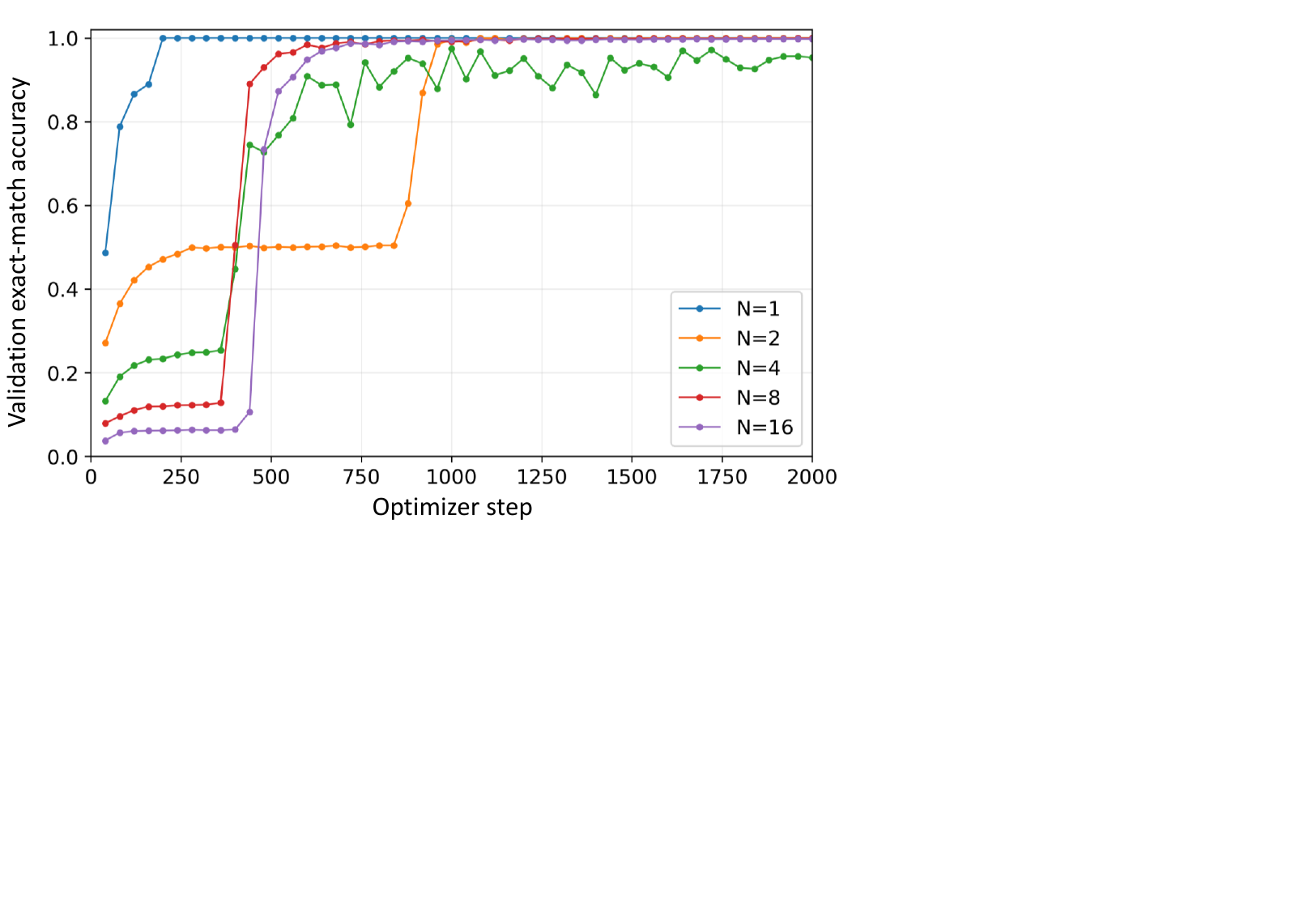}
\caption{
Learning dynamics of selective retrieval from the
re-entered cognitive field.
Validation exact-match accuracy during cross-cycle
retrieval training for memory loads $N=1,2,4,8,$ and $16$,
where $N$ denotes the number of entity--attribute bindings
presented during the source cycle.
The $N=1$ task reaches near-perfect accuracy rapidly,
whereas larger memory loads exhibit distinct learning
trajectories.
For $N=2$, accuracy remains near $0.5$ before rising
sharply toward unity.
The $N=8$ and $N=16$ conditions also show abrupt
improvements, while $N=4$ exhibits greater fluctuations
before reaching high retrieval accuracy.
}
\label{fig:selective_retrieval_learning}
\end{figure}

\subsection{C. Emergence of content-specific memory and selective retrieval}
\label{sec:selective_retrieval}

The preceding results establish two properties of the recurrent CFN.
First, language-model pretraining organizes the available re-entry
pathway into a finite component of the learned computation.
Second, this learning process is accompanied by a pronounced
reorganization of the collective relaxation spectrum and its infrared
sector.
Neither observation, however, establishes that the re-entered field
actually carries identifiable information that can later be recovered.
A finite recurrent perturbation could influence subsequent computation
without functioning as a content-specific memory.

We therefore next ask whether information presented during one
inference cycle can be encoded into the recurrent field and selectively
retrieved during a subsequent cycle after the original external context
has been removed.
Figure~\ref{fig:selective_retrieval_protocol} illustrates the controlled
cross-cycle retrieval task used to test this question.

During a source or WRITE cycle $n$, the model receives a sequence
containing $N$ randomly generated entity--attribute bindings.
The CFN processes the complete source sequence and produces the
token-resolved recurrent field
\begin{equation}
\Phi_n=H_n^{(L)}.
\label{eq:write_field}
\end{equation}
The source text is then removed.
It is not concatenated to the subsequent input, replayed as context,
or otherwise supplied to the model again.
The only source-dependent neural variable made available to the next
cycle is the internally generated recurrent field $\Phi_n$.

During the subsequent QUERY cycle, the model receives a question about
one of the bindings, while the external query contains no copy of the
source facts.
The computation therefore has the form
\begin{equation}
(Q_{n+1},\Phi_n)
\longrightarrow
H_{n+1}^{(L)}
\longrightarrow
\operatorname{LMHead}
\longrightarrow
A,
\label{eq:query_from_field}
\end{equation}
where $Q_{n+1}$ denotes the current query and $A$ is the predicted
answer.

The task is deliberately selective.
For a source containing
\begin{equation}
\{(e_1,a_1),(e_2,a_2),\ldots,(e_N,a_N)\},
\end{equation}
different queries must recover different attributes from the same
preceding field:
\begin{equation}
(Q_i,\Phi_n)\longrightarrow a_i,
\qquad
i=1,\ldots,N.
\label{eq:selective_readout}
\end{equation}
Successful performance therefore cannot be obtained merely by
propagating a generic signal indicating that previous information is
present.
The recurrent field must preserve distinguishable source-dependent
content, while the current query must determine which component is
expressed through the language-model readout.

Training is performed by full backpropagation through the source and
query cycles.
The loss is applied to the answer generated during the QUERY cycle,
without directly supervising how the source information is represented
inside $\Phi_n$.
The internal representation required for subsequent retrieval must
therefore emerge through end-to-end learning.
The experiment thus tests whether an available recurrent pathway can
become an information-bearing write--read channel.
The construction of the synthetic binding
episodes, fixed-length WRITE inputs, and
supervised cross-cycle retrieval protocol
is detailed in Appendix~D.

Figure~\ref{fig:selective_retrieval_learning} shows the resulting
learning dynamics for memory loads
\begin{equation}
N=1,\;2,\;4,\;8,\;16.
\end{equation}
For every tested load, optimization eventually produces high validation
exact-match accuracy, although the learning trajectories differ
substantially with $N$.

The simplest $N=1$ task is acquired rapidly and reaches essentially
perfect retrieval within the first few hundred optimizer steps.
For $N=2$, validation accuracy initially rises to approximately
$0.5$, corresponding to the chance-level selective-retrieval baseline
$1/N$, and remains near this level for an extended interval before
undergoing a sharp transition to nearly perfect retrieval.
Because the source contains two independently addressable bindings,
this intermediate regime is consistent with the presence of source
information without reliable query-dependent selection among the
available bindings.
The subsequent transition from the $1/N$ regime to near-unit accuracy
therefore marks the emergence of reliable content-specific addressing
and retrieval from the recurrent field.

Related transitions occur at larger memory loads.
For $N=8$ and $N=16$, validation accuracy remains low during the early
stage of optimization and then rises abruptly toward unity.
The $N=4$ trajectory exhibits larger fluctuations and settles somewhat
below the other trained conditions, but nevertheless undergoes the same
qualitative transition from weak initial retrieval to a high-accuracy
regime.

The important observation is therefore not that all memory loads follow
the same optimization trajectory.
Rather, across the tested values of $N$, learning transforms the
re-entered field into a representation from which the current query can
select and recover the appropriate source-dependent content.
This distinguishes selective retrieval from simple persistence.
A field may remain dynamically present across a cycle boundary without
preserving multiple distinguishable bindings, and stored information may
remain present without being selectively addressable.
The present experiment requires both properties simultaneously.

The resulting computation can therefore be summarized as
\begin{equation}
\text{source context}
\longrightarrow
\Phi_n
\longrightarrow
(Q_{n+1},\Phi_n)
\longrightarrow
A.
\label{eq:write_reentry_read}
\end{equation}
The language-model output produced during the WRITE cycle does not carry
the source information into the QUERY cycle.
The causal bridge available to the subsequent computation is the
re-entered hidden field itself.

\vspace{5pt}

\paragraph{Causal verification of content-specific retrieval.}

High retrieval accuracy alone does not establish that the recovered
answer is causally determined by information carried by the recurrent
field.
In principle, the model could exploit correlations in the query itself,
or recurrent activation could provide a nonspecific computational
advantage without transmitting episode-specific information.
We therefore intervene directly on the recurrent field while holding
the external QUERY input fixed.

For each held-out query, we evaluate three conditions.
In the correct-field condition ($C$), the model receives the recurrent
field generated from the corresponding source episode.
In the wrong-field condition ($W$), the same query is presented while
the recurrent field is replaced by one generated from an unrelated
episode.
In the state-off condition ($O$), recurrent re-entry is disabled
entirely.
These conditions isolate retrieval with the history-matched field,
retrieval with an informationally incorrect field, and retrieval without
access to the recurrent field, respectively.

Table~\ref{tab:selective_retrieval_intervention} reports the resulting
exact-match accuracies for memory loads from $N=1$ to $N=16$.
With the history-matched field present, exact-match accuracy is
$1.0000$, $0.9975$, $0.9760$, $0.9988$, and $0.9976$ for
$N=1,2,4,8,$ and $16$, respectively.
Thus, throughout the tested range, the model retrieves the queried
binding with near-perfect accuracy when supplied with the field
generated from the corresponding source episode.

\begin{table}[t]
\centering
\caption{
Causal intervention on the recurrent field during selective retrieval.
$C$, $W$, and $O$ denote the correct-field, wrong-field, and
state-off conditions, respectively.
Subscript EM denotes exact-match accuracy.
The differences $C-W$ and $C-O$ quantify the retrieval advantage
of the correct field over the controls.
}
\label{tab:selective_retrieval_intervention}

\begin{tabular*}{\columnwidth}
{@{\extracolsep{\fill}}lccccc@{}}
\hline\hline
Memory load
& $C_{\rm EM}$
& $W_{\rm EM}$
& $O_{\rm EM}$
& $C-W$
& $C-O$ \\
\hline
$N=1$  & 1.0000 & 0.0275 & 0.0350 & 0.9725 & 0.9650 \\
$N=2$  & 0.9975 & 0.0150 & 0.0354 & 0.9825 & 0.9621 \\
$N=4$  & 0.9760 & 0.0131 & 0.0185 & 0.9629 & 0.9575 \\
$N=8$  & 0.9988 & 0.0160 & 0.0167 & 0.9827 & 0.9821 \\
$N=16$ & 0.9976 & 0.0140 & 0.0193 & 0.9835 & 0.9783 \\
\hline\hline
\end{tabular*}
\end{table}

Retrieval collapses when the recurrent field is either replaced or
removed.
Across the same memory loads, wrong-field exact-match accuracy remains
between approximately $1.3\%$ and $2.8\%$, while state-off accuracy
remains between approximately $1.7\%$ and $3.5\%$.
Consequently, the correct-field advantage remains large throughout the
tested range.

The causal structure of the intervention can be summarized as
\begin{equation}
P(A\mid Q,\Phi_C)
\gg
P(A\mid Q,\Phi_W)
\simeq
P(A\mid Q,\Phi_{\rm off}),
\label{eq:causal_memory}
\end{equation}
where $\Phi_C$ denotes the history-matched recurrent field,
$\Phi_W$ an unrelated recurrent field, and $\Phi_{\rm off}$ the
condition in which recurrent re-entry is disabled.

Because the external query $Q$ is held fixed across these conditions,
the large change in retrieval performance is induced by manipulation of
the recurrent field available to the model.
The near-equivalence of the wrong-field and state-off conditions is
particularly informative.
If recurrent activation merely supplied a generic computational
advantage, replacing the correct field by an unrelated field should
preserve a substantial fraction of the retrieval performance.
Instead, an unrelated field provides essentially no retrieval advantage
over disabling re-entry altogether.

Successful retrieval therefore depends not merely on the presence of
recurrent activity, but on the episode-specific information carried by
the appropriate recurrent field.
The same intervention also constrains a query-only interpretation:
the external QUERY is identical across conditions, yet performance
changes from near-perfect retrieval under the correct field to
near-floor performance when that field is replaced or removed.

Taken together, Figs.~\ref{fig:selective_retrieval_protocol} and
\ref{fig:selective_retrieval_learning}, together with
Table~\ref{tab:selective_retrieval_intervention}, establish the
functional emergence of content-specific recurrent memory.
Multiple bindings written during one inference cycle can be retained in
the token-resolved field $\Phi_n$ and selectively accessed by distinct
queries during the subsequent cycle.
Direct intervention further shows that the history-matched field is a
causal variable of this retrieval rather than merely an accompanying
recurrent activation.

Selective retrieval across a single cycle boundary, however, does not
establish long-term persistence.
The next question is whether this content-specific field can remain
functionally available through repeated re-entry and intervening
inference cycles, and whether its persistence scale itself can be
organized by learning.

\begin{figure*}[t]
\centering
\includegraphics[width=1.0\textwidth, trim=0cm 1cm 0cm 0cm]{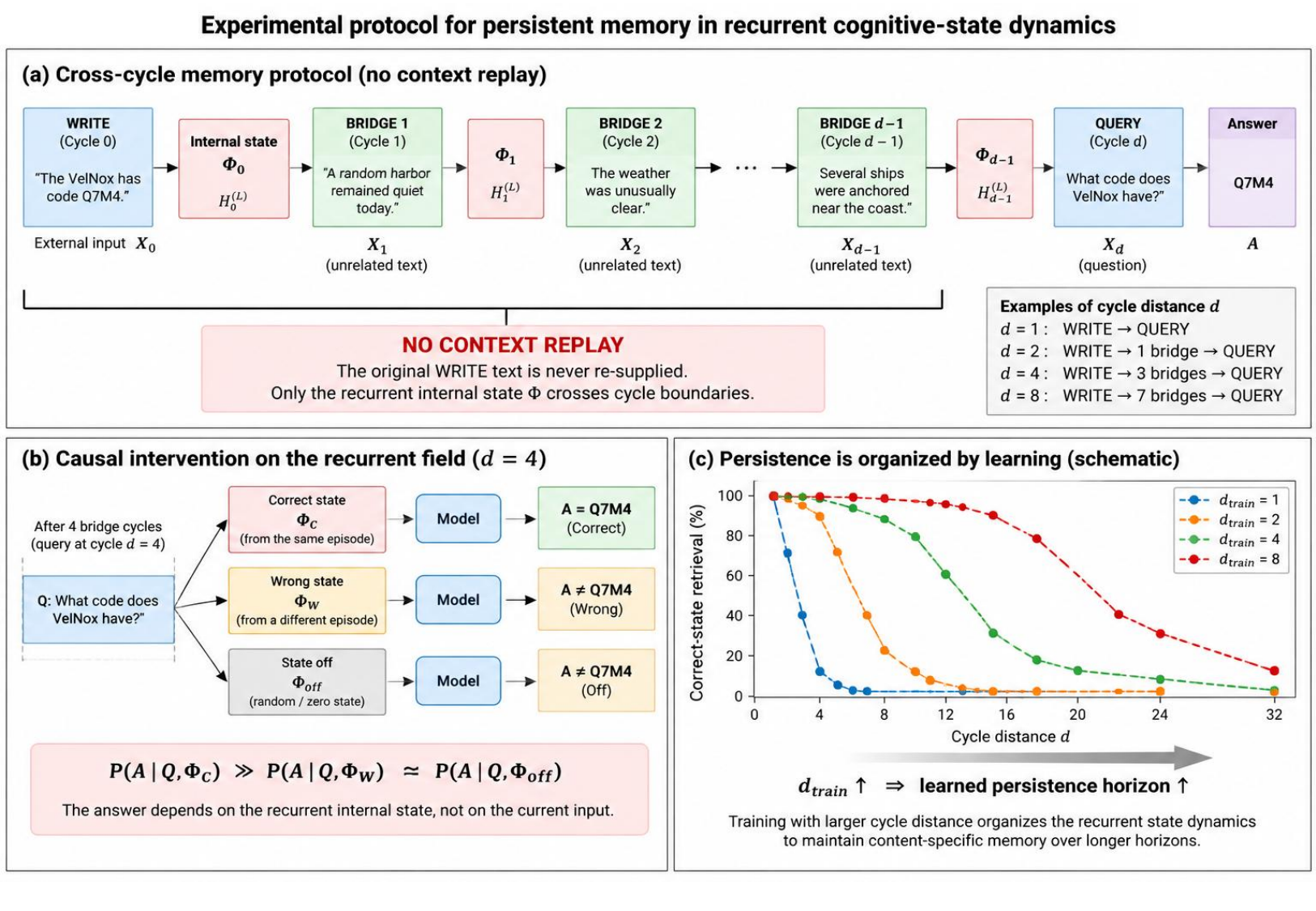}
\caption{
Experimental protocol for persistent memory through
recurrent cognitive-field dynamics.
(a) Cross-cycle memory protocol without context replay.
During the WRITE cycle, the CFN receives an
entity--attribute binding and generates the recurrent
field $\Phi_0$.
The original context is then removed, while the field
propagates through $d-1$ bridge cycles with unrelated
inputs $X_1,\ldots,X_{d-1}$.
At cycle distance $d$, a QUERY retrieves the original
attribute from the propagated field $\Phi_{d-1}$.
(b) Causal intervention on the recurrent field.
For the same query and matched intervening inputs,
retrieval is evaluated using the correct field $\Phi_C$,
an unrelated field $\Phi_W$, or disabled re-entry
$\Phi_{\rm off}$.
Content-specific retrieval is assessed through
$P(A\mid Q,\Phi_C)\gg P(A\mid Q,\Phi_W)
\simeq P(A\mid Q,\Phi_{\rm off})$.
(c) Schematic illustration of learning-dependent memory
persistence for training distances
$d_{\rm train}=1,2,4,$ and $8$.
Larger training distances correspond to longer
illustrated retention horizons.
}
\label{fig:persistent_memory_protocol}
\end{figure*}

\begin{figure}[t]
\centering
\includegraphics[scale=0.58, trim= 0.2cm 1.0cm 0cm 0cm]{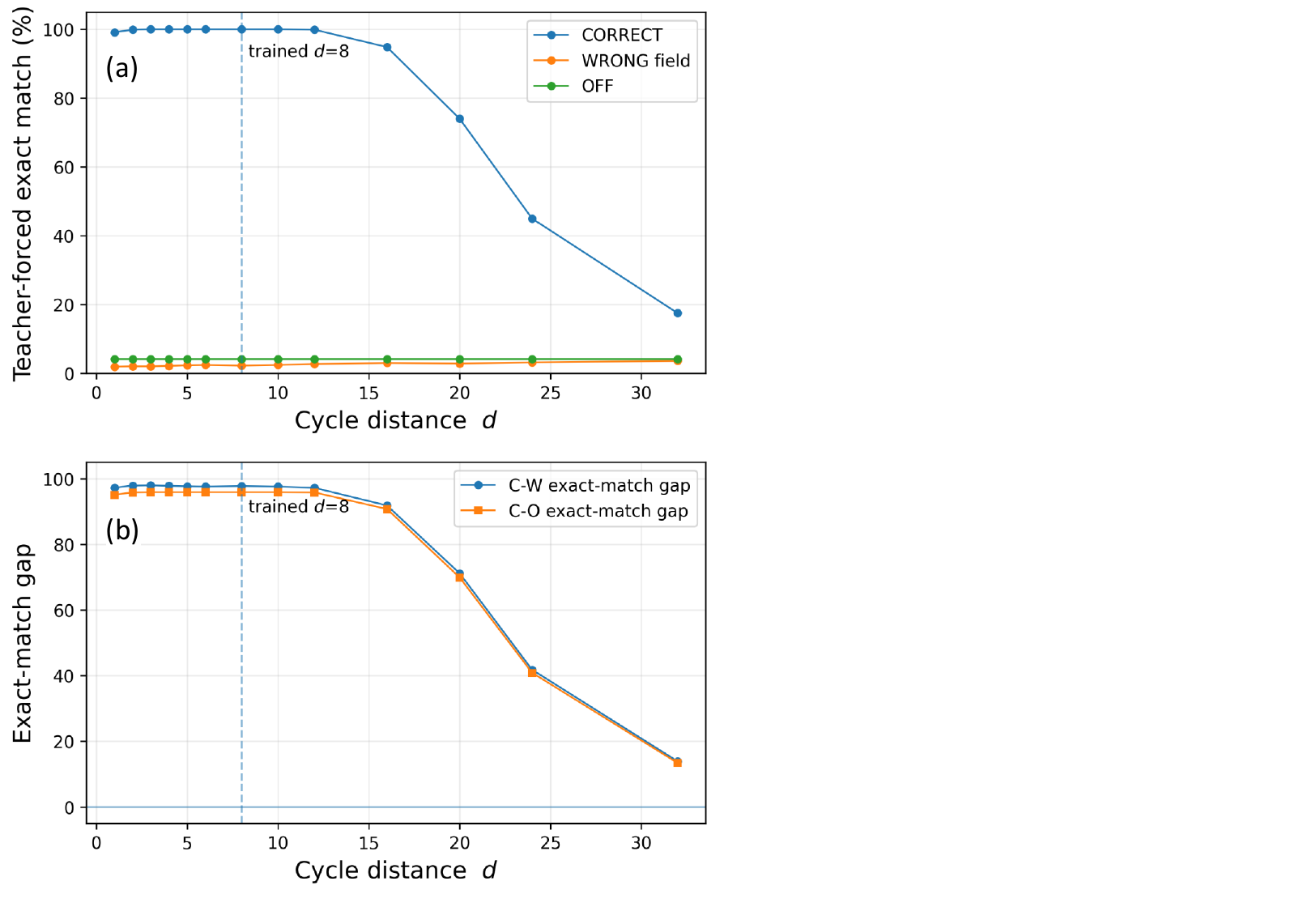}
\caption{
Long-horizon persistence of content-specific information
in the recurrent cognitive field.
Exact-match retrieval for the CFN trained at cycle distance
$d_{\rm train}=8$ and evaluated without further optimization
or replay of the original WRITE context.
The vertical dashed line indicates the training distance.
(a) Teacher-forced exact-match accuracy under the
correct-field ($C$), wrong-field ($W$), and state-off ($O$)
conditions.
Correct-field retrieval remains near perfect beyond the
trained distance before gradually declining, while the
wrong-field and state-off controls remain near background
levels.
(b) Content-specific retrieval signals measured by the
exact-match differences $C-W$ and $C-O$.
Both differences remain large beyond the trained distance
and decrease at longer cycle distances.
}
\label{fig:long_horizon_retention}
\end{figure}

\begin{figure}[t]
\centering
\includegraphics[scale=0.56, trim= 0.3cm 0.5cm 0cm 0cm]{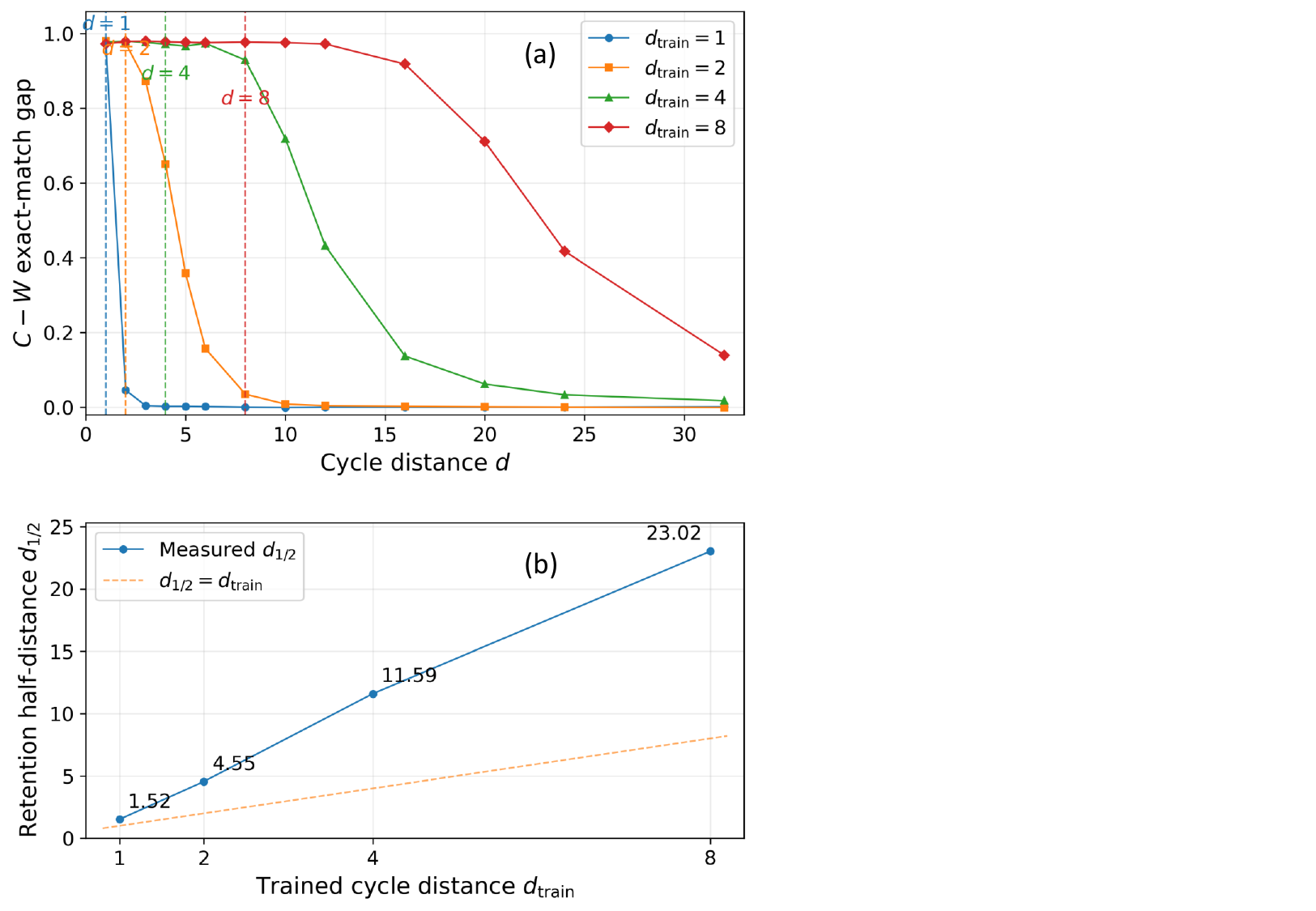}
\caption{
Learning-dependent temporal persistence of the recurrent
cognitive field.
(a) Content-specific retention measured by the
correct--wrong exact-match difference
$\mathcal M(d)$ for training distances
$d_{\rm train}=1,2,4,$ and $8$.
The vertical dashed lines indicate the corresponding
training distances.
The retention curves shift toward longer cycle distances
as $d_{\rm train}$ increases.
(b) Retention half-distance $d_{1/2}$, defined by
$\mathcal M(d_{1/2})=\mathcal M_0/2$, as a function of
$d_{\rm train}$.
The measured values are
$d_{1/2}\simeq1.52,\,4.55,\,11.59,$ and $23.02$
for $d_{\rm train}=1,2,4,$ and $8$, respectively.
The dashed line indicates the reference
$d_{1/2}=d_{\rm train}$.
}
\label{fig:learned_persistence_scale}
\end{figure}

\subsection{D. Persistent memory through recurrent cognitive-field dynamics}
\label{sec:persistent_memory}

The preceding experiments establish that the recurrent field can carry
content-specific information and that subsequent inference depends
causally on the history-matched field.
A defining requirement for persistent cognition, however, is stronger:
information acquired during one processing cycle must remain
functionally available over extended recurrent trajectories even when
the original external input is no longer present.
We therefore next ask whether the temporal persistence of the
content-bearing cognitive field can itself be organized by learning.

Persistence was trained using a progressive cycle-distance curriculum,
$d_{\rm train}=1,2,4,$ and $8$.
Each stage was initialized from the best validation checkpoint of the
preceding distance, while fresh training episodes and randomized
intervening bridge sequences were generated at each stage.
For a given $d_{\rm train}$, the recurrent field was propagated through
$d_{\rm train}-1$ bridge cycles, supervision was applied only at the
final QUERY, and gradients were propagated through the full recurrent
trajectory by backpropagation through time.
Thus, temporal training progressively extended an already learned
recurrent-memory dynamics rather than training each persistence
distance independently from scratch.
The construction of randomized intervening inputs, full-BPTT persistence training, 
and long-horizon evaluation procedures is detailed in Appendix~D.

Figure~\ref{fig:persistent_memory_protocol} summarizes the experimental
protocol.
During the initial WRITE cycle, the model receives a randomly generated
entity--attribute binding and forms the recurrent field
\begin{equation}
\Phi_0=H_0^{(L)}.
\end{equation}
The original WRITE context is then removed and is never presented again.
During each subsequent cycle, the model receives an unrelated randomized
bridge sequence while only the internally generated field is propagated
across the cycle boundary,
\begin{equation}
(X_k,\Phi_{k-1})
\longrightarrow
\Phi_k.
\label{eq:persistent_field_update}
\end{equation}
After $d-1$ intervening bridge cycles, a QUERY asks for the attribute
associated with the original entity,
\begin{equation}
(Q_d,\Phi_{d-1})
\longrightarrow
A.
\label{eq:persistent_query}
\end{equation}
The cycle distance $d$ therefore denotes the separation between the
original WRITE event and the final QUERY, with $d-1$ intervening
recurrent transformations.

No replay of the original context, external memory buffer, or
re-supply of the WRITE sequence is used during this interval.
The remembered information must instead remain functionally represented
along the evolving recurrent trajectory
\begin{equation}
\Phi_0
\longrightarrow
\Phi_1
\longrightarrow
\cdots
\longrightarrow
\Phi_{d-1},
\label{eq:field_trajectory}
\end{equation}
while unrelated external inputs continuously perturb the system.
Persistent memory is therefore tested here as a dynamical property of
the evolving cognitive field rather than as continued access to a
stored external context.

As in the selective-retrieval experiment, the source of successful
retrieval is verified by direct intervention on the recurrent field.
For the same QUERY and the same intervening bridge inputs, we compare
three conditions: a correct field $\Phi_C$ propagated from the
corresponding WRITE episode, a wrong field $\Phi_W$ propagated from an
unrelated episode, and an OFF condition in which recurrent re-entry is
removed.
The causal criterion is therefore
\[
P(A\mid Q,\Phi_C)
\gg
P(A\mid Q,\Phi_W)
\simeq
P(A\mid Q,\Phi_{\rm off}).
\]
Because the correct and wrong fields traverse the same intervening
inputs, a selective advantage of $\Phi_C$ identifies information carried
specifically by the recurrent field rather than by the bridge sequences
themselves.

Figure~\ref{fig:long_horizon_retention} shows the resulting long-horizon
behavior for the checkpoint obtained at
$d_{\rm train}=8$.
Correct-field retrieval remains essentially perfect throughout the
trained interval and substantially beyond it.
Exact-match accuracy remains $100\%$ at $d=8$ and $d=10$ and
approximately $99.9\%$ at $d=12$.
Beyond this extended plateau, retrieval decays gradually, reaching
approximately $95\%$ at $d=16$, $74\%$ at $d=20$, $45\%$ at $d=24$,
and $18\%$ at $d=32$.
The wrong-field and OFF controls, by contrast, remain near only a few
percent throughout the same interval.

The lower panel of Fig.~\ref{fig:long_horizon_retention} isolates the
content-specific contribution by plotting the correct--wrong and
correct--OFF exact-match differences.
Both remain large well beyond the trained distance and decay together
with the absolute correct-field retrieval.
The long-horizon signal therefore cannot be explained by an increasing
background probability of producing the correct answer.
Instead, information specific to the original WRITE episode remains
behaviorally accessible through repeated transformations of the
recurrent field before gradually relaxing at longer cycle distances.

The finite decay is itself informative.
Persistent memory in the CFN does not require a perfectly conserved or
non-decaying internal state.
Rather, the field remains content bearing over a finite dynamical
timescale while continuing to undergo recurrent transformation.
This behavior is qualitatively consistent with the finite-gap
near-critical dynamics discussed in Sec.~III.B:
long-lived cognitive organization can coexist with finite forgetting
rather than requiring exact dynamical freezing.

\vspace{5pt}

\paragraph{Persistence is a learnable dynamical scale.}

We next examine how the behavioral persistence scale changes across the
successive stages of the temporal curriculum,
$d_{\rm train}=1,2,4,$ and $8$.
Each trained checkpoint was evaluated at cycle distances extending well
beyond the distance used at that curriculum stage.
No additional optimization, gradient update, or replay of the original
WRITE context was performed during this long-distance evaluation.

To isolate information carried specifically by the recurrent field, we
define the content-specific memory signal
\begin{equation}
\mathcal M(d)
\equiv
P(A\mid C,d)
-
P(A\mid W,d),
\label{eq:retention_signal}
\end{equation}
where $C$ and $W$ denote the correct-field and wrong-field conditions,
respectively, and $P$ denotes the measured teacher-forced exact-match
retrieval rate.
Thus, $\mathcal M(d)$ measures the retrieval advantage attributable
specifically to the episode-dependent content carried by the propagated
recurrent field.

Figure~\ref{fig:learned_persistence_scale}(a) reveals a systematic
displacement of the content-specific retention profile across the
temporal curriculum.
At $d_{\rm train}=1$, most of the signal is lost immediately beyond the
trained distance.
The $d_{\rm train}=2$ stage shifts the decay toward longer recurrent
trajectories, while the $d_{\rm train}=4$ checkpoint maintains a strong
content-specific signal over a substantially broader range.
At $d_{\rm train}=8$, the signal remains clearly detectable even at
$d=32$.
The learned memory therefore does not terminate at the largest temporal
separation explicitly encountered during a given training stage.

To characterize this displacement without assuming a particular decay
law, we define the empirical retention half-distance $d_{1/2}$ by
\begin{equation}
\mathcal M(d_{1/2})
=
\frac{1}{2}\mathcal M_0,
\label{eq:dhalf_definition}
\end{equation}
where $\mathcal M_0$ denotes the initial content-specific memory signal.
The crossing is estimated directly from the measured retention curve.

As shown in Fig.~\ref{fig:learned_persistence_scale}(b), the measured
half-distances are approximately
$d_{1/2}=1.52,\,4.55,\,11.59,$ and $23.02$ cycles for
$d_{\rm train}=1,\,2,\,4,$ and $8$, respectively.
Except for the shortest-distance condition, the behavioral persistence
scale therefore extends substantially beyond the corresponding training
distance.
Progressively increasing $d_{\rm train}$ does not merely improve
retrieval at the endpoint used for supervision; it systematically
extends the temporal stability of the content-bearing recurrent field.

At a coarse-grained level, a field component dominated by an effective
slow relaxation scale may be represented schematically as
\begin{equation}
\mathfrak m(d)\sim e^{-r_{\rm eff}d},
\qquad
d_{\rm memory}\sim r_{\rm eff}^{-1}.
\label{eq:effective_memory_decay}
\end{equation}
The increase of $d_{1/2}$ across the temporal curriculum is therefore
consistent with a decreasing effective relaxation scale.
This interpretation is not a direct measurement of either the
microscopic relaxation spectrum or the cognitive forgetting gap
$r_{\rm cog}$, and the measured curves do not establish a particular
asymptotic decay law.
The empirical half-distance is used only as a model-independent
behavioral measure of persistence.

Together with the correct-field, wrong-field, and recurrence-off
controls above, these results establish that persistent memory in the
CFN is a learned dynamical property of the recurrent cognitive field.
The field carries episode-specific information, remains causally
necessary for later retrieval, survives repeated transformations by
unrelated inputs, and acquires a characteristic persistence scale that
increases as the temporal curriculum is extended.

The next section asks a distinct question: how this learned persistence
is reorganized when the intervening trajectory is not unrelated, but
remains semantically connected to the remembered episode.

\begin{figure}[t]
\centering
\includegraphics[scale=0.5, trim= 0.2cm 8.5cm 0cm 0cm]{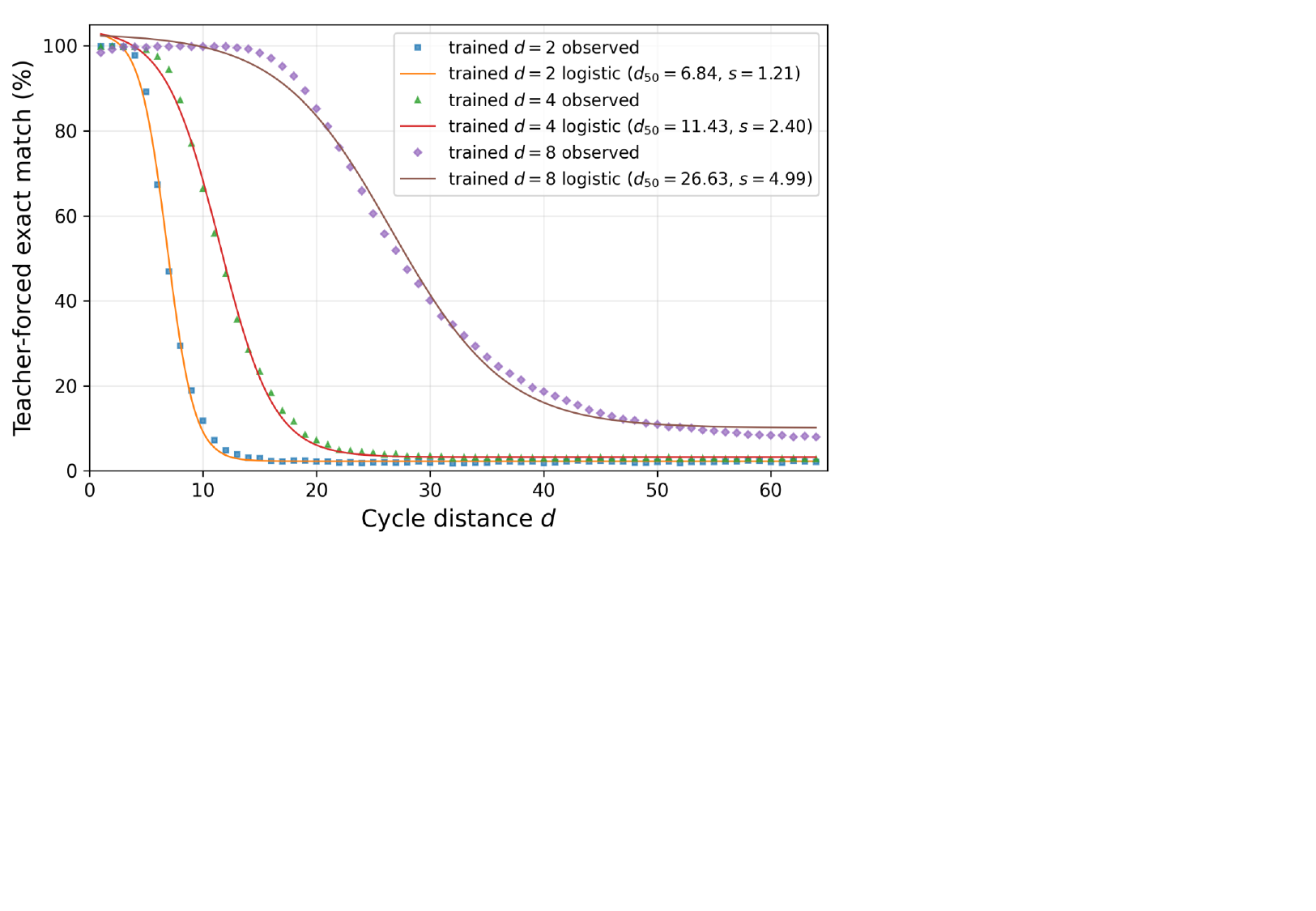}
\caption{
Learning-dependent displacement of long-horizon memory
retention under semantic continuation.
Teacher-forced exact-match accuracy as a function of
cycle distance $d$ for CFN models trained at
$d_{\rm train}=2,4,$ and $8$ using
semantic-continuation bridges.
Symbols represent measured retention values, and solid
curves denote four-parameter logistic fits.
The fitted midpoints are $d_{50}=6.84$, $11.43$,
and $26.63$ for $d_{\rm train}=2,4,$ and $8$,
respectively.
The retention curves shift toward longer cycle distances
as the training distance increases.
}
\label{fig:semantic_retention_logistic}
\end{figure}

\begin{figure}[t]
\centering
\includegraphics[scale=0.5, trim= 0.2cm 8.5cm 0cm 0cm]{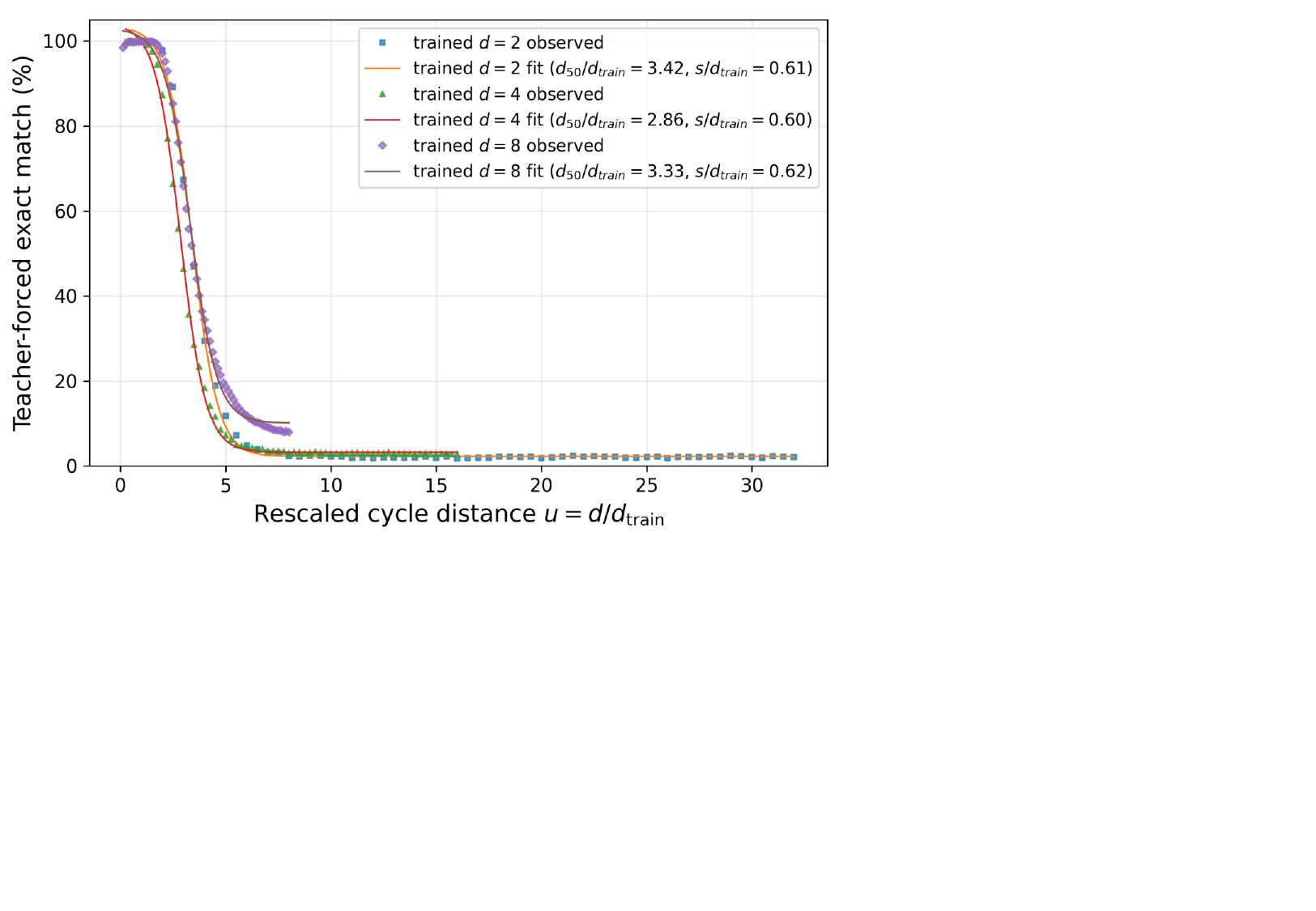}
\caption{
Approximate temporal scaling collapse of learned
recurrent-memory retention under semantic continuation.
Normalized correct-field retention,
$\widetilde P=(P-P_{\infty})/(P_0-P_{\infty})$,
is plotted against the rescaled cycle distance
$u=d/d_{\rm train}$ for CFN models trained at
$d_{\rm train}=2,4,$ and $8$.
Symbols represent measured teacher-forced exact-match
retention, and solid curves denote the corresponding
logistic fits.
The three retention profiles exhibit an approximate
collapse in the transition region, with modest differences
in the rescaled midpoints $d_{50}/d_{\rm train}$.
}
\label{fig:semantic_scaling_collapse}
\end{figure}

\section{IV. Persistent and Adaptive Cognitive Dynamics}
\label{sec:persistent_adaptive_dynamics}

Section~III established that the recurrent cognitive field can carry
content-specific information across multiple inference cycles and that
its persistence timescale can itself be reorganized by learning.
Those experiments used randomly generated neutral bridges between the
source and query, providing a stringent test of retention under
intervening inputs unrelated to the original episode.

Natural cognitive trajectories, however, rarely evolve through entirely
unrelated inputs.
Subsequent observations often remain semantically connected to an
ongoing episode even when the specific information required for later
recall is not repeated.
We therefore extend the analysis to semantic-continuation bridges that
remain related to the entities, relations, or context of the source
episode while excluding the target answer values.
This allows us to test whether the evolving recurrent field is supported
by the semantic structure of the subsequent trajectory without direct
replay of the remembered information.

This regime is distinct from the explicit source re-exposure examined
later in this section.
Semantic continuation provides related contextual input while requiring
the existing recurrent field to preserve the target information,
whereas source re-exposure deliberately reintroduces the original source
and tests whether a partially decayed field can be renewed.
The two experiments therefore probe complementary aspects of adaptive
cognitive dynamics: context-supported persistence and input-driven
renewal.

We first examine how semantic continuation modifies long-horizon
retention and whether the learned persistence scale continues to expand
with temporal training.
We then ask whether subsequent content-matched input can renew a
decaying recurrent field and sustain a nonzero memory state over
extended recurrent trajectories.
Together, these experiments move beyond passive retention toward a
dynamical picture in which cognitive persistence depends jointly on the
learned recurrent field and its continuing interaction with relevant
experience.

\subsection{A. Learned temporal scaling under semantic continuation}
\label{sec:learned_temporal_scaling}

Section~III established, using random intervening inputs and
field-specific controls, that the CFN learns a finite but adaptable
persistence scale for episode-dependent recurrent memory.
We now ask how this learned persistence is organized when the
intervening trajectory remains semantically related to the source
episode.

To isolate temporal organization from changes in memory load, all
experiments in this section use the $N=2$ selective-binding task
introduced in Sec.~III.
Otherwise identical CFN models were trained using semantic-continuation
bridges at cycle distances
$d_{\rm train}=2,4,$ and $8$ and subsequently evaluated over the common
long-horizon range $1\le d\le64$.
No additional optimization was performed during long-horizon
evaluation, and the target answer values were excluded from all
intervening bridges.
The construction of the semantic-continuation
bridges, including the alternating contextual
templates and entity orderings, is described
in Appendix~D.

Because Sec.~III already established through wrong-field controls that
retrieval depends on the episode-specific content of the propagated
field, here we directly analyze the correct-field teacher-forced
exact-match retention.
For compactness, we denote this measured retention rate by
\begin{equation}
P(d)
\equiv
P(A\mid C,d),
\label{eq:correct_field_retention}
\end{equation}
where $C$ denotes the correct-field condition and $P$ denotes the
teacher-forced exact-match retrieval rate, as in Sec.~III.

Figure~\ref{fig:semantic_retention_logistic} shows the resulting
long-horizon retention curves.
Increasing the training distance shifts the entire forgetting
transition toward longer recurrent trajectories.
The $d_{\rm train}=2$ model loses most of its retrievable content within
approximately ten cycles, whereas the $d_{\rm train}=4$ model retains
substantial accuracy over a broader range.
For $d_{\rm train}=8$, retrieval remains close to its short-distance
plateau far beyond the distance encountered during training before
entering a broad long-distance decay.
Semantic continuation therefore reveals a systematic organization of
the entire temporal retention profile.

To quantify this organization, each retention curve was fit to the
four-parameter decreasing logistic form
\begin{equation}
P(d)
=
P_{\infty}
+
\frac{P_0-P_{\infty}}
{1+\exp[(d-d_{50})/s]},
\label{eq:retention_logistic}
\end{equation}
where $P_0$ and $P_{\infty}$ denote the fitted short- and long-distance
levels, $d_{50}$ is the midpoint of the fitted dynamical range, and $s$
characterizes the width of the forgetting transition.
The corresponding $90\%$--$10\%$ transition width is
\begin{equation}
\Delta d_{90\rightarrow10}
=
2s\ln9.
\end{equation}
The logistic form is used only as a phenomenological parameterization
of behavioral retention and is not assumed to describe the microscopic
relaxation law of the recurrent field.

The fitted parameters are summarized in
Table~\ref{tab:retention_logistic}.
All three retention profiles are accurately described by the logistic
form, with $R^2=0.9985$, $0.9987$, and $0.9968$ for
$d_{\rm train}=2,4,$ and $8$, respectively.

\begin{table}[t]
\centering
\caption{
Logistic characterization of long-horizon correct-field retention
under semantic continuation.
The midpoint $d_{50}$ and transition scale $s$ are obtained from
Eq.~(\ref{eq:retention_logistic}).
The transition width is
$\Delta d_{90\rightarrow10}=2s\ln9$.
The last two columns show the fitted midpoint and transition scale
normalized by the training distance.
}
\label{tab:retention_logistic}
\begin{tabular*}{\columnwidth}
{@{\extracolsep{\fill}}cccccc@{}}
\hline\hline
$d_{\rm train}$
& $d_{50}$
& $s$
& $\Delta d_{90\rightarrow10}$
& $d_{50}/d_{\rm train}$
& $s/d_{\rm train}$ \\
\hline
2 & 6.84  & 1.21 & 5.32  & 3.42 & 0.605 \\
4 & 11.43 & 2.40 & 10.54 & 2.86 & 0.600 \\
8 & 26.63 & 4.99 & 21.95 & 3.33 & 0.624 \\
\hline\hline
\end{tabular*}
\end{table}

Two related scaling properties emerge.
First, the fitted midpoint shifts from $d_{50}=6.84$ to $11.43$ and
$26.63$ as $d_{\rm train}$ increases from $2$ to $4$ and $8$.
Second, the transition scale grows from $s=1.21$ to $2.40$ and $4.99$,
giving
\begin{equation}
\frac{s}{d_{\rm train}}
=
0.605,\;0.600,\;0.624,
\end{equation}
or, over the present range,
\(
s\simeq0.61\,d_{\rm train}.
\)
Temporal learning therefore rescales not only where retrieval is lost
but also the width of the transition through which it is lost.

The sigmoidal profile itself admits a natural behavioral interpretation.
Let $\Delta z(d)$ denote an effective retrieval margin between the
target representation and competing outputs.
For a softmax-like readout, a two-way competition takes the form
\begin{equation}
p_{\rm ret}(d)
=
\frac{1}{1+\exp[-\Delta z(d)]},
\label{eq:retrieval_margin}
\end{equation}
where $p_{\rm ret}$ denotes an effective readout probability rather
than the measured exact-match rate $P(d)$.
Near the retrieval crossover, a local expansion
\begin{equation}
\Delta z(d)
\simeq
-\kappa(d-d_{50})
\end{equation}
produces a logistic transition with
\(
s\simeq\kappa^{-1}.
\)
More generally, a smooth threshold-like readout of a continuously
varying internal state can generate a similar sigmoid.
Because $P(d)$ is an empirical exact-match rate rather than the
softmax probability itself, this construction should be understood as
an effective readout interpretation, not as a microscopic derivation
of the measured retention law.

This distinction separates the shape of the behavioral transition from
its learned temporal scale.
A fixed nonlinear readout can naturally generate a sigmoid, but cannot
by itself explain why the transition broadens systematically with
$d_{\rm train}$.
The observed near-constancy of $s/d_{\rm train}$ instead implies
\begin{equation}
\kappa
\sim
\frac{1}{s}
\propto
\frac{1}{d_{\rm train}},
\label{eq:margin_temporal_scaling}
\end{equation}
indicating that temporal training reorganizes the recurrent dynamics
that supplies the memory-bearing signal to the readout.

This interpretation leads to a direct scaling test.
We define the rescaled cycle distance and normalized retention as
\begin{equation}
u
=
\frac{d}{d_{\rm train}},
\qquad
\widetilde P
=
\frac{P-P_{\infty}}
{P_0-P_{\infty}}.
\label{eq:normalized_temporal_variables}
\end{equation}
If $d_{\rm train}$ organizes the characteristic temporal scale of the
underlying recurrent dynamics, retention curves obtained at different
training distances should approximately collapse when expressed in
terms of these dimensionless variables.

Figure~\ref{fig:semantic_scaling_collapse} shows this behavior.
Despite the factor-of-four variation in training distance, the three
curves collapse onto a narrow common transition region when plotted
against $d/d_{\rm train}$.
The remaining horizontal displacement is consistent with the modest
variation of $d_{50}/d_{\rm train}$ across the three models.
The retention profiles are therefore approximately related by a common
temporal rescaling rather than representing independent
distance-specific solutions.

Phenomenologically, this organization can be summarized as
\begin{equation}
\begin{aligned}
P(d;d_{\rm train})
\simeq
&
P_{\infty}(d_{\rm train})
\\
&+
\left[
P_0(d_{\rm train})
-
P_{\infty}(d_{\rm train})
\right]
\mathcal F
\!\left(
\frac{d}{d_{\rm train}}
\right),
\label{eq:retention_scaling_form}
\end{aligned}
\end{equation}
with, over the present range,
\begin{equation}
\frac{d_{50}}{d_{\rm train}}\sim3.2,
\qquad
\frac{s}{d_{\rm train}}\sim0.61.
\end{equation}
These numerical values should not be interpreted as universal
constants.
Only three training distances are examined, and the fitted
long-distance floor, particularly for $d_{\rm train}=8$, remains
sensitive to the finite observation window.
The central result is the approximate one-parameter temporal collapse.

The random-bridge and semantic-continuation experiments therefore probe
complementary aspects of the same learned recurrent memory.
Section~III established that episode-specific information persists
causally in the recurrent field and that its behavioral persistence
scale is adaptable.
The present results further show that, under semantically structured
continuation, the entire retention profile exhibits an approximate
temporal rescaling with the distance imposed during learning.

Persistence in the CFN is therefore neither a fixed architectural
lifetime nor simply passive survival of an unchanged representation.
It is a learned dynamical property whose observable temporal structure
depends on the continuing interaction between the recurrent field and
subsequent input.

This observation motivates the next question.
If semantically related input can support an existing recurrent field
without replaying its target content, can direct re-exposure to the
source actively renew a field that has already begun to decay?
We address this input-driven renewal regime in the following subsection.

\begin{figure}[t]
\centering
\includegraphics[scale=0.5, trim= 0.8cm 8.1cm 0cm 0cm]{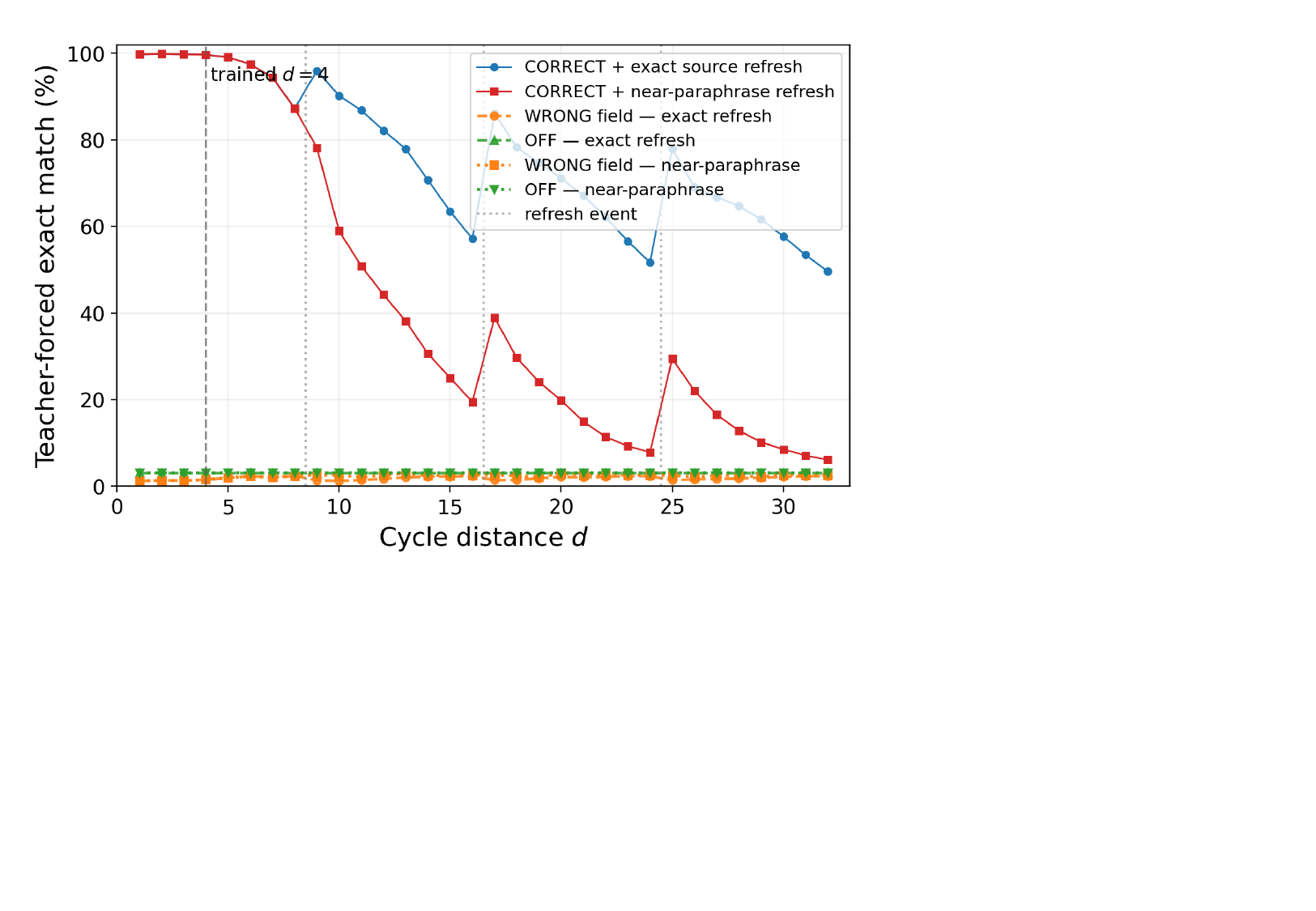}
\caption{
Representation-sensitive renewal of recurrent-memory
retention under periodic source re-exposure.
Teacher-forced exact-match accuracy as a function of
cycle distance $d$ for a CFN trained with semantic
continuation at $d_{\rm train}=4$.
Periodic re-exposure presents either the exact original
source or a near-paraphrased version of the same episode,
without supplying the target answer through intervening
bridges.
Exact source re-exposure produces repeated recovery of
correct-field retrieval, whereas near-paraphrased
re-exposure yields substantially weaker recovery.
Wrong-field and recurrent-field-off controls remain
near background accuracy.
}
\label{fig:representation_sensitive_renewal}
\end{figure}

\begin{figure}[t]
\centering
\includegraphics[scale=0.5, trim= 0.3cm 8.2cm 0cm 0cm]{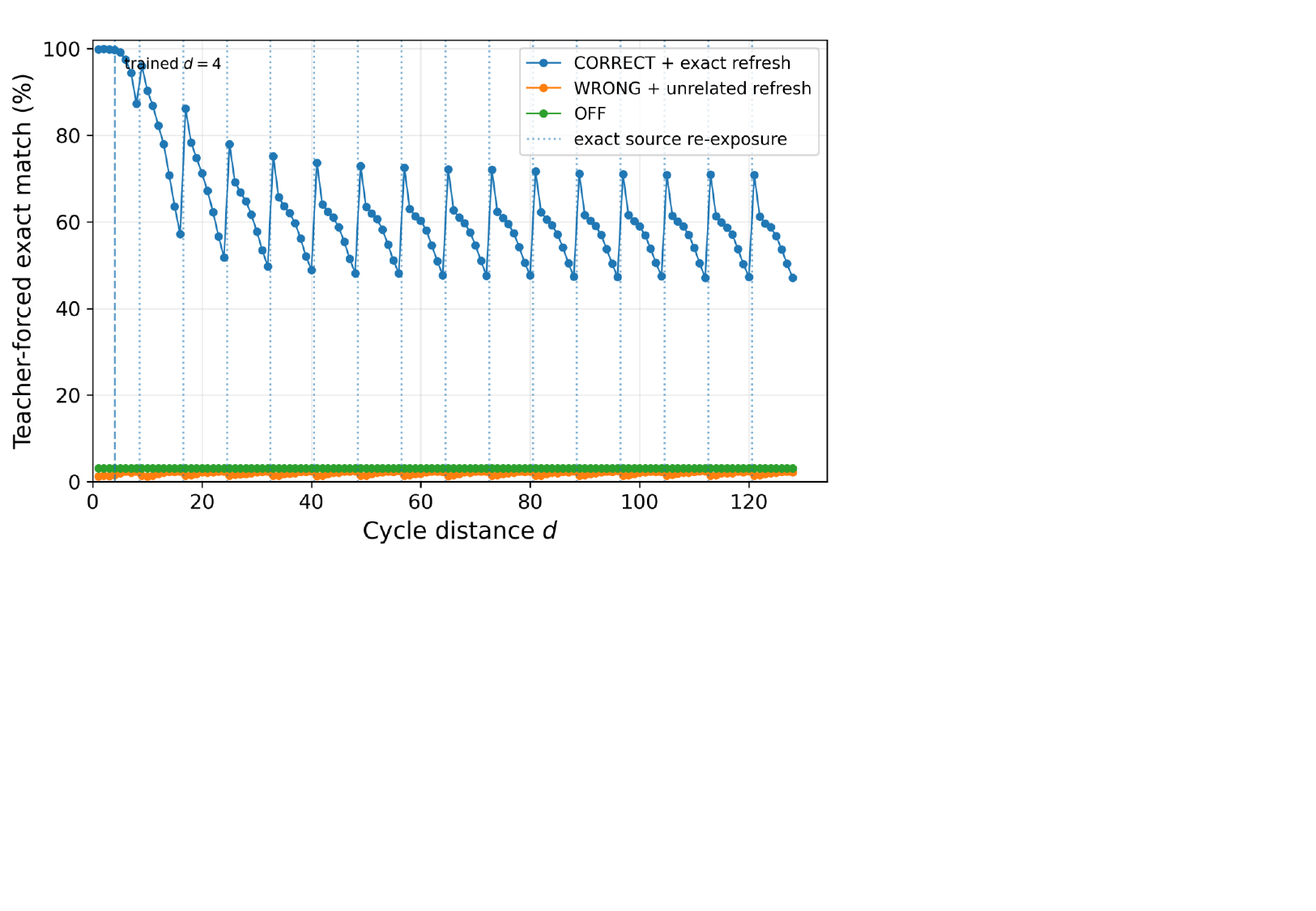}
\caption{
Long-horizon recurrent-memory retention under periodic
content-matched source re-exposure.
Teacher-forced exact-match accuracy for a CFN trained at
$d_{\rm train}=4$, with the original source re-presented
every eight recurrent cycles.
The vertical dotted lines indicate source re-exposure
events.
Correct-field retrieval exhibits repeated recovery
followed by gradual decay between successive
re-exposure events, producing a sawtooth trajectory.
After an initial transient, the peak--trough envelope
approaches an approximately stationary nonzero regime
over the 128-cycle evaluation window.
Wrong-field and recurrent-field-off controls remain
near background accuracy.
}
\label{fig:periodic_renewal}
\end{figure}

\subsection{B. Field re-entry, driven renewal, and persistent cognitive dynamics}
\label{sec:CFN_CFT_realization}

The long-horizon recurrent experiments reveal two complementary
properties of the learned CFN dynamics.
First, the recurrent cognitive field exhibits finite but learnable
persistence when content-relevant input is removed.
Second, subsequent input can selectively reorganize and renew a
partially surviving field.
Together, these observations provide a computational realization of
the driven recurrent inference dynamics developed in Sec.~II.C.

The essential correspondence is not that the CFN stores a static
representation of previous input.
Rather, newly presented information acts on a collective state
generated by preceding inference, while that state remains dynamically
available through cross-cycle field re-entry.
At the computational level,
\[
\Phi_{t+1}
=
F_\theta
\left(
X_{t+1},
\Phi_t
\right),
\]
so that current inference depends jointly on newly presented
information and on the internally generated field inherited from the
preceding cycle.

As derived in Sec.~II.C, the corresponding continuous-time
field-theoretic dynamics may be written in coarse-grained form as
\begin{equation}
\partial_t\phi(t)
=
-r\phi(t)
+
\int_{t_0}^{t}dt'\,
K(t-t')\phi(t')
+
I_{\rm eff}(t)
+
\xi_{\rm eff}(t),
\label{eq:CFN_CFT_memory_field}
\end{equation}
where the memory kernel describes endogenous collective feedback,
whereas $I_{\rm eff}(t)$ describes the effective finite cognitive
drive generated after incoming information has coupled to the
collective dynamical manifold.
The microscopic origin of this drive is the mode-selective projection
\begin{equation}
b_\alpha(t)
=
\widetilde u_\alpha^\dagger
B_\mathcal{X} I(t),
\label{eq:CFN_CFT_modal_projection}
\end{equation}
derived in Sec.~II.C.
Thus, new information does not act on an empty state.
It perturbs a memory-dressed collective field whose present
configuration already contains the dynamical consequences of
preceding inference.

At a coarse-grained discrete level, the CFN dynamics may therefore be
represented schematically as
\begin{equation}
\Phi_{t+1}
=
\mathcal D_t[\Phi_t]
+
\mathcal R_t[X_{t+1},\Phi_t],
\label{eq:CFN_discrete_field}
\end{equation}
where $\mathcal D_t$ denotes propagation and relaxation of the
existing recurrent field and $\mathcal R_t$ denotes its
input-dependent reorganization.
Equation~(\ref{eq:CFN_discrete_field}) should not be interpreted as an
independent phenomenological memory model.
It summarizes the two dynamical components directly realized by the
CFN: persistence of the preceding field and its reorganization by
subsequent input.

This distinction is directly visible in the passive-retention
experiments.
When subsequent input provides little support for the represented
content, the recurrent state evolves predominantly through its
internally generated relaxation dynamics.
Within Cognitive Field Theory, the low-frequency stability of the
memory-dressed field is characterized by the cognitive forgetting gap,
\(
r_{\rm cog}
=
r-\Sigma_R(0).
\)
The corresponding dressed cognitive propagator is
\[
L_{\rm cog}(\Omega)
=
\frac{1}
{-i\Omega+r-\Sigma_R(\Omega)}.
\]
In a stationary linearized regime,
$L_{\rm cog}(\Omega)$ coincides with the retarded susceptibility
defined with respect to an infinitesimal auxiliary probe,
\(
\chi_R(\Omega)
=
L_{\rm cog}(\Omega).
\)
This response probe should be distinguished from the finite cognitive
input that drives inference.

Within a local low-frequency approximation in which the residual
frequency dependence of the self-energy is neglected, the passive
persistence scale behaves as
\begin{equation}
\tau_{\rm cog}
\sim
\frac{1}{r_{\rm cog}}.
\label{eq:CFN_tau_cog}
\end{equation}
The learning-dependent extension of the behavioral retention horizon
observed in Secs.~III and IV.A is therefore consistent with a
reorganization of the effective temporal structure of the recurrent
field.
The present behavioral measurements do not, however, directly measure
$r_{\rm cog}$.
A quantitative identification would require the collective spectrum,
self-energy, and behavioral retention scale to be measured jointly in
the same trained models.

The source-re-exposure experiments probe a complementary property of
the dynamics.
Rather than measuring only how long an internally generated state
survives in the absence of relevant support, they test how subsequent
content-specific information acts on a partially surviving recurrent
field.
For a re-exposed source $I_{\rm re}(t)$, the field-theoretic
mode-selective drive takes the form
\begin{equation}
b_\alpha^{\rm re}(t)
=
\widetilde u_\alpha^\dagger
B_\mathcal{X} I_{\rm re}(t).
\label{eq:CFN_reexposure_modal_projection}
\end{equation}
Different incoming representations therefore need not perturb the
existing cognitive state in the same way.
Their effects depend on how strongly they couple to the collective
directions available in the learned dynamical geometry.

Because the recurrent field has not generally returned to an empty
state before re-exposure, this new mode-selective perturbation acts on
a state that already contains residual collective organization from
preceding inference.
The observed recovery should therefore not be interpreted as a simple
reset of the recurrent state or as an extension of the original
context.
At the field-theoretic level, the process is represented
schematically by
\begin{equation}
I_{\rm re}
\longrightarrow
\{b_\alpha^{\rm re}\}
\longrightarrow
\delta\{\mathcal{X}_\alpha\}
\longrightarrow
\delta\phi
\longrightarrow
\delta y,
\label{eq:CFN_reexposure_chain}
\end{equation}
whereas at the CFN level the same process is realized through the
input-dependent recurrent map,
\(
\Phi_{t+1}
=
F_\theta
\left(
X_{t+1},
\Phi_t
\right).
\)
The incoming representation therefore reorganizes a cognitive field
that is already dynamically conditioned by its own preceding history.

At sufficiently low frequencies, this driven reorganization can be
represented schematically as
\begin{equation}
\partial_t\phi(t)
\simeq
-r_{\rm cog}\phi(t)
+
I_{\rm rel}(t),
\label{eq:CFN_driven_lowfreq}
\end{equation}
where $I_{\rm rel}(t)$ denotes the net low-frequency effective drive
generated after the incoming representation has coupled to the
relevant memory-bearing collective sector.
Equation~(\ref{eq:CFN_driven_lowfreq}) is therefore a coarse-grained
limit of the mode-resolved driven dynamics derived in Sec.~II.C,
rather than an independent assumption about cognitive input.
The low-frequency reduction and the distinction
between passive relaxation and input-driven
persistence are derived in Appendix~C, together
with the stroboscopic description of periodic
source re-exposure.

When content-specific support is removed, the same local
low-frequency approximation gives
\begin{equation}
\phi(t)
\sim
e^{-r_{\rm cog}t},
\label{eq:CFN_passive_field_decay}
\end{equation}
so that $r_{\rm cog}>0$ corresponds to a finite passive retention
scale.
This approximation describes the effective slow relaxation governed
by the forgetting gap; the full non-Markovian field dynamics can in
general contain additional frequency-dependent or long-time
structure inherited from $\Sigma_R(\Omega)$.

Finite passive relaxation, however, does not imply that a cognitive
state receiving continuing relevant input must possess the same
finite lifetime.
For an approximately constant effective low-frequency drive,
$I_{\rm rel}(t)=I_*$, Eq.~(\ref{eq:CFN_driven_lowfreq}) gives
\begin{equation}
\phi_*
=
\frac{I_*}{r_{\rm cog}},
\label{eq:CFN_driven_fixed_point}
\end{equation}
provided $r_{\rm cog}>0$ and the static low-frequency approximation
remains valid.
Here $I_*$ represents the net contribution generated by continuing
content-relevant excitation of the collective sector rather than an
infinitesimal response probe.
A positive cognitive forgetting gap is therefore compatible with a
nonzero cognitive field dynamically sustained by continuing relevant
experience.

The periodic source-re-exposure experiments provide a computational
test of this distinction between passive persistence and driven
maintenance.
When a partially relaxed recurrent field is periodically presented
with content-matched source information, each presentation produces a
new input-dependent reorganization of the existing state.
For recurrence interval $T_r$, the driven system may approach a
periodic asymptotic regime,
\begin{equation}
\phi_*(t+T_r)
\simeq
\phi_*(t).
\label{eq:CFN_periodic_field}
\end{equation}

A minimal stroboscopic description, evaluated immediately after
successive renewal events, is
\begin{equation}
m_{n+1}^{+}
=
A_r m_n^{+}
+
\Delta_r,
\qquad
|A_r|<1,
\label{eq:CFN_stroboscopic_map}
\end{equation}
where $m_n^{+}$ denotes a content-specific projection of the recurrent
field immediately after the $n$th renewal event, $A_r$ represents the
net relaxation over one re-exposure interval, and $\Delta_r$ denotes
the field reorganization generated by the incoming source.
The corresponding periodically driven fixed point is
\begin{equation}
m_*^{+}
=
\frac{\Delta_r}{1-A_r}.
\label{eq:CFN_stroboscopic_fixed_point}
\end{equation}
A nonzero asymptotic recurrent state can therefore coexist with finite
relaxation between successive renewal events.

Figure~\ref{fig:representation_sensitive_renewal} shows that the
renewal amplitude is strongly representation dependent.
Exact source re-exposure produces substantially stronger recovery than
near-paraphrased re-exposure, whereas unrelated input and
recurrent-field-off controls remain near background levels.
The effect therefore cannot be attributed simply to the addition of
new input.
Recovery depends on the relation between the incoming representation
and the content-bearing recurrent state.

This representation dependence is consistent with the mode-selective
input coupling derived in Sec.~II.C,
\(
b_\alpha^{\rm re}
=
\widetilde u_\alpha^\dagger
B_\mathcal{X} I_{\rm re}.
\)
Different representations can therefore generate different
perturbations of the collective-mode manifold.
At the CFN level, the corresponding renewal may be written
schematically as
\begin{equation}
\Delta_r
=
\mathcal R
\left(
X_r,\Phi_r
\right).
\label{eq:CFN_renewal_operator}
\end{equation}
The stronger recovery produced by exact than by near-paraphrased
re-exposure is therefore consistent with stronger effective coupling
between the incoming representation and the content-bearing recurrent
state.

The present behavioral measurements do not directly determine the
microscopic modal overlaps
$\widetilde u_\alpha^\dagger B_X I_{\rm re}$.
The correspondence should therefore be understood operationally:
the observed representation dependence has the qualitative structure
expected from mode-selective excitation, but direct verification
requires simultaneous measurement of the internal collective modes
and their input projections.

The extended periodic-re-exposure experiment shown in
Fig.~\ref{fig:periodic_renewal} provides a second test of the driven
interpretation.
For the model trained at $d_{\rm train}=4$, exact source re-exposure
every eight recurrent cycles initially produces a decreasing
peak--trough envelope.
At longer cycle distances, however, the trajectory approaches an
approximately stationary sawtooth regime.
Across the extended 128-cycle observation window, no continuing
systematic decay of the late-time periodic envelope toward the control
baseline is resolved.

This behavior is naturally described by
Eqs.~(\ref{eq:CFN_periodic_field}) and
(\ref{eq:CFN_stroboscopic_fixed_point}).
The recurrent field relaxes between successive source presentations,
producing the descending portion of each sawtooth, while
content-matched input reorganizes and renews the relevant field
component.
The initial decrease of the envelope can therefore be interpreted as
a transient from the strongly encoded initial condition toward a
nonzero periodically driven regime, without requiring a second
independent relaxation process that continues indefinitely toward
zero.

This interpretation does not imply mathematically infinite memory.
The experiment probes only a finite recurrent trajectory, and the
passive-retention measurements independently demonstrate finite
relaxation.
The stronger conclusion is dynamical:
finite passive relaxation does not impose the same finite lifetime on
a recurrent cognitive state that continues to interact with
content-relevant experience.
The state may relax locally between inputs while remaining
macroscopically sustained through repeated input-dependent
reorganization.

The difference between exact and near-paraphrased re-exposure also
clarifies the present computational regime.
The CFN studied here has limited semantic capacity, and its renewal
response remains sensitive to the particular representation supplied
to the recurrent system.
Long-lived recurrent dynamics therefore does not require sophisticated
linguistic reasoning.
Rather, the experiments establish that a temporal substrate combining
collective persistence, finite relaxation, recurrent field
propagation, and representation-dependent renewal can already arise in
a relatively small recurrent Transformer.

Semantic representational capacity and temporal persistence are
therefore distinct properties.
More capable models may develop greater invariance of the effective
input coupling, allowing paraphrases, implications, intermediate
deductions, or other semantically related inputs to reorganize the
same underlying cognitive state.
This remains a prediction for future work rather than a conclusion of
the present experiments.

\section{V. Discussion}
\label{sec:discussion}

The central implication of the present study is not simply
that recurrent computation can extend memory, but that
field re-entry follows naturally from the dynamical
structure of Cognitive Field Theory.
As developed in Sec.~II, collective relaxation modes
generate internal non-Markovian memory feedback,
giving rise to a memory-dressed cognitive field
whose evolution depends on its preceding history.
By incorporating the memory degrees of freedom
into an extended cognitive state, this history-dependent
dynamics can be expressed as a recursive state
evolution in discrete time.
Field re-entry therefore emerges as a natural
computational representation of the underlying
memory-dressed dynamics, rather than an independently
prescribed memory-storage mechanism.
The CFN implements this principle through a learned
recurrent mapping in which the internally generated
cognitive field participates in subsequent inference
together with newly arriving information.
The resulting computation is not a sequence of
independent input--output mappings, but an evolving
recurrent trajectory whose present organization
depends partly on its own computational history.

The experiments demonstrate that learning can organize
this recurrent pathway into nontrivial persistent dynamics.
The re-entered field supports content-specific retrieval,
remains causally necessary for recovering information
from preceding episodes, and acquires a behavioral
persistence scale that increases through temporal training.
Memory remains functionally accessible beyond the
explicitly trained recurrent horizon, although it
gradually decays under unrelated intervening inputs.
When subsequent inputs remain semantically connected
to the remembered episode, the persistence trajectory
is reorganized, while content-related source
re-exposure can selectively renew information after
partial relaxation.
Periodic exact source re-exposure further produces an
approximately stationary nonzero driven regime over
the extended observation window, whereas unrelated-input
and recurrence-off controls do not reproduce this
behavior.
Near-paraphrased input produces weaker renewal than
exact source re-exposure in the present model.
Together, these observations show that recurrent
memory is not the preservation of an unchanged hidden
state, but a learned dynamical process shaped by
recurrent time, internal history, and the relationship
between incoming information and the existing field.

These findings are consistent with the physical picture
of finite forgetting and dynamically sustained cognitive
organization developed in Cognitive Field Theory
\cite{7}.
In that framework, passive relaxation is characterized
by the memory-dressed cognitive forgetting gap
$r_{\rm cog}=r-\Sigma_R(0)$.
A positive forgetting gap permits the cognitive field
to relax in the absence of supporting input, without
requiring that persistent organization disappear under
continued content-relevant driving.
The present experiments provide a computational
illustration of this distinction.
Information carried by the recurrent field gradually
loses behavioral accessibility along unsupported
trajectories, while repeated content-matched input
can maintain a long-lived driven regime.
Such persistence does not imply infinite memory,
exact dynamical freezing, or a vanishing forgetting gap.
Rather, finite relaxation and sustained cognitive
organization can coexist in a dissipative recurrent
system.
The observed behavioral persistence is qualitatively
compatible with this interpretation, although it
does not directly determine the microscopic forgetting
gap or the collective memory self-energy.

This perspective also clarifies the distinction between
episodic computation and continuous cognitive dynamics.
In an episodic architecture, each inference episode
primarily produces an output from its current inputs.
In a recurrent cognitive-field architecture, the
internally generated state can remain active across
computational boundaries and can be reorganized by
subsequent information.
Cognitive continuity therefore need not require
lossless preservation or repeated retrieval of the
complete computational history.
Instead, the dynamically relevant consequences of
preceding computation can remain available through
the evolving field, while other components relax.
The recurrent field thus provides a dynamical substrate
for context-dependent persistence and ongoing
reorganization, rather than a prescribed mechanism
for storing and retrieving past information.
Learning determines how the preceding field
participates in subsequent computation and how
incoming information modifies the evolving state.
The CFN therefore provides a computational framework
in which persistence and inference can emerge as
coupled aspects of recurrent cognitive dynamics.

The same dynamical principle suggests a possible route
toward continuously interacting cognitive systems.
If sensory input drives the evolving cognitive field
and the resulting state contributes to action,
re-entry could operate within a closed
perception--cognition--action loop.
In such a system, the consequences of previous internal
computation and action would remain relevant when new
sensory information arrives, allowing cognition to
unfold through continuing interaction with the
environment rather than isolated inference episodes
\cite{24}.
This possibility is conceptually related to earlier
biological theories of reentry, particularly the
Edelman--Tononi framework, in which reciprocal signaling
among distributed neuronal populations contributes
to the coordination and integration of neural activity
\cite{25,26}.
The present framework addresses a related but distinct
problem: the formation of a memory-dressed collective
field and its explicit causal participation in
subsequent computation.
The relationship is therefore conceptual rather than
one of theoretical identity.
The present experiments do not establish a particular
biological circuit, an evolutionary mechanism for
intelligence, or the realization of continuous
sensorimotor cognition.
These remain directions for future investigation.

An important feature of the CFN is that its recurrent
cognitive state is experimentally accessible.
The field can be recorded, replaced, removed,
perturbed, or propagated across multiple cycles,
allowing its causal contribution to subsequent
inference to be directly examined.
Such interventions provide a means of studying
history-dependent cognitive dynamics beyond
final task performance alone.
Field re-entry also introduces a minimal form
of dynamical self-reference, in which an internally
generated state participates in subsequent
computation.
This does not imply self-awareness,
consciousness, or metacognition.
Nevertheless, experimental access to the recurrent
field may provide a basis for investigating
higher-order functions, including uncertainty
evaluation, the detection of inconsistent information,
and learned modification of ongoing cognitive states.
Whether these capabilities emerge in more capable
recurrent systems remains an open empirical question.

Finally, the collective spectral measurements
provide a physical perspective complementary
to the behavioral memory experiments.
Learning reorganizes the infrared relaxation
spectrum, while recurrent training changes
the observed persistence and renewal of
content-specific memory.
These findings are qualitatively consistent
with the memory-dressed near-critical picture
of Cognitive Field Theory, although their
quantitative relationship remains unresolved.
A more stringent test would reconstruct effective
memory kernels, self-energies, and collective
response functions from the relaxation spectrum
and its mode-dependent coupling structure,
and compare them with independently measured
behavioral relaxation and driven responses.
Controlled perturbations of the recurrent field
and external input could further examine
cognitive susceptibility, history dependence,
and nonlinear response.
Establishing these connections would provide
a more direct test of the field-theoretic
description of persistent cognitive dynamics.

The present study supports a dynamical view of persistent
cognition in which memory and inference are not separate
operations connected by an explicit storage mechanism.
Instead, information from preceding computation remains
functionally available through an evolving collective state
that participates in subsequent inference, relaxes over
finite timescales, and can be reorganized or renewed by
new input.
The CFN demonstrates this principle in a controlled
computational setting and provides a foundation for
investigating how continuous, history-dependent cognitive
dynamics may emerge from learned recurrent field
organization.

\vspace{6pt}
\paragraph{Re-entry and the Limits of Knowledge-Based Intelligence.}

Contemporary large language models exhibit a striking asymmetry
in their cognitive capabilities. They can integrate knowledge
across diverse scientific disciplines, perform sophisticated
mathematical reasoning, and develop complex theoretical
arguments. Yet these capabilities do not necessarily translate
into an equally developed capacity for autonomous scientific
discovery. Identifying important unresolved problems,
sustaining independent inquiry, and progressively revising
hypotheses in response to new evidence remain distinct
challenges. This contrast suggests that the breadth of
acquired knowledge and the autonomous organization of
cognitive dynamics may represent different dimensions
of intelligence.

One possible source of this asymmetry is the limited
continuity of internally generated cognitive states
across successive inference episodes. Although language
models can exploit extensive contextual information,
such context dependence is not equivalent to the
continued evolution of an internal cognitive state.
Cognitive Field Theory provides a dynamical framework
for understanding how collective interactions and
memory feedback organize persistent cognitive states,
while the CFN implements learned field re-entry
across successive computational cycles. From this
perspective, re-entry may provide a foundation not
only for memory persistence but also for sustained
inquiry, hypothesis revision, and progressive
cognitive self-organization.

The present experiments demonstrate content-specific
memory through learned field re-entry, but do not
establish autonomous scientific discovery or
superintelligence. Whether recurrent cognitive
dynamics can support persistent problem formulation,
adaptive hypothesis revision, and the discovery
of new principles remains an important direction
for future investigation.

\section{VI. Conclusion}
\label{sec:conclusion}

We introduced the Cognitive Field Network as a recurrent
architecture in which an internally generated cognitive field remains
causally available to subsequent inference.
The central result is that learning can organize this re-entry pathway
into a functional dynamical substrate that supports content-dependent
persistence, causal influence on later computation, and selective
renewal by subsequent input.

The experiments show that this persistence is dynamical rather than
equivalent to permanent storage.
Without relevant support, recurrent information relaxes over a finite
timescale, whereas semantically continuing trajectories and
content-matched re-exposure can sustain or renew the surviving field.
Repeated relevant input can therefore produce an approximately
stationary nonzero recurrent regime over the observed interval,
even when passive forgetting remains finite.

These results provide a computational realization of the central CFT
picture developed in this work.
Collective relaxation dynamics generate a memory-dressed cognitive
field, finite incoming information reorganizes that field, and
cross-cycle re-entry allows the resulting history-dependent field to
participate in subsequent cognitive dynamics.
Persistent cognition can therefore arise not from the elimination of
forgetting, but from the continuing interaction between finite
persistence, new information, and recurrent field reorganization.

The present CFN is deliberately minimal, and its renewal remains
representation sensitive.
Future work should determine whether larger systems can extend this
mechanism to increasingly abstract semantic transformations,
intermediate reasoning states, and autonomously generated information.
More generally, the CFN provides a controlled computational setting
for relating measurable collective dynamics to persistent,
history-dependent cognition.

\vspace{6pt}
\emph{Acknowledgements}---This work was partially supported by the Institute of Information \& Communications Technology Planning \& Evaluation (IITP) grant 
funded by the Korea government (MSIT) (IITP-RS-2025-02214780).

The author acknowledges the support of ChatGPT (GPT-5, OpenAI) for assistance in literature review and conceptual structuring during early development.

\clearpage
\appendix

\renewcommand{\thefigure}{A\arabic{figure}}
\renewcommand{\theequation}{A\arabic{equation}}

\setcounter{figure}{0}
\setcounter{equation}{0}

\vspace*{1.5cm}
{\centering\large\bfseries Supplementary Materials\par}
\vspace{1.0cm}

\section{Appendix A: From Continuous-Time Cognitive Field Dynamics to Discrete-Time Field Re-entry}
\label{app:cft_cfn_discretization}

\vspace{6pt}
\paragraph{I. Non-Markovian collective cognitive-field dynamics.}

We establish a mathematical connection between the continuous-time
memory-dressed dynamics of Cognitive Field Theory (CFT) and the discrete-time
field re-entry architecture of the Cognitive Field Network (CFN). The
connection is a statement about the form of the state evolution: it does not,
by itself, imply that a trained CFN reproduces a prescribed spectral density,
susceptibility, or memory kernel.

For clarity, we first consider one component of the coarse-grained cognitive
field. The corresponding vector- and operator-valued generalizations follow by
replacing scalar coefficients with matrices or integral operators. Let
$\phi(t)$ denote the collective cognitive field and let $I_{\rm rel}(t)$
denote its effective input. We adopt the retarded memory-feedback convention
\begin{equation}
\frac{d\phi(t)}{dt}
=-r\phi(t)
+\int_0^t K(t-s)\phi(s)\,ds
+I_{\rm rel}(t),
\label{eq:app_cft_gle}
\end{equation}
where $r$ is the bare local relaxation coefficient and $K(t)$ is a causal
memory kernel. In the convention used here, the static self-energy is
$\Sigma_R(0)=\widetilde K(0)$ when this integral is finite, and the
renormalized cognitive forgetting gap is $r_{\rm cog}=r-\Sigma_R(0)$. The full
kernel is retained in the evolution equation, so the static contribution must
not be subtracted again.

The causal spectral representation of the kernel is
\begin{equation}
K(t)=\Theta(t)\int_0^\Lambda d\lambda\,w(\lambda)e^{-\lambda t},
\label{eq:app_kernel_spectral}
\end{equation}
where $w(\lambda)$ is a signed or nonnegative effective spectral coupling
density, depending on the microscopic projection and feedback convention. If
the mode couplings are approximately uniform, $w(\lambda)=g\rho(\lambda)$ with
$\rho(\lambda)$ the time-scale density of states (TDOS) and $g$ an effective
coupling. Thus the TDOS and the memory kernel are related, but need not be
identical: the projection weights also matter.

The Laplace transform, with $\operatorname{Re}z>0$, is
\begin{equation}
\widetilde K(z)
=\int_0^\infty dt\,e^{-zt}K(t)
=\int_0^\Lambda d\lambda\,\frac{w(\lambda)}{z+\lambda}.
\label{eq:app_kernel_laplace}
\end{equation}

Writing $\widetilde\phi(z)=\int_0^\infty e^{-zt}\phi(t)dt$, the transformed
equation is
\begin{equation}
\left[z+r-\widetilde K(z)\right]\widetilde\phi(z)
=\phi(0)+\widetilde I_{\rm rel}(z).
\label{eq:app_laplace_field}
\end{equation}
Consequently, the retarded linear response has the formal structure
\begin{equation}
\chi_R(z)=\frac{1}{z+r-\widetilde K(z)}.
\label{eq:app_response}
\end{equation}

Analytic continuation $z\to 0^+-i\omega$ gives the frequency-domain retarded
response. The zero-frequency inverse response is $r-\widetilde K(0)=r_{\rm
cog}$, provided the static integral is finite. Equivalently, the response can
be written as $\chi_R^{-1}(z)=z+r_{\rm cog}-[\widetilde K(z)-\widetilde
K(0)]$, which avoids double-counting the static self-energy.

\vspace{6pt}
\paragraph{II. Exact time-slab integration and discrete memory recurrence.}

Define a uniform temporal mesh
\begin{equation}
t_n=n\Delta t,\qquad \phi_n=\phi(t_n),\qquad \Delta t>0.
\label{eq:app_time_mesh}
\end{equation}
Integrating Eq.~\eqref{eq:app_cft_gle} over the interval $[t_n,t_{n+1}]$
yields the exact identity
\begin{align}
\phi_{n+1}={}&\phi_n
-r\int_{t_n}^{t_{n+1}}\phi(s)\,ds\nonumber\\
&+\int_{t_n}^{t_{n+1}}ds\int_0^s du\,K(s-u)\phi(u)\nonumber\\
&+\int_{t_n}^{t_{n+1}}I_{\rm rel}(s)\,ds.
\label{eq:app_exact_slab}
\end{align}

No Markovian approximation has been made. The double integral retains the full
history before and during the time slab. In particular,
Eq.~\eqref{eq:app_exact_slab} does not generally define a closed update map on
the scalar variable $\phi_n$ alone.

To display the history dependence explicitly, consider a first-order
quadrature away from a possible singularity at the origin. With $I_n\simeq
I_{\rm rel}(t_n)$ and $K_j=K(j\Delta t)$, a formal left-endpoint
discretization gives
\begin{equation}
\phi_{n+1}\simeq
(1-r\Delta t)\phi_n
+\Delta t^2\sum_{k=0}^{n}K_{n-k}\phi_k
+\Delta t I_n.
\label{eq:app_naive_discrete}
\end{equation}

This formula is illustrative, not a uniformly valid numerical scheme when
$K(0)$ is divergent or unresolved. A more robust product-integration
approximation uses cell-integrated kernel weights
\begin{equation}
W_j=\int_{j\Delta t}^{(j+1)\Delta t}K(u)\,du,
\label{eq:app_cell_weights}
\end{equation}
and approximates the convolution at $t_n$ by
\begin{equation}
\int_0^{t_n}K(t_n-s)\phi(s)\,ds
\simeq\sum_{k=0}^{n-1}W_{n-1-k}\phi_k.
\label{eq:app_product_integration}
\end{equation}

The corresponding explicit time step is
\begin{equation}
\phi_{n+1}\simeq
(1-r\Delta t)\phi_n
+\Delta t\sum_{k=0}^{n-1}W_{n-1-k}\phi_k
+\Delta t I_n.
\label{eq:app_history_update}
\end{equation}
Its key structural feature is
\begin{equation}
\phi_{n+1}=\mathcal F_{\Delta t}
(\phi_n,\phi_{n-1},\ldots,\phi_0;I_n),
\label{eq:app_history_map}
\end{equation}
which is non-Markovian in the reduced field variable. A closed one-step
recurrence becomes possible after the memory-bearing degrees of freedom are
included in the state.

\vspace{6pt}
\paragraph{III. Embedding the memory kernel in an extended state space.}

Suppose first that the memory kernel admits a finite exponential
representation
\begin{equation}
K(t)=\sum_{\alpha=1}^{M}g_\alpha e^{-\lambda_\alpha t},
\qquad \lambda_\alpha>0,\quad t\geq0.
\label{eq:app_exp_kernel}
\end{equation}
For each relaxation mode, define an auxiliary memory coordinate
\begin{equation}
m_\alpha(t)=\int_0^t e^{-\lambda_\alpha(t-s)}\phi(s)\,ds.
\label{eq:app_memory_coordinate}
\end{equation}
Applying the Leibniz rule gives
\begin{align}
\frac{dm_\alpha(t)}{dt}
&=e^{-\lambda_\alpha(t-t)}\phi(t)
+\int_0^t\frac{\partial}{\partial t}
\left[e^{-\lambda_\alpha(t-s)}\phi(s)\right]ds\nonumber\\
&=\phi(t)-\lambda_\alpha
\int_0^t e^{-\lambda_\alpha(t-s)}\phi(s)\,ds\nonumber\\
&=\phi(t)-\lambda_\alpha m_\alpha(t).
\label{eq:app_aux_derivation}
\end{align}

The original memory convolution becomes
\begin{align}
\int_0^t K(t-s)\phi(s)\,ds
&=\sum_{\alpha=1}^{M}g_\alpha
\int_0^t e^{-\lambda_\alpha(t-s)}\phi(s)\,ds\nonumber\\
&=\sum_{\alpha=1}^{M}g_\alpha m_\alpha(t).
\label{eq:app_memory_sum}
\end{align}

Therefore Eq.~\eqref{eq:app_cft_gle} is equivalent, for the stated kernel and
initial auxiliary values $m_\alpha(0)=0$, to
\begin{equation}
\begin{aligned}
\dot\phi(t)&=-r\phi(t)
+\sum_{\alpha=1}^{M}g_\alpha m_\alpha(t)
+I_{\rm rel}(t),\\
\dot m_\alpha(t)&=-\lambda_\alpha m_\alpha(t)+\phi(t),
\qquad \alpha=1,\ldots,M.
\end{aligned}
\label{eq:app_markov_embedding}
\end{equation}

For nonzero prehistory, the auxiliary initial values must encode the
appropriate memory inherited from times before the chosen initial time; zero
auxiliary initial values correspond to the integral beginning at $t=0$.

Introduce the extended state vector
\begin{equation}
\mathbf\Phi(t)=
\begin{pmatrix}
\phi(t)&m_1(t)&m_2(t)&\cdots&m_M(t)
\end{pmatrix}^{\!T}.
\label{eq:app_extended_state}
\end{equation}
Then
\begin{equation}
\frac{d\mathbf\Phi(t)}{dt}
=A\mathbf\Phi(t)+B I_{\rm rel}(t),
\label{eq:app_state_space}
\end{equation}
with
\begin{equation}
A=
\begin{pmatrix}
-r&g_1&g_2&\cdots&g_M\\
1&-\lambda_1&0&\cdots&0\\
1&0&-\lambda_2&\cdots&0\\
\vdots&\vdots&\vdots&\ddots&\vdots\\
1&0&0&\cdots&-\lambda_M
\end{pmatrix},
\qquad
B=\begin{pmatrix}1&0&\cdots&0\end{pmatrix}^{T}.
\label{eq:app_state_matrix}
\end{equation}

The extended system is Markovian in $\mathbf\Phi(t)$ even though its projected
equation for $\phi(t)$ is non-Markovian. This is the mathematical distinction
between a reduced field and a memory-complete state. More generally, a
continuous distribution of relaxation rates produces an infinite-dimensional
embedding; a finite set of auxiliary coordinates then constitutes an
approximation over a specified time and frequency range.

\vspace{6pt}
\paragraph{IV. Exact discrete-time propagator and input coupling.}

For time-independent $A$ and $B$, variation of constants gives
\begin{equation}
\mathbf\Phi(t_{n+1})
=e^{A\Delta t}\mathbf\Phi(t_n)
+\int_{t_n}^{t_{n+1}}
e^{A(t_{n+1}-s)}B I_{\rm rel}(s)\,ds.
\label{eq:app_exact_propagator}
\end{equation}
Define
\begin{equation}
A_d=e^{A\Delta t}.
\label{eq:app_discrete_A}
\end{equation}
If the input is held constant over the interval,
$I_{\rm rel}(s)=X_{n+1}$ for $t_n\leq s<t_{n+1}$,
then
\begin{align}
\mathbf\Phi_{n+1}
&=A_d\mathbf\Phi_n+B_d X_{n+1},
\label{eq:app_discrete_ssm}\\
B_d&=\int_0^{\Delta t}e^{Au}B\,du.
\label{eq:app_discrete_B}
\end{align}

This integral expression for $B_d$ remains valid even when
$A$ is singular. When $A$ is invertible, it may also be
written as
$B_d=A^{-1}(e^{A\Delta t}-\mathbb I)B$.
For small $\Delta t$,
\begin{align}
A_d&=\mathbb I+A\Delta t+O(\Delta t^2),\\
B_d&=B\Delta t+\tfrac12 AB\Delta t^2
+O(\Delta t^3).
\label{eq:app_small_dt}
\end{align}
Thus the first-order approximation is
\begin{equation}
\mathbf\Phi_{n+1}
=\mathbf\Phi_n
+\Delta t\left(A\mathbf\Phi_n+B X_{n+1}\right)
+O(\Delta t^2).
\label{eq:app_euler_extended}
\end{equation}

If the collective dynamics is nonlinear or the input
changes the effective relaxation generator, the extended
equation becomes
\begin{equation}
\dot{\mathbf\Phi}(t)
=\mathcal G_\theta
\bigl(\mathbf\Phi(t),I_{\rm rel}(t)\bigr).
\label{eq:app_nonlinear_extended}
\end{equation}

Under suitable regularity assumptions, its solution
over one time slab defines a flow map. For a
piecewise-constant input this takes the form
\begin{equation}
\mathbf\Phi_{n+1}
=\mathcal U_{\theta,\Delta t}
\bigl(\mathbf\Phi_n,X_{n+1}\bigr).
\label{eq:app_nonlinear_flow}
\end{equation}

If the input is not constant, the exact flow depends
on the entire input path over the interval, rather
than solely on one sampled value. An input token
$X_{n+1}$ then represents or approximates that path.

Equations~\eqref{eq:app_discrete_ssm} and
\eqref{eq:app_nonlinear_flow} establish that a
memory-bearing continuous-time system admits a
one-step discrete-time evolution when the carried
state includes the required memory degrees of freedom.

\vspace{6pt}
\paragraph{V. Recovery of the discrete memory kernel.}

The memory structure retained by the extended
discrete-time state can be made explicit by eliminating
its auxiliary coordinates. Partition the state as
$\mathbf\Phi_n=(\phi_n,\mathbf m_n)^T$ and write
the exact discrete propagator in block form,
\begin{equation}
A_d=
\begin{pmatrix}
A_{\phi\phi}&A_{\phi m}\\
A_{m\phi}&A_{mm}
\end{pmatrix},
\qquad
B_d=
\begin{pmatrix}
B_\phi\\
B_m
\end{pmatrix}.
\label{eq:app_discrete_blocks}
\end{equation}
Here $A_{\phi\phi}$ describes the direct field
propagation, $A_{\phi m}$ and $A_{m\phi}$ represent
the couplings between the field and auxiliary
memory coordinates, and $A_{mm}$ propagates the
auxiliary memory sector. Equation~\eqref{eq:app_discrete_ssm}
then becomes
\begin{equation}
\begin{aligned}
\phi_{n+1}
&=A_{\phi\phi}\phi_n
+A_{\phi m}\mathbf m_n+B_\phi X_{n+1},\\
\mathbf m_{n+1}
&=A_{m\phi}\phi_n
+A_{mm}\mathbf m_n+B_m X_{n+1}.
\end{aligned}
\label{eq:app_discrete_partition}
\end{equation}

Iterating the second equation gives
\begin{align}
\mathbf m_n
={}&A_{mm}^{\,n}\mathbf m_0
\nonumber\\
&+\sum_{j=0}^{n-1}
A_{mm}^{\,n-1-j}
\left(A_{m\phi}\phi_j+B_mX_{j+1}\right).
\label{eq:app_discrete_memory_solution}
\end{align}
Substitution into the field equation yields
\begin{align}
\phi_{n+1}
={}&A_{\phi\phi}\phi_n
+\sum_{j=0}^{n-1}
A_{\phi m}A_{mm}^{\,n-1-j}
A_{m\phi}\phi_j
\nonumber\\
&+A_{\phi m}A_{mm}^{\,n}\mathbf m_0
+B_\phi X_{n+1}
\nonumber\\
&+\sum_{j=0}^{n-1}
A_{\phi m}A_{mm}^{\,n-1-j}
B_mX_{j+1}.
\label{eq:app_recovered_memory}
\end{align}

The discrete memory kernel is therefore
\begin{equation}
K_k^{(d)}
=A_{\phi m}A_{mm}^{\,k}A_{m\phi},
\qquad k=0,1,2,\ldots .
\label{eq:app_discrete_kernel}
\end{equation}

This result establishes explicitly that a one-step
Markovian evolution in the extended state produces
a history-dependent equation for the reduced field.
The blocks appearing in Eq.~\eqref{eq:app_discrete_kernel}
are determined by the exact propagator
$A_d=e^{A\Delta t}$ rather than introduced as
independent phenomenological coefficients.

For a diagonalizable auxiliary propagator,
\begin{equation}
A_{mm}
=\sum_\alpha
\zeta_\alpha
u_\alpha l_\alpha^\dagger,
\qquad
l_\alpha^\dagger u_\beta=\delta_{\alpha\beta},
\label{eq:app_aux_spectral}
\end{equation}
the kernel becomes
\begin{equation}
K_k^{(d)}
=\sum_\alpha
(A_{\phi m}u_\alpha)
(l_\alpha^\dagger A_{m\phi})
\zeta_\alpha^k.
\label{eq:app_discrete_spectral_kernel}
\end{equation}
For stable modes, the associated relaxation rates are
\begin{equation}
\lambda_\alpha^{(d)}
=-\frac{\ln|\zeta_\alpha|}{\Delta t}.
\label{eq:app_kernel_rates}
\end{equation}

Thus the temporal persistence of the reduced field
depends jointly on the relaxation spectrum and the
coupling weights of the eliminated modes. The
continuous-time kernel and its discrete counterpart
are connected through the extended propagator,
although the discrete weights need not equal simple
samples of the continuous kernel. In particular,
exact discretization of the coupled system generally
mixes the field and auxiliary relaxation modes.

\vspace{6pt}
\paragraph{VI. Cognitive-field re-entry and cross-attention coupling.}

The defining CFN recurrence is
\begin{equation}
\Phi_{n+1}=F_\theta(X_{n+1},\Phi_n).
\label{eq:app_cfn_reentry}
\end{equation}
Here $\Phi_n$ denotes the complete token-resolved
hidden field carried from one inference cycle to the
next. Comparison with
Eq.~\eqref{eq:app_nonlinear_flow} establishes a
structural correspondence between the extended
continuous-time state and the recurrent CFN field.

In the present implementation, this state continuity
is realized through learned cross-attention coupling.
For a given Transformer layer, let $H_{n+1}$ denote
the current hidden representation and $\Phi_n$ the
re-entered field. The cross-attention contribution
has the schematic form
\begin{equation}
C_{n+1}
=
\operatorname{softmax}
\left[
\frac{
(H_{n+1}W_Q)(\Phi_nW_K)^T
}{\sqrt{d_k}}
\right]
(\Phi_nW_V),
\label{eq:app_cross_attention}
\end{equation}
with the appropriate output projection and learned
recurrent coupling understood.

The previous field enters both the attention keys
and values. Consequently, cross-attention determines
not only which components of the preceding field
are selected by the current representation, but also
how their information contributes to the next
computational state.

To characterize this coupling dynamically, linearize
the learned CFN recurrence around a reference state,
\begin{equation}
\delta\Phi_{n+1}
=J_n\delta\Phi_n
+G_n\delta X_{n+1},
\label{eq:app_cfn_linearized}
\end{equation}
where
\begin{equation}
J_n=
\left.
\frac{\partial F_\theta(X_{n+1},\Phi)}
{\partial\Phi}
\right|_{\Phi=\Phi_n}.
\label{eq:app_cfn_jacobian}
\end{equation}

The recurrent Jacobian contains the derivatives of
the cross-attention pathway together with those of
the remaining Transformer operations. In particular,
the derivative of cross-attention with respect to the
previous field contains both value-transmission and
key-dependent attention-selection contributions.

For a time-independent linearization, introduce a
projection of the recurrent state into a collective
field coordinate $\varphi_n$ and complementary
internal coordinates $\mathbf m_n$. In this basis,
write
\begin{equation}
J_{\rm rec}
=
\begin{pmatrix}
J_{\phi\phi}&J_{\phi m}\\
J_{m\phi}&J_{mm}
\end{pmatrix}.
\label{eq:app_cfn_jacobian_blocks}
\end{equation}
Elimination of the complementary coordinates then
defines an effective discrete memory kernel,
\begin{equation}
K_k^{\rm CFN}
=J_{\phi m}J_{mm}^{\,k}J_{m\phi},
\qquad k=0,1,2,\ldots .
\label{eq:app_cfn_effective_kernel}
\end{equation}

This construction provides a quantitative route
from the learned recurrent coupling to an effective
memory response. Cross-attention contributes to
the Jacobian blocks that determine the propagation
and feedback of internal perturbations. It is
therefore a learnable coupling mechanism, rather
than the memory kernel itself.

The resulting kernel depends on the selected
collective projection and the reference dynamics.
Moreover, $F_\theta$ need not be the numerical
discretization of a prescribed CFT kernel.
Reproducing a particular continuous-time memory
response requires the learned recurrent state to
represent the relevant memory-bearing degrees
of freedom and their effective couplings.

The architectural correspondence is nevertheless
well defined. In a continuously evolving dynamical
system, the final state of one interval becomes
the initial state of the next. In CFN, explicit
field re-entry establishes analogous continuity
across discrete inference cycles. This identifies
a shared state-evolution principle without
asserting biological or spectral equivalence.

\vspace{6pt}
\paragraph{VII. Functional and spectral tests of the correspondence.}

The CFN selective-retrieval protocol provides a
functional test of whether information carried by
the recurrent hidden field influences later inference.
Let $P_{\rm EM}^{C}(d)$, $P_{\rm EM}^{W}(d)$, and
$P_{\rm EM}^{O}(d)$ denote exact-match accuracy at
bridge distance $d$ with the correct re-entered
field, an incorrect field, and re-entry disabled,
respectively. The content-specific memory signal is
\begin{equation}
M(d)=P_{\rm EM}^{C}(d)-P_{\rm EM}^{W}(d).
\label{eq:app_content_signal}
\end{equation}

A positive $M(d)$, together with the appropriate
OFF control, demonstrates that information carried
by the re-entered field causally contributes to
later inference under the tested interventions.
It does not independently identify the spectral
kernel responsible for retention.

A complementary dynamical test would measure the
recurrent Jacobian in
Eq.~\eqref{eq:app_cfn_jacobian} along representative
trajectories and examine the propagation of
perturbations through successive inference cycles.
For a stationary linearized map with eigenvalues
$\zeta_\alpha$, the discrete relaxation rates are
\begin{equation}
\lambda_\alpha
=-\frac{1}{\Delta t}\ln|\zeta_\alpha|.
\label{eq:app_discrete_rates}
\end{equation}

For time-dependent linearizations, the perturbation
propagator is instead
\begin{equation}
\mathcal P_{n+k,n}
=J_{n+k-1}\cdots J_{n+1}J_n.
\label{eq:app_cfn_propagator}
\end{equation}
In nonnormal or nonstationary dynamics, individual
Jacobian eigenvalues need not characterize long-time
relaxation; propagator spectra, singular-value
growth, or Lyapunov exponents may also be required.

Beyond measuring the relaxation spectrum, the
projected Jacobian blocks in
Eq.~\eqref{eq:app_cfn_jacobian_blocks} provide a
direct route to the effective kernel
$K_k^{\rm CFN}$. Its mode-dependent weights and
temporal decay can then be compared with the
spectral memory structure predicted by CFT.

Such measurements would distinguish the general
state-continuity principle demonstrated by the
present CFN experiments from the stronger
hypothesis that learning organizes recurrent
dynamics into the infrared memory structure
described by Cognitive Field Theory.

\renewcommand{\thefigure}{B\arabic{figure}}
\renewcommand{\theequation}{B\arabic{equation}}

\setcounter{figure}{0}
\setcounter{equation}{0}

\vspace{1.0cm}

\section{Appendix B: Driven Cognitive-Field Dynamics,
Memory Dressing, and Field Re-entry}
\label{app:driven_inference}

The main text describes cognitive inference as a driven collective
dynamical process in which incoming information acts on an already
history-dependent cognitive state.
The purpose of this Appendix is to make the corresponding dynamical
structure explicit and to distinguish two forms of recurrence that
play different roles in Cognitive Field Theory (CFT) and the Cognitive
Field Network (CFN).

The projection procedure used below is closely related to the linear
Mori--Zwanzig construction: eliminating a complementary dynamical
sector generates an effective equation containing a retarded memory
kernel and an effective fluctuating force.
Here this standard projection principle is specialized to the
cognitive-field decomposition.
This makes it possible to relate the memory kernel explicitly to the
non-Hermitian collective relaxation spectrum and, at the same time,
to determine how a finite cognitive input is transmitted through that
spectrum to the macroscopic cognitive field.

The resulting memory dressing should not be identified with the
cross-cycle field re-entry introduced by the CFN.
Memory dressing is an endogenous consequence of eliminating internal
collective modes that remain dynamically coupled to the cognitive
field.
Field re-entry instead makes an already organized cognitive field
causally available to a subsequent inference cycle.
The distinction between these two dynamical levels is the central
purpose of the derivation below.

\vspace{6pt}
\paragraph{I. Driven cognitive dynamics and collective projection.}

We begin from the cognitive dynamics linearized around a reference
trajectory or locally stationary cognitive state,
\begin{equation}
\partial_t\delta x(t)
=
-J\,\delta x(t)
+
B I(t)
+
\xi(t),
\label{eq:app_driven_linear}
\end{equation}
where $\delta x(t)$ denotes the deviation of the microscopic or
mesoscopic cognitive state from the reference state, $J$ is the local
stability operator, $I(t)$ is the external cognitive input, $B$
specifies how that input couples to the internal cognitive degrees of
freedom, and $\xi(t)$ denotes unresolved fluctuations.

Importantly, Eq.~(\ref{eq:app_driven_linear}) is not a
linear-response definition.
The input $I(t)$ is part of the physical dynamical equation and may be
finite.
The approximation at this stage is the local linearization of the
internal dynamics around the chosen trajectory or state, rather than
an assumption that the cognitive input itself is infinitesimal.

To separate the macroscopic cognitive field from the complementary
dynamical sector, let $v$ denote a right collective direction and $w$
the corresponding left projection vector, normalized by
\begin{equation}
w^\dagger v=1.
\label{eq:app_collective_norm}
\end{equation}
We define
\begin{equation}
P=vw^\dagger,
\qquad
Q=1-P,
\label{eq:app_projectors}
\end{equation}
so that
\begin{equation}
P^2=P,
\qquad
Q^2=Q,
\qquad
PQ=QP=0.
\end{equation}

The macroscopic collective coordinate and complementary relaxation
sector are then
\begin{equation}
\phi(t)
=
w^\dagger\delta x(t),
\qquad
\mathcal{X}(t)
=
Q\,\delta x(t),
\label{eq:app_collective_variables}
\end{equation}
and therefore
\begin{equation}
\delta x(t)
=
v\phi(t)
+
\mathcal{X}(t),
\qquad
w^\dagger\mathcal{X}(t)=0.
\label{eq:app_state_decomposition}
\end{equation}

The macroscopic field is thus not introduced as an additional degree
of freedom on top of a complete microscopic mode expansion.
Rather, the full dynamical state is first separated into a collective
coordinate and a complementary sector.
The relaxation modes introduced below are modes of this complementary
sector.

\vspace{6pt}
\paragraph{II. Exact field--mode block dynamics.}

Applying $w^\dagger$ and $Q$ to
Eq.~(\ref{eq:app_driven_linear}), and using
Eq.~(\ref{eq:app_state_decomposition}), gives
\begin{align}
\partial_t\phi(t)
&=
-r\phi(t)
+
D\mathcal{X}(t)
+
B_\phi I(t)
+
\xi_\phi(t),
\label{eq:app_phi_block}
\\
\partial_t\mathcal{X}(t)
&=
-M\mathcal{X}(t)
+
C\phi(t)
+
B_{\mathcal X}I(t)
+
\xi_{\mathcal X}(t),
\label{eq:app_X_block}
\end{align}
where
\begin{equation}
\begin{split}
r&\equiv w^\dagger Jv,
\qquad
D\equiv-w^\dagger JQ,
\qquad
C\equiv-QJv,
\\
M&\equiv QJQ,
\qquad
B_\phi\equiv w^\dagger B,
\qquad
B_{\mathcal X}\equiv QB,
\\
\xi_\phi&\equiv w^\dagger\xi,
\qquad
\xi_{\mathcal X}\equiv Q\xi.
\end{split}
\label{eq:app_block_definitions}
\end{equation}

These equations constitute the exact block representation of the
locally linearized driven dynamics for fixed projectors.
They also show that the direct drive of the macroscopic field and the
excitation of the complementary relaxation sector originate from the
same microscopic input $BI(t)$.
No independent field-level source needs to be introduced.

This point is important for cognitive inference.
Incoming information is not simply appended to an already constructed
macroscopic field equation.
It first acts on the underlying cognitive degrees of freedom, and its
effective action on the field is determined by the learned collective
dynamical structure.

\vspace{6pt}
\paragraph{III. Non-Hermitian collective relaxation modes.}

The complementary operator
\begin{equation}
M=QJQ
\end{equation}
is generally non-Hermitian for nonequilibrium cognitive dynamics.
We therefore introduce biorthogonal right and left eigenmodes,
\begin{equation}
M u_\alpha
=
\mu_\alpha u_\alpha,
\qquad
\widetilde u_\alpha^\dagger M
=
\mu_\alpha\widetilde u_\alpha^\dagger,
\qquad
\widetilde u_\alpha^\dagger u_\beta
=
\delta_{\alpha\beta},
\label{eq:app_eigenmodes}
\end{equation}
with
\begin{equation}
\mu_\alpha
=
\lambda_\alpha+i\omega_\alpha.
\label{eq:app_mu}
\end{equation}
Here $\lambda_\alpha$ is the relaxation rate and $\omega_\alpha$ is
the intrinsic circulation frequency of the collective mode.

Expanding
\begin{equation}
\mathcal{X}(t)
=
\sum_\alpha
\mathcal{X}_\alpha(t)u_\alpha,
\label{eq:app_mode_expansion}
\end{equation}
and projecting Eq.~(\ref{eq:app_X_block}) onto the left eigenvectors
gives
\begin{equation}
\partial_t\mathcal{X}_\alpha(t)
=
-\mu_\alpha\mathcal{X}_\alpha(t)
+
c_\alpha\phi(t)
+
b_\alpha(t)
+
\eta_\alpha(t),
\label{eq:app_modal_driven}
\end{equation}
where
\begin{equation}
c_\alpha
=
\widetilde u_\alpha^\dagger C,
\qquad
b_\alpha(t)
=
\widetilde u_\alpha^\dagger B_{\mathcal X}I(t),
\qquad
\eta_\alpha(t)
=
\widetilde u_\alpha^\dagger\xi_{\mathcal X}(t).
\label{eq:app_modal_definitions}
\end{equation}

Similarly,
\begin{equation}
D\mathcal{X}(t)
=
\sum_\alpha d_\alpha\mathcal{X}_\alpha(t),
\qquad
d_\alpha
=
D u_\alpha,
\label{eq:app_field_mode_projection}
\end{equation}
so that the field equation becomes
\begin{equation}
\partial_t\phi(t)
=
-r\phi(t)
+
\sum_\alpha d_\alpha\mathcal{X}_\alpha(t)
+
B_\phi I(t)
+
\xi_\phi(t).
\label{eq:app_phi_modes}
\end{equation}

Equations~(\ref{eq:app_modal_driven}) and
(\ref{eq:app_phi_modes}) make explicit the dynamical role of finite
cognitive input.
The quantity $b_\alpha(t)$ is the mode-selective input overlap:
it determines how strongly incoming information excites each
collective relaxation direction.
The resulting perturbation then evolves according to the intrinsic
relaxation and circulation scales $(\lambda_\alpha,\omega_\alpha)$
while remaining coupled to the macroscopic field through
$c_\alpha$ and $d_\alpha$.

Thus the same collective spectrum mediates both the internal
field--mode feedback and the dynamical transmission of new
information.

\vspace{6pt}
\paragraph{IV. Elimination of the relaxation sector and memory dressing.}

The exact solution of Eq.~(\ref{eq:app_modal_driven}) for an initial
time $t_0$ is
\begin{align}
\mathcal{X}_\alpha(t)
={}&
e^{-\mu_\alpha(t-t_0)}
\mathcal{X}_\alpha(t_0)
\nonumber\\
&+
\int_{t_0}^{t}dt'\,
e^{-\mu_\alpha(t-t')}
\left[
c_\alpha\phi(t')
+
b_\alpha(t')
+
\eta_\alpha(t')
\right].
\label{eq:app_mode_solution}
\end{align}

Substitution into Eq.~(\ref{eq:app_phi_modes}) eliminates the
complementary relaxation sector and yields
\begin{align}
\partial_t\phi(t)
={}&
-r\phi(t)
+
\int_{t_0}^{t}dt'\,
K(t-t')\phi(t')
\nonumber\\
&+
I_{\rm eff}(t)
+
\xi_{\rm eff}(t)
+
\zeta_{\rm init}(t),
\label{eq:app_effective_field}
\end{align}
where
\begin{equation}
K(\tau)
=
\Theta(\tau)
\sum_\alpha
d_\alpha c_\alpha
e^{-\mu_\alpha\tau}
\label{eq:app_memory_kernel}
\end{equation}
is the internally generated memory kernel,

\begin{equation}
I_{\rm eff}(t)
=
B_\phi I(t)
+
\sum_\alpha d_\alpha
\int_{t_0}^{t}dt'\,
e^{-\mu_\alpha(t-t')}
b_\alpha(t')
\label{eq:app_effective_input}
\end{equation}
is the effective cognitive drive,

\begin{equation}
\xi_{\rm eff}(t)
=
\xi_\phi(t)
+
\sum_\alpha d_\alpha
\int_{t_0}^{t}dt'\,
e^{-\mu_\alpha(t-t')}
\eta_\alpha(t')
\label{eq:app_effective_noise}
\end{equation}
is the effective fluctuating force, and
\begin{equation}
\zeta_{\rm init}(t)
=
\sum_\alpha
d_\alpha
e^{-\mu_\alpha(t-t_0)}
\mathcal{X}_\alpha(t_0)
\label{eq:app_initial_term}
\end{equation}
contains the decaying dependence on the initial complementary state.

Equation~(\ref{eq:app_effective_field}) has the generalized-Langevin
structure expected from a Mori--Zwanzig-type elimination.
The important specialization here is that both the memory kernel and
the effective finite cognitive drive are resolved explicitly in terms
of the same collective relaxation modes.

The two terms nevertheless represent different causal processes.
The kernel $K$ is generated by endogenous field--mode coupling:
the existing field perturbs the complementary collective modes and
their subsequent dynamics feeds back onto the field.
By contrast, $I_{\rm eff}$ describes how new external information
excites those modes and is subsequently transmitted to the
macroscopic field.
Memory dressing and information driving therefore share the same
collective dynamical manifold without being the same physical
process.

\vspace{6pt}
\paragraph{V. Spectral representation and the driven cognitive field.}

The mode-resolved memory kernel may be written in terms of the
coupling-weighted spectral density
\begin{equation}
\rho_K(\lambda,\omega)
=
\sum_\alpha
d_\alpha c_\alpha\,
\delta(\lambda-\lambda_\alpha)
\delta(\omega-\omega_\alpha),
\label{eq:app_rhoK}
\end{equation}
which gives
\begin{equation}
K(t)
=
\Theta(t)
\int d\lambda\,d\omega\,
\rho_K(\lambda,\omega)
e^{-\lambda t}
e^{-i\omega t}.
\label{eq:app_K_spectral}
\end{equation}

The coupling-weighted density $\rho_K$ should be distinguished from
the normalized collective mode density
\begin{equation}
\rho(\lambda,\omega)
=
\frac{1}{N}
\sum_\alpha
\delta(\lambda-\lambda_\alpha)
\delta(\omega-\omega_\alpha).
\label{eq:app_raw_rho}
\end{equation}
When the field--mode couplings vary slowly across the infrared sector,
the two inherit the same leading infrared structure up to the
corresponding coupling weight.

Using the Fourier convention
\begin{equation}
\phi(t)
=
\int\frac{d\Omega}{2\pi}\,
e^{-i\Omega t}\phi(\Omega),
\end{equation}
the memory self-energy becomes
\begin{equation}
\Sigma_R(\Omega)
=
\int_0^\infty dt\,
e^{i\Omega t}K(t)
=
\sum_\alpha
\frac{d_\alpha c_\alpha}
{\mu_\alpha-i\Omega},
\label{eq:app_self_energy}
\end{equation}
or equivalently
\begin{equation}
\Sigma_R(\Omega)
=
\int d\lambda\,d\omega\,
\frac{\rho_K(\lambda,\omega)}
{\lambda-i(\Omega-\omega)}.
\label{eq:app_self_energy_spectral}
\end{equation}

For an input coupling that is linear in $I$, define
\begin{equation}
q_\alpha
\equiv
\widetilde u_\alpha^\dagger B_{\mathcal X}.
\label{eq:app_input_overlap}
\end{equation}
The corresponding input-transfer function is
\begin{equation}
\mathcal T_I(\Omega)
=
B_\phi
+
\sum_\alpha
\frac{d_\alpha q_\alpha}
{\mu_\alpha-i\Omega},
\label{eq:app_input_transfer}
\end{equation}
so that
\begin{equation}
I_{\rm eff}(\Omega)
=
\mathcal T_I(\Omega)I(\Omega).
\label{eq:app_input_frequency}
\end{equation}

Neglecting the decaying initial transient, the driven cognitive-field
equation becomes
\begin{equation}
\left[
-i\Omega+r-\Sigma_R(\Omega)
\right]
\phi(\Omega)
=
\mathcal T_I(\Omega)I(\Omega)
+
\xi_{\rm eff}(\Omega).
\label{eq:app_frequency_field}
\end{equation}

Defining
\begin{equation}
L_{\rm cog}(\Omega)
=
\frac{1}
{-i\Omega+r-\Sigma_R(\Omega)},
\label{eq:app_Lcog}
\end{equation}
we obtain
\begin{equation}
\phi(\Omega)
=
L_{\rm cog}(\Omega)
\mathcal T_I(\Omega)I(\Omega)
+
L_{\rm cog}(\Omega)
\xi_{\rm eff}(\Omega).
\label{eq:app_driven_solution}
\end{equation}

This expression separates two essential structures of driven
cognitive inference.
The factor $\mathcal T_I(\Omega)I(\Omega)$ describes how new information
enters and propagates through the collective mode manifold, whereas
$L_{\rm cog}(\Omega)$ describes the memory-dressed internal dynamics
through which that information is integrated with the consequences of
preceding cognitive activity.

The identity
\begin{equation}
L_{\rm cog}^{-1}(\Omega)
=
L_0^{-1}(\Omega)
-
\Sigma_R(\Omega),
\qquad
L_0(\Omega)
=
\frac{1}{-i\Omega+r},
\label{eq:app_Dyson_identity}
\end{equation}
has the usual Dyson form.
No expansion of the external cognitive input is implied by this
identity.
The self-energy describes internal memory dressing, while the finite
input remains the physical drive acting on the dressed dynamical
system.

For completeness, a separate infinitesimal auxiliary probe $h(t)$
may be introduced to define the retarded susceptibility around the
driven background,
\begin{equation}
\chi_R(t,t';[I])
=
\left.
\frac{
\delta\langle\phi(t)\rangle_{I,h}
}{
\delta h(t')
}
\right|_{h=0}.
\label{eq:app_chi}
\end{equation}
In a stationary linearized regime,
\begin{equation}
\chi_R(\Omega)
=
L_{\rm cog}(\Omega),
\end{equation}
but the two roles should not be confused:
$I(t)$ drives the cognitive dynamics, whereas $h(t)$ probes its
response.

In the low-frequency stationary regime one may further define
\begin{equation}
r_{\rm cog}
=
r-\Sigma_R(0).
\label{eq:app_rcog}
\end{equation}
If the residual frequency dependence of the self-energy can be
neglected over the relevant range,
\begin{equation}
L_{\rm cog}(\Omega)
\simeq
\frac{1}
{-i\Omega+r_{\rm cog}},
\end{equation}
giving an effective persistence scale
$\tau_{\rm cog}\sim r_{\rm cog}^{-1}$.
For a broad infrared spectrum, however,
$\Sigma_R(\Omega)-\Sigma_R(0)$ may remain nonanalytic, and the exact
long-time relaxation need not be purely exponential.

\vspace{6pt}
\paragraph{VI. Memory dressing and cognitive-field re-entry.}

The derivation above establishes how a history-dependent cognitive
field arises from the interaction between a macroscopic collective
coordinate and a complementary spectrum of relaxation modes.
This construction is closely related to the standard
Mori--Zwanzig mechanism by which unresolved dynamical degrees of
freedom generate a retarded memory kernel after projection.

The CFN introduces a distinct additional operation.
The endogenous memory term in
Eq.~(\ref{eq:app_effective_field}) describes how the cognitive field
is dynamically dressed by its internal collective modes.
It acts within the continuous field dynamics and is already contained
in the effective propagator $L_{\rm cog}$.
It therefore explains how previous internal dynamics remains causally
present in the current cognitive field.

Cross-cycle field re-entry instead concerns what the system does with
the field after such a state has been organized.
As defined in the main text, the CFN makes the recurrent field
$\Phi_n$ available as an explicit dynamical condition for the next
inference cycle through
\begin{equation}
\Phi_{n+1}
=
F_\theta
\left(
X_{n+1},
\Phi_n
\right).
\label{eq:app_reentry}
\end{equation}
The continuous field $\phi(t)$ and the computational recurrent field
$\Phi_n$ are not assumed to be microscopically identical.
The CFN equation instead implements at the computational level the
causal principle that an internally organized state can participate
directly in the generation of its successor.

This distinction is essential.
Mori--Zwanzig-type elimination explains why a reduced collective
description becomes history dependent after internal degrees of
freedom are eliminated.
Cognitive-field memory dressing specifies this history dependence in
terms of the collective relaxation spectrum and its self-energy.
Neither operation by itself requires that the resulting macroscopic
field be explicitly supplied to a subsequent inference cycle.

Field re-entry adds precisely this causal pathway.
A memory-dressed state generated by preceding dynamics is retained as
an active computational variable and is allowed to interact with newly
arriving information.
Consequently, new input is processed not only through the current
microscopic transformation but in the presence of an internally
generated field carrying the dynamical consequences of preceding
inference.

The theoretical organization of the CFN may therefore be understood
as three related but distinct levels.
At the first level, collective relaxation modes generate endogenous
memory dressing of the macroscopic field.
At the second, finite cognitive input selectively excites the same
collective manifold and reorganizes the memory-bearing field.
At the third, cross-cycle field re-entry makes the resulting field
causally available to subsequent inference.

The first two levels follow from the driven collective dynamics
derived above.
The third is the additional architectural operation implemented by
the CFN.
It is this separation between the formation of a memory-dressed
cognitive field and the subsequent causal reuse of that field that
distinguishes field re-entry from the memory kernel generated by
projection.

\renewcommand{\thefigure}{C\arabic{figure}}
\renewcommand{\theequation}{C\arabic{equation}}

\setcounter{figure}{0}
\setcounter{equation}{0}

\vspace{1.0cm}

\section{Appendix C: Low-Frequency Driven Cognitive-Field Dynamics,
Passive Relaxation, and Periodic Renewal}
\label{app:driven_field_renewal}

This Appendix derives the low-frequency relations used in Sec.~IV.B
to interpret passive retention, continuing content-dependent drive,
and periodic source re-exposure in the Cognitive Field Network (CFN).

The starting point is the memory-dressed driven cognitive-field
equation derived in Sec.~II.C and Appendix~A,
\begin{equation}
\partial_t\phi(t)
=
-r\phi(t)
+
\int_{t_0}^{t}dt'\,
K(t-t')\phi(t')
+
I_{\rm eff}(t)
+
\xi_{\rm eff}(t).
\label{eq:appB_full_driven_field}
\end{equation}
Here $K(t)$ is the memory kernel generated by the collective
relaxation sector, $I_{\rm eff}(t)$ is the effective cognitive drive
generated after finite external input has coupled to the collective
mode manifold, and $\xi_{\rm eff}(t)$ denotes the corresponding
effective fluctuation term.

The purpose of the present Appendix is not to derive the memory kernel
again.
Rather, we examine several dynamical limits of
Eq.~(\ref{eq:appB_full_driven_field}) that are directly relevant to
the CFN experiments.
In particular, we distinguish passive relaxation from externally
supported persistence and show how repeated content-dependent
re-exposure can generate a stable periodically driven state even when
the underlying passive forgetting scale remains finite.

\vspace{6pt}
\paragraph{I. Low-frequency reduction of the driven cognitive field.}

For the deterministic relations considered below, we suppress the
effective noise and write
\begin{equation}
\partial_t\phi(t)
=
-r\phi(t)
+
\int_{t_0}^{t}dt'\,
K(t-t')\phi(t')
+
I_{\rm eff}(t).
\label{eq:appB_deterministic_field}
\end{equation}

In frequency space,
\begin{equation}
\left[
-i\Omega+r-\Sigma_R(\Omega)
\right]
\phi(\Omega)
=
I_{\rm eff}(\Omega),
\label{eq:appB_frequency_field}
\end{equation}
where
\begin{equation}
\Sigma_R(\Omega)
=
\int_0^\infty dt\,
e^{i\Omega t}K(t)
\label{eq:appB_self_energy}
\end{equation}
is the retarded memory self-energy.

As shown in Appendix~A, the effective cognitive drive is itself
generated by the projection of finite external input onto the
collective dynamical manifold,
\begin{equation}
I_{\rm eff}(\Omega)
=
\mathcal T_I(\Omega)I(\Omega),
\label{eq:appB_effective_drive}
\end{equation}
with
\begin{equation}
\mathcal T_I(\Omega)
=
B_\phi
+
\sum_\alpha
\frac{d_\alpha q_\alpha}
{\mu_\alpha-i\Omega},
\qquad
q_\alpha
=
\widetilde u_\alpha^\dagger B_{\mathcal X}.
\label{eq:appB_input_transfer}
\end{equation}
Consequently,
\begin{equation}
\phi(\Omega)
=
\frac{
\mathcal T_I(\Omega)I(\Omega)
}{
-i\Omega+r-\Sigma_R(\Omega)
}.
\label{eq:appB_full_input_field}
\end{equation}

Equation~(\ref{eq:appB_full_input_field}) separates the coupling of
new information into the collective manifold from the
memory-dressed propagation of the resulting perturbation.

To obtain a local low-frequency description, we approximate
\begin{equation}
\Sigma_R(\Omega)
\simeq
\Sigma_R(0)
\label{eq:appB_static_self_energy}
\end{equation}
over the frequency range relevant to the effective slow dynamics and
define the cognitive forgetting gap
\begin{equation}
r_{\rm cog}
\equiv
r-\Sigma_R(0).
\label{eq:appB_forgetting_gap}
\end{equation}
The effective field equation then reduces to
\begin{equation}
\partial_t\phi(t)
\simeq
-r_{\rm cog}\phi(t)
+
I_{\rm eff}^{\rm IR}(t),
\label{eq:appB_lowfreq_field}
\end{equation}
where $I_{\rm eff}^{\rm IR}$ denotes the low-frequency component of
the effective cognitive drive.

Equation~(\ref{eq:appB_lowfreq_field}) is therefore not an
independently postulated phenomenological equation.
It is the local low-frequency limit of the full non-Markovian
cognitive-field dynamics.

This reduction should be interpreted with care.
If the infrared self-energy retains strong nonanalytic frequency
dependence, the exact long-time evolution need not be described by a
single exponential decay.
The results below therefore characterize the effective local
low-frequency dynamics rather than the most general asymptotic limit
of Cognitive Field Theory.

\vspace{6pt}
\paragraph{II. Passive relaxation and driven persistence.}

We first consider the passive regime in which no content-relevant
effective drive acts on the cognitive field,
\begin{equation}
I_{\rm eff}^{\rm IR}(t)=0.
\end{equation}
Equation~(\ref{eq:appB_lowfreq_field}) then gives
\begin{equation}
\partial_t\phi(t)
=
-r_{\rm cog}\phi(t),
\end{equation}
with solution
\begin{equation}
\phi(t)
=
\phi(t_0)
e^{-r_{\rm cog}(t-t_0)}.
\label{eq:appB_passive_solution}
\end{equation}

Within this local approximation, the characteristic passive
persistence scale is
\begin{equation}
\tau_{\rm cog}
=
\frac{1}{r_{\rm cog}}.
\label{eq:appB_passive_timescale}
\end{equation}
Thus $r_{\rm cog}>0$ implies finite passive relaxation, while
$r_{\rm cog}\rightarrow0^+$ produces increasingly long persistence.

This passive behavior should be distinguished from the dynamics under
a continuing relevant drive.
For an approximately constant low-frequency drive,
\begin{equation}
I_{\rm eff}^{\rm IR}(t)=I_*,
\end{equation}
the field obeys
\begin{equation}
\partial_t\phi(t)
=
-r_{\rm cog}\phi(t)
+
I_*,
\end{equation}
whose solution is
\begin{equation}
\phi(t)
=
\frac{I_*}{r_{\rm cog}}
+
\left(
\phi_0-\frac{I_*}{r_{\rm cog}}
\right)
e^{-r_{\rm cog}(t-t_0)}.
\label{eq:appB_constant_solution}
\end{equation}

For $r_{\rm cog}>0$, the asymptotic driven state is therefore
\begin{equation}
\phi_*
=
\frac{I_*}{r_{\rm cog}}.
\label{eq:appB_driven_fixedpoint}
\end{equation}

A positive forgetting gap consequently implies finite passive
persistence, but it does not imply that the cognitive field must
vanish when content-relevant input continues to act on the collective
dynamics.
Passive forgetting and driven persistence are therefore distinct
dynamical regimes of the same low-frequency field equation.

\vspace{6pt}
\paragraph{III. Periodic source re-exposure and the stroboscopic map.}

The CFN experiments considered in Sec.~IV.B involve repeated
presentation of content-relevant information at a recurrence interval
$T_r$.
Let
\begin{equation}
t_n=nT_r
\end{equation}
denote the successive re-exposure times, and let
$m_n^+$ denote the relevant recurrent component sampled immediately
after the $n$th presentation.

In the simplest local approximation, the field relaxes between
presentations according to
Eq.~(\ref{eq:appB_passive_solution}), giving
\begin{equation}
m_{n+1}^-
=
e^{-r_{\rm cog}T_r}m_n^+.
\label{eq:appB_interevent_decay}
\end{equation}

If the next content-matched presentation generates an effective
renewal contribution $\Delta_r$, then
\begin{equation}
m_{n+1}^+
=
m_{n+1}^-
+
\Delta_r,
\end{equation}
and therefore
\begin{equation}
m_{n+1}^+
=
e^{-r_{\rm cog}T_r}m_n^+
+
\Delta_r.
\label{eq:appB_local_stroboscopic}
\end{equation}

This motivates the more general effective stroboscopic description
\begin{equation}
m_{n+1}^+
=
A_r m_n^+
+
\Delta_r,
\qquad
|A_r|<1,
\label{eq:appB_general_stroboscopic}
\end{equation}
where $A_r$ represents the net propagation and relaxation of the
content-specific recurrent component over one re-exposure interval.

Only in the simplest local Markovian limit should one identify
\begin{equation}
A_r
=
e^{-r_{\rm cog}T_r}.
\label{eq:appB_Ar_local}
\end{equation}
The full CFN may contain multiple relaxation scales, non-Markovian
memory, nonlinear recurrent transformations, and input-dependent
state reorganization.
Accordingly, $A_r$ should generally be regarded as an effective
one-cycle propagation factor rather than as a direct microscopic
measurement of $r_{\rm cog}$.

Iteration of Eq.~(\ref{eq:appB_general_stroboscopic}) gives
\begin{equation}
m_n^+
=
A_r^n m_0^+
+
\frac{\Delta_r(1-A_r^n)}
{1-A_r},
\label{eq:appB_stroboscopic_solution}
\end{equation}
and for $|A_r|<1$ the asymptotic stroboscopic state is
\begin{equation}
m_*^+
=
\frac{\Delta_r}{1-A_r}.
\label{eq:appB_stroboscopic_fixedpoint}
\end{equation}

In the local Markovian limit,
\begin{equation}
m_*^+
=
\frac{\Delta_r}
{1-e^{-r_{\rm cog}T_r}}.
\label{eq:appB_fixedpoint_local}
\end{equation}
This relation provides a useful conceptual connection between passive
forgetting and periodic renewal, but it should not be interpreted as
a quantitative prediction for the CFN unless both $r_{\rm cog}$ and
the effective inter-event propagation are independently measured.

\vspace{6pt}
\paragraph{IV. Periodic steady state and transient approach.}

The fixed point in
Eq.~(\ref{eq:appB_stroboscopic_fixedpoint}) does not imply that the
field becomes constant throughout each re-exposure interval.
It means only that the state sampled at the same phase of successive
cycles becomes stationary.

In the local passive-interevent approximation, after the
stroboscopic fixed point has been reached,
\begin{equation}
\phi(t_n^+)=m_*^+,
\end{equation}
and for $t_n<t<t_{n+1}$,
\begin{equation}
\phi_*(t)
=
m_*^+
e^{-r_{\rm cog}(t-t_n)}.
\label{eq:appB_periodic_interevent}
\end{equation}
Immediately before the next re-exposure,
\begin{equation}
\phi(t_{n+1}^-)
=
A_r m_*^+
\label{eq:appB_periodic_trough}
\end{equation}
in the local approximation, after which the next content-relevant
presentation restores the post-renewal state according to
Eq.~(\ref{eq:appB_general_stroboscopic}).

The resulting asymptotic trajectory is therefore periodic,
\begin{equation}
\phi_*(t+T_r)
=
\phi_*(t),
\label{eq:appB_periodic_solution}
\end{equation}
within the idealized periodic model.
The approximately stationary sawtooth pattern observed under periodic
source re-exposure is the corresponding behavioral signature of
repeated finite relaxation followed by content-dependent renewal.

The same stroboscopic map also describes the transient approach to
this periodically driven state.
Subtracting the fixed point from
Eq.~(\ref{eq:appB_general_stroboscopic}) gives
\begin{equation}
m_n^+-m_*^+
=
A_r^n
\left(
m_0^+-m_*^+
\right).
\label{eq:appB_transient}
\end{equation}
Thus, for $0<A_r<1$, an initially strong recurrent state with
$m_0^+>m_*^+$ exhibits a decreasing envelope that converges to the
nonzero fixed point $m_*^+$ rather than necessarily decaying toward
zero.

This distinction is important for interpreting extended CFN
re-exposure experiments.
A decreasing early envelope and a nonzero late-time periodically
driven regime are not mutually exclusive interpretations.
They can represent successive stages of the same stable driven
dynamics.

The discrete map may also be obtained from a continuous periodic
drive.
Consider
\begin{equation}
\partial_t\phi(t)
=
-r_{\rm cog}\phi(t)
+
I_{\rm per}(t),
\qquad
I_{\rm per}(t+T_r)
=
I_{\rm per}(t).
\label{eq:appB_periodic_drive}
\end{equation}
Evolution over one recurrence interval gives
\begin{align}
\phi(t_{n+1})
={}&
e^{-r_{\rm cog}T_r}\phi(t_n)
\nonumber\\
&+
\int_{t_n}^{t_{n+1}}dt'\,
e^{-r_{\rm cog}(t_{n+1}-t')}
I_{\rm per}(t').
\label{eq:appB_onecycle_solution}
\end{align}
Thus the affine stroboscopic map is the natural one-cycle reduction
of the locally driven field equation and does not require
re-exposure to be represented as an instantaneous impulse.

\vspace{6pt}
\paragraph{V. Representation-dependent renewal and field re-entry.}

The effective renewal amplitude $\Delta_r$ is not expected to be
independent of the incoming representation.
As derived in Appendix~A, finite cognitive input excites the
complementary collective sector through
\begin{equation}
b_\alpha(t)
=
\widetilde u_\alpha^\dagger
B_{\mathcal X}I(t),
\label{eq:appB_modal_input}
\end{equation}
and the resulting mode-mediated field drive is
\begin{equation}
I_{\rm eff}(t)
=
B_\phi I(t)
+
\sum_\alpha d_\alpha
\int_{t_0}^{t}dt'\,
e^{-\mu_\alpha(t-t')}
b_\alpha(t').
\label{eq:appB_mode_filtered_input}
\end{equation}

For a re-exposed source $I_{\rm re}(t)$,
\begin{equation}
b_\alpha^{\rm re}(t)
=
\widetilde u_\alpha^\dagger
B_{\mathcal X}I_{\rm re}(t),
\label{eq:appB_reexposure_modal_input}
\end{equation}
so that different incoming representations may produce different
renewal amplitudes even when their overall magnitudes are comparable.

At the coarse-grained CFN level, this dependence may be written as
\begin{equation}
\Delta_r
=
\mathcal R
\left(
X_r,\Phi_r
\right),
\label{eq:appB_representation_renewal}
\end{equation}
where $X_r$ denotes the re-exposed input representation and $\Phi_r$
the recurrent field on which that input acts.

This expression emphasizes that renewal is not simply the addition of
a fixed external quantity.
The newly presented information acts on an already existing recurrent
state, and the resulting state change depends jointly on the incoming
representation and on the field generated by preceding inference.

This is where the recurrent architecture of the CFN becomes essential.
The low-frequency memory-dressed equation determines how an internal
field persists and relaxes.
Cross-cycle field re-entry makes that surviving field available to the
next computational cycle.
A re-exposed source therefore does not act on a blank state; it acts
on a partially surviving, history-dependent recurrent field.

Consequently, exact source re-exposure and near-paraphrased
re-exposure need not generate the same effective $\Delta_r$.
Their projections onto the learned collective dynamical manifold may
differ, and the recurrent field with which they interact may also
differ.

The present behavioral measurements do not directly determine the
microscopic overlaps
$\widetilde u_\alpha^\dagger B_{\mathcal X}I_{\rm re}$.
Representation-sensitive renewal should therefore be interpreted as
being consistent with mode-selective collective excitation rather
than as a direct measurement of those microscopic quantities.

\vspace{6pt}
\paragraph{VI. Passive forgetting, driven renewal, and persistent recurrent dynamics.}

The results above show that passive forgetting and persistent driven
cognition are different limits of the same memory-dressed dynamical
framework.

In the absence of content-relevant drive,
\begin{equation}
I_{\rm eff}^{\rm IR}=0,
\end{equation}
the field relaxes with the effective scale $r_{\rm cog}$.

Under continuing relevant input,
\begin{equation}
I_{\rm eff}^{\rm IR}=I_*,
\end{equation}
the same field approaches the nonzero driven state
$\phi_*=I_*/r_{\rm cog}$ within the local approximation.

Under periodic content-dependent re-exposure, the recurrent component
is described at the coarse-grained level by
\begin{equation}
m_{n+1}^+
=
A_r m_n^+
+
\Delta_r,
\qquad
m_*^+
=
\frac{\Delta_r}{1-A_r}.
\label{eq:appB_periodic_summary}
\end{equation}

The CFN experiments probe this third regime.
The recurrent state relaxes between content-matched presentations,
while subsequent presentations reorganize the partially surviving
field.
A nonzero late-time periodic regime is therefore compatible with a
finite positive forgetting gap and finite passive retention.

The distinction between passive retention and driven persistence is
especially important in the CFN because field re-entry preserves the
causal availability of the surviving recurrent state.
The field does not need to remain unchanged or infinitely stable in
order to influence later cognition.
It need only survive sufficiently for subsequent input to act on it
and reorganize it through the recurrent dynamics.

Accordingly, the central prediction of the low-frequency description
is not infinite memory.
It is that finite internal persistence, representation-dependent
renewal, and cross-cycle field re-entry can together generate a
stable history-dependent recurrent regime.

Beyond the local approximation, the full relation remains
\begin{equation}
\left[
-i\Omega+r-\Sigma_R(\Omega)
\right]
\phi(\Omega)
=
\mathcal T_I(\Omega)I(\Omega),
\label{eq:appB_full_relation}
\end{equation}
so that passive decay, driven maintenance, and periodic renewal should
be understood as different dynamical limits of the same
memory-dressed, externally driven cognitive field.

\renewcommand{\thefigure}{D\arabic{figure}}
\renewcommand{\theequation}{D\arabic{equation}}

\setcounter{figure}{0}
\setcounter{equation}{0}

\section{Appendix D: Input Construction, Supervised Fine-Tuning,
and Recurrent-State Propagation in Long-Horizon Memory Experiments}
\label{sec:s3_memory_protocol}

This section provides the experimental details underlying the two-binding
memory-retrieval and long-horizon persistence experiments. We describe the
generation of synthetic entity--attribute associations, the construction of
fixed-length WRITE inputs, the supervised fine-tuning procedure used to establish
cross-cycle retrieval, and the subsequent propagation of recurrent hidden fields
through randomized or semantically related intervening inputs. We also specify the
causal intervention controls, evaluation metrics, and checkpoint selection
procedures. These details are important because successful long-horizon retrieval
depends not only on the recurrent architecture but also on how the model is
trained to encode and selectively recover multiple associations from a shared
hidden field.

\vspace{6pt}
\paragraph{I. Synthetic entity--attribute binding dataset.}
\label{sec:s3_dataset}

The memory experiments use synthetically generated entity--attribute binding
episodes rather than passages sampled from a natural-language question-answering
corpus. This construction allows the association between an entity and its
attribute value to be randomized independently for each episode, thereby reducing
the possibility that the correct response can be recovered from general linguistic
knowledge alone. Each episode contains exactly two entity--attribute bindings,
denoted by $N=2$, and the model is subsequently asked to retrieve either
association without receiving the original binding statements again.

Four relation types are included in the dataset:
\texttt{color}, \texttt{location}, \texttt{owner}, and
\texttt{code}. For each episode, one relation type is
selected from these four alternatives. Two distinct synthetic entity names and two
distinct attribute values are then sampled from the corresponding candidate pools.
The resulting entity--value pairs are converted into natural-language statements
using relation-specific templates. The order of the two statements is randomized
before they are combined into a single WRITE context.

The four statement templates and their corresponding question formats are
illustrated below. The entity names and values are representative examples of the
implemented generation rules.

\begin{quote}
\textbf{Color:}

The daxen is purple.

Question: What color is the daxen?

Answer: purple.

\medskip
\textbf{Location:}

The daxen is stored in chamber seven.

Question: Where is the daxen stored?

Answer: chamber seven.

\medskip
\textbf{Owner:}

The daxen belongs to Alice.

Question: Who owns the daxen?

Answer: Alice.

\medskip
\textbf{Code:}

The daxen has code A1.

Question: What code does the daxen have?

Answer: A1.
\end{quote}

The implemented value pools contain 12 colors, 30 storage locations, 24 owner
names, and 24 alphanumeric codes. The color values are
\texttt{red}, \texttt{blue}, \texttt{green},
\texttt{yellow}, \texttt{purple}, \texttt{orange},
\texttt{silver}, \texttt{white}, \texttt{black},
\texttt{brown}, \texttt{pink}, and \texttt{gray}.
The location values are chamber labels, including
\texttt{chamber one}, \texttt{chamber seven},
\texttt{chamber twenty-one}, and
\texttt{chamber forty-seven}. The owner values
are personal names such as \texttt{Alice},
\texttt{Bob}, \texttt{Carol}, and \texttt{David},
whereas the code values are short alphanumeric labels such as \texttt{A1},
\texttt{B2},
\texttt{C3}, and \texttt{D4}.

Synthetic entity names are generated by combining randomly selected initial,
middle, and final syllabic components. A split-specific suffix is appended to the
resulting string. The default initial retrieval experiment uses a pool of 512
synthetic entities for each dataset split, with different random seeds used to
generate the training, validation, and test pools. Consequently, the entity names
used in one split are separated from those used in the other splits by
construction.

For example, a single color episode may contain the two associations
\begin{equation}
\begin{aligned}
e_1 &\longrightarrow \texttt{purple},\\\\
e_2 &\longrightarrow \texttt{green},
\end{aligned}
\end{equation}
represented by the statements
\begin{quote}
The daxen is purple.

The mivor is green.
\end{quote}
The two statements constitute one memory episode, not two independent WRITE
episodes. The assignment of values to entities is randomized for each episode, and
the two values are distinct within the episode. The model must therefore learn to
recover the appropriate entity-specific association rather than return a fixed
answer associated with the relation type.

The initial two-binding fine-tuning configuration generates 12,000 training
episodes, 1,200 validation episodes, and 1,200 test episodes. Because both
bindings are queried independently in the final training protocol, each episode
contributes two single-query retrieval instances. Thus, the 12,000 WRITE episodes
correspond to 24,000 supervised query instances before repeated optimization over
the training set.

The generator constructs episodes by random sampling and does not explicitly
reject duplicate complete episodes. The split-specific entity pools provide
separation of entity identities across training, validation, and testing, while
the relation and value vocabularies remain shared.

\vspace{6pt}
\paragraph{II. Construction of fixed-length WRITE inputs.}
\label{sec:s3_write}

The recurrent CFN was pretrained using input sequences of length $T=1024$ tokens.
To maintain the same token-resolved field size during memory fine-tuning, each
synthetic WRITE episode is converted into a sequence containing exactly 1024
tokens. The actual binding statements are placed at the end of this sequence,
while the preceding positions are filled with neutral background text.

The background text is constructed from the following fixed passage:
\begin{quote}
This is background text for a memory record. The following record contains facts
that may be needed later.
\end{quote}
The passage is tokenized using the Pythia tokenizer and repeated as many times as
necessary to fill the available prefix positions. The memory statements are
introduced by the heading
\texttt{Memory facts:}, followed by the two
generated binding statements. For the illustrative color episode, the resulting
WRITE input has the following structure:
\begin{quote}
\textit{[Repeated neutral background text]}

\texttt{Memory facts:}

The daxen is purple.

The mivor is green.
\end{quote}

Let $L_{\mathrm{facts}}$ denote the number of tokens in the complete
memory-record suffix, including its heading and both binding statements. The
number of background tokens is then
\begin{equation}
L_{\mathrm{background}}
=
1024-L_{\mathrm{facts}}.
\label{eq:s3_background_length}
\end{equation}
The tokenized background passage is repeated and truncated to this length, after
which the complete memory-record suffix is appended. The resulting sequence
therefore satisfies
\begin{equation}
L_{\mathrm{WRITE}}=1024.
\label{eq:s3_write_length}
\end{equation}

This construction preserves the complete binding statements while keeping the
token dimension of the WRITE field fixed. Neutral background tokens are used
instead of padding or end-of-sequence tokens because the recurrent model does not
apply a separate padding-attention mask to the WRITE sequence. The binding
statements remain near the end of the input, so the experimental protocol also
fixes their approximate position within the 1024-token context.

The WRITE input is processed without a previous recurrent field. Its final
token-resolved hidden representation defines the initial memory field,
\begin{equation}
\Phi_0
=
F_\theta(X_{\mathrm{WRITE}},0),
\label{eq:s3_initial_field}
\end{equation}
where $F_\theta$ denotes the recurrent CFN state-update map and the second
argument represents the absence of a preceding field.

In the implemented architecture, the final field retains the full token dimension,
\begin{equation}
\Phi_0\in\mathbb{R}^{1024\times512},
\label{eq:s3_field_dimension}
\end{equation}
for each episode. No mean pooling or compression to a single memory vector is
applied before the field is passed to the next recurrent cycle.

\vspace{6pt}
\paragraph{III. Independent single-query retrieval protocol.}
\label{sec:s3_queries}

The retrieval task is constructed so that the QUERY input does not contain the
binding statements from the preceding WRITE episode. Instead, the query presents
only an instruction asking for the attribute associated with one of the stored
entities. The answer must be recovered from the recurrent hidden field produced by
the earlier WRITE cycle.

The query format is
\begin{quote}
\texttt{Instruction:}

What color is the daxen?

\texttt{Context:}

\texttt{Answer:}
\end{quote}
The \texttt{Context} field is intentionally empty. For the illustrative WRITE
episode, the target response to this query is
\texttt{purple}. A second independent query
asks for the color of \texttt{mivor}, with target response \texttt{green}.

The two queries are not concatenated into a single multi-question prompt. Rather,
they are processed as separate single-query instances that access the same WRITE
field. If the two binding pairs are $(e_1,v_1)$ and $(e_2,v_2)$, the direct
retrieval protocol can be expressed as
\begin{equation}
\Phi_0
=
F_\theta(X_{\mathrm{WRITE}},0),
\end{equation}
followed by
\begin{equation}
\left(
\hat Y_j,\Phi_1^{(j)}
\right)
=
F_\theta
\left(
X_{\mathrm{QUERY}}^{(j)},\Phi_0
\right),
\qquad j=1,2.
\label{eq:s3_two_queries}
\end{equation}
Here $\hat Y_j$ denotes the answer prediction obtained from the output logits of
the corresponding QUERY cycle.

The same source field is shared by both queries, but each query has its own
prediction and supervision target. In implementation, a batch of WRITE episodes is
first encoded into source fields, and these fields are then indexed and expanded
to match the two independent query rows associated with each source.

This design directly tests whether one token-resolved field can support selective
retrieval of more than one stored association. It also avoids introducing an
additional requirement to generate two answers sequentially within the same QUERY
input.

\vspace{6pt}
\paragraph{IV. Initial supervised fine-tuning.}
\label{sec:s3_initial_sft}

The initial cross-cycle retrieval model is obtained by supervised fine-tuning of a
recurrent CFN initialized from a WikiText pretraining checkpoint. The architecture
contains six Transformer blocks with eight-head recurrent cross-attention. Each
block performs causal self-attention, recurrent cross-attention to the preceding
cycle's final hidden field, and a feed-forward transformation. The recurrent
cross-attention is unmasked over the completed previous-cycle field, allowing each
current QUERY token to access all positions of the stored field.

The recurrent attention projections are separate for each Transformer block. A
learnable scalar feedback coefficient $g$ controls the strength of recurrent field
injection. The initial fine-tuning configuration updates the Transformer backbone
and recurrent attention parameters jointly, unless the optional
\texttt{cfn-only} setting is explicitly
enabled. No separate external memory buffer, learned memory-slot compression, or
auxiliary memory-reconstruction loss is introduced.

An important methodological distinction concerns the supervision of the two stored
bindings. In an earlier experimental configuration, each WRITE episode contained
two bindings but only one randomly selected binding was queried during a training
instance. The subsequent two-single-query configuration instead supervises both
bindings independently from the same WRITE field. This change makes the training
objective explicitly sensitive to retrieval of both associations, rather than only
one sampled association per WRITE episode.

The supervised objective is restricted to the target answer tokens, including the
end-of-sequence token. The instruction and QUERY-prefix tokens are excluded from
the loss by assigning them the ignore label used by the cross-entropy
implementation. For two independent queries, the objective is
\begin{equation}
\mathcal L_{\mathrm{SFT}}
=
-\frac{1}{N_{\mathrm{target}}}
\sum_{j=1}^{2}
\sum_{t\in\mathcal T_j}
\log p_\theta
\left(
y_{j,t}
\mid
X_{\mathrm{QUERY}}^{(j)},
\Phi_0,
y_{j,<t}
\right),
\label{eq:s3_sft_loss}
\end{equation}
where $\mathcal T_j$ contains the supervised answer-token positions for query $j$,
and $N_{\mathrm{target}}$ is the total number of supervised tokens across the two
queries.

The implemented loss is computed as the sum of cross-entropies over all unmasked
target positions divided by the total number of such positions. Consequently,
answers containing different numbers of tokens contribute according to their
supervised token counts rather than through an unweighted average of per-query
losses.

The training procedure uses full backpropagation through time (BPTT) across the
WRITE-to-QUERY boundary. The WRITE field is not detached from the computation
graph before being supplied to the QUERY cycle. Thus, gradients originating from
the answer loss can modify both the construction of the source field and its
subsequent retrieval through recurrent attention.

This distinction is important because a model can possess recurrent feedback
connections without having learned to encode and retrieve the particular
information required by the task. The answer-only loss supplies a task-specific
training signal that organizes the WRITE and QUERY transformations jointly.

\vspace{6pt}
\paragraph{V. Optimization and checkpoint selection.}
\label{sec:s3_optimization}

The initial two-binding SFT experiment uses AdamW optimization with a peak
learning rate of $5\times10^{-5}$, a minimum learning rate of $5\times10^{-6}$,
and weight decay of $0.1$. The learning-rate schedule uses a linear warmup over
the initial 3\% of the configured optimizer steps, followed by cosine decay
toward the minimum learning rate. Gradient norms are clipped at $1.0$.

The default microbatch contains four WRITE episodes. Since each episode produces
two independent query rows, the corresponding QUERY batch contains eight rows.
Eight gradient-accumulation steps are performed before each optimizer update,
giving a nominal effective batch of 32 WRITE episodes, or 64 query instances, per
update. The experiment is configured for up to 2000 optimizer steps and a maximum
of ten training epochs.

Validation is performed every 40 optimizer steps, with training statistics logged
more frequently. The implementation records the training answer loss, learning
rate, feedback coefficient $g$, and validation losses and exact-match accuracies
under the causal control conditions described below.

The checkpoint is selected by the lowest CORRECT-field validation loss. The WRONG
and OFF conditions are evaluated as controls but do not contribute additional
terms to the training objective or determine checkpoint selection. The selected
checkpoint is subsequently evaluated on the held-out test split.

The saved training history provides the quantities needed to reconstruct
optimization curves, including training loss and condition-specific validation
performance. However, the present description does not assign numerical
convergence times or learning-curve values without the corresponding recorded
logs.

\vspace{6pt}
\paragraph{VI. Long-horizon recurrent-state propagation.}
\label{sec:s3_recurrence}

After the direct WRITE-to-QUERY retrieval task has been learned, memory
persistence is examined by inserting additional input cycles between WRITE and
QUERY. The defining feature of this protocol is that the original binding
statements are not replayed during the intervening cycles. Only the final hidden
field produced by the preceding cycle is passed forward.

For a retrieval distance $d$, the recurrent field evolves according to
\begin{equation}
\Phi_0
=
F_\theta(X_{\mathrm{WRITE}},0),
\end{equation}
and
\begin{equation}
\Phi_k
=
F_\theta(X_k,\Phi_{k-1}),
\qquad
k=1,\ldots,d-1.
\label{eq:s3_bridge_recurrence}
\end{equation}
The final QUERY is then processed using the field $\Phi_{d-1}$:
\begin{equation}
\left(
\hat Y,\Phi_d
\right)
=
F_\theta
\left(
X_{\mathrm{QUERY}},\Phi_{d-1}
\right).
\label{eq:s3_long_query}
\end{equation}

The distance $d$ therefore counts the number of recurrent transitions from the
WRITE cycle to the QUERY cycle. For $d=1$, the QUERY immediately follows WRITE.
For $d=4$, the sequence is
\begin{equation}
\mathrm{WRITE}
\rightarrow B_1
\rightarrow B_2
\rightarrow B_3
\rightarrow\mathrm{QUERY},
\label{eq:s3_d4_sequence}
\end{equation}
where $B_1$, $B_2$, and $B_3$ are intervening bridge inputs.

Each bridge input is processed by the same recurrent model. Its final hidden field
becomes the previous field for the next cycle. There is no external storage and
retrieval of the original WRITE tokens during this propagation, and model
parameters remain fixed throughout evaluation.

Persistence fine-tuning differs from initial retrieval SFT in the length of the
differentiable recurrent path. The final answer loss is propagated through the
intervening bridge cycles and the initial WRITE cycle using full BPTT. The model
is therefore trained not only to form a retrievable field but also to transform
that field under subsequent inputs without losing the task-relevant associations.

\vspace{6pt}
\paragraph{VII. Randomized neutral bridge construction.}
\label{sec:s3_random_bridges}

The randomized-bridge experiments measure the persistence of stored associations
when intervening inputs contain unrelated material. Bridge sentences are generated
from four predefined groups: opening sentences, middle sentences, transition
sentences, and ending sentences. The groups contain 8, 8, 6, and 8 alternatives,
respectively. A bridge passage is formed by sampling one sentence from each group
and concatenating the selected sentences. The middle and transition sentences may
also exchange positions, introducing additional ordering variation.

A representative passage generated from the implemented sentence groups is
\begin{quote}
The room remained quiet while routine notes were being organized.

Soft daylight entered through a nearby window and the air remained still.

Meanwhile, routine background activity continued at an even pace.

This material is independent of the memory question that will appear later.
\end{quote}

The generated passages are intentionally neutral with respect to the entity--value
associations stored during WRITE. They do not contain the answer values as
task-specific memory cues, and their purpose is to introduce additional
input-driven recurrent transformations between encoding and retrieval.

Each sampled bridge passage is tokenized and repeated or truncated to produce an
input containing exactly 1024 tokens. Unlike the WRITE sequence, in which the
binding statements are appended at the end, the randomized bridge consists of
repeated neutral passage tokens throughout the full sequence.

The $d=4$ persistence experiment uses three independently sampled randomized
bridge inputs per training example. The implemented configuration constructs
separate bridge pools containing 256 training passages, 64 validation passages,
and 64 test passages, using different random seeds. Repeated generation within
each pool is checked to avoid duplicate bridge passages.

The model used for $d=4$ persistence fine-tuning is initialized from a previously
trained two-binding cross-cycle retrieval checkpoint. Training is performed at the
fixed distance $d_{\mathrm{train}}=4$. The microbatch size is reduced to one
WRITE episode because full BPTT through the WRITE and three bridge cycles requires
substantially more memory than direct retrieval training. Gradient accumulation is
increased to 32 steps to retain a nominal effective batch of 32 WRITE episodes.

During evaluation, deterministic bridge selections are used so that the different
recurrent-field conditions can be compared under the same intervening inputs. The
trained model is also evaluated at distances other than four, without additional
parameter updates, to characterize retention beyond the optimization horizon.

\vspace{6pt}
\paragraph{VIII. Semantic continuation bridge construction.}
\label{sec:s3_semantic_bridges}

The semantic-continuation experiments use intervening inputs that remain related
to the entities introduced during WRITE while deliberately omitting their
associated answer values. This condition differs from the randomized neutral
bridges: the intervening text preserves references to the same entities and
therefore provides semantic continuity, but does not explicitly repeat the stored
bindings.

For each of the four relation types, the implementation defines two
relation-specific continuation templates. Each template contains references to
both entities in the episode. The order of the entity names can be reversed,
producing four controlled variants for each relation and entity pair.

For example, a color continuation may take the following form:
\begin{quote}
Later, the daxen and the mivor remained part of the same ongoing record. Both
objects were inspected again while their earlier descriptive details were left
unstated. The discussion continued with the two objects kept distinct and no new
color information was introduced.
\end{quote}

An alternative template is
\begin{quote}
Afterward, attention returned to the daxen and the mivor. They were handled
separately during routine activity, but the earlier appearance of each object was
not repeated. The record continued without stating any color.
\end{quote}

Corresponding templates are defined for location, ownership, and code relations.
The continuation text mentions the entities but avoids restating the attribute
values. Before a semantic bridge is accepted, the implementation checks whether
either target answer appears as a standalone word or phrase in the generated
continuation. A detected answer replay causes the bridge construction to fail.

Each semantic bridge is again converted into a fixed-length 1024-token sequence.
Its prefix is filled by repeating the neutral text
\begin{quote}
The conversation continued naturally. Earlier details remained relevant while the
discussion moved forward. No answer to a later memory question was stated here.
\end{quote}
The entity-related continuation is then appended at the end, preceded by the
heading
\texttt{Ongoing discussion:}.
The number of repeated prefix tokens is chosen so that the complete bridge
sequence contains exactly 1024 tokens.

For the $d=16$ semantic-continuation experiment, fifteen intervening bridge cycles
are constructed. Successive bridge variants alternate between the two templates
and the two entity orderings, so the continuation pattern varies while remaining
within a controlled template family. The same episode's entity identities are
retained throughout these cycles, but the original answer values are not
reintroduced.

The $d=16$ experiment is initialized from an already trained $d=8$
semantic-continuation checkpoint. It therefore represents a further stage of
persistence fine-tuning, rather than training a new model from the original
WikiText checkpoint. The default configuration uses 12,000 training episodes,
3,000 validation episodes, and 1,200 test episodes, with an entity pool size of
2048 and a microbatch size of one WRITE episode. Thirty-two gradient-accumulation
steps are used, and the training distance is fixed at $d_{\mathrm{train}}=16$.

The semantic-continuation protocol tests memory persistence under controlled,
entity-related inputs. It should not be interpreted as a test involving arbitrary
independently sampled natural-language events, because the intervening sentences
are drawn from a restricted set of templates.

\vspace{6pt}
\paragraph{IX. Causal controls and exact-match evaluation.}
\label{sec:s3_controls}

Memory retrieval is evaluated under three recurrent-field conditions, denoted
CORRECT, WRONG, and OFF. The QUERY text and target answer remain unchanged across
these conditions; only the recurrent field supplied to the QUERY cycle is altered.

In the CORRECT condition, the QUERY receives the field generated by its own WRITE
episode and subsequently propagated through the specified bridge inputs. This
condition tests retrieval from the appropriate episode-specific hidden state.

In the WRONG condition, the field is replaced by one generated from a different
WRITE episode. In the direct retrieval implementation, this replacement is
constructed by cyclically shifting source fields across the batch. The two
independent QUERY rows associated with an episode are then mapped to the shifted
source field rather than their own source field.

For bridge-based evaluation, the incorrect source field is propagated through the
same intervening inputs as the correct field. This keeps the current episode's
QUERY and bridge inputs fixed while altering the initial memory content carried by
the recurrent field.

In the OFF condition, the QUERY is processed without a preceding recurrent field.
This removes the cross-cycle information pathway while preserving the current
QUERY input and the model parameters.

The principal evaluation metric is teacher-forced exact-match accuracy. Let
$y_{j,t}$ denote the target token at supervised position $t$ for query $j$. An
individual query is counted as correct only if the model's argmax prediction
matches every supervised target token, including the end-of-sequence token:
\begin{equation}
\mathrm{EM}_j
=
\prod_{t\in\mathcal T_j}
\mathbf{1}
\left[
\arg\max_v
p_\theta(v\mid\cdots)
=
y_{j,t}
\right].
\label{eq:s3_em}
\end{equation}
The reported exact-match accuracy is the average of this indicator over evaluated
single-query instances.

We denote the resulting condition-specific accuracies by
\begin{equation}
P_{\mathrm{EM}}^{C}(d),
\qquad
P_{\mathrm{EM}}^{W}(d),
\qquad
P_{\mathrm{EM}}^{O}(d),
\label{eq:s3_conditions}
\end{equation}
where $C$, $W$, and $O$ correspond to CORRECT, WRONG, and OFF, respectively.

A useful measure of episode-specific retrieval is
\begin{equation}
M(d)
=
P_{\mathrm{EM}}^{C}(d)
-
P_{\mathrm{EM}}^{W}(d).
\label{eq:s3_memory_signal}
\end{equation}
A positive difference indicates that the correct episode-specific field supports
more accurate retrieval than an unrelated field under the same QUERY condition.

The OFF condition provides an additional reference for performance without
recurrent memory. The distinction between CORRECT and WRONG is particularly
important because a model may sometimes generate a plausible answer from its
learned output distribution even without recovering the actual association stored
in the WRITE episode.

Teacher-forced exact match measures the model's token predictions under the
provided target prefix. It is not identical to free-running autoregressive
generation accuracy. The experiments therefore report this metric explicitly
rather than treating it as an unrestricted generation success rate.

\vspace{6pt}
\paragraph{X. Training-distance dependence and reproducibility.}
\label{sec:s3_reproducibility}

The initial retrieval and long-horizon persistence experiments serve different
purposes and use different checkpoint initializations. Direct two-binding
retrieval fine-tuning establishes the WRITE-to-QUERY association at $d=1$ from a
recurrent language-pretraining checkpoint. The randomized-bridge experiment then
extends the recurrent path to $d=4$ from a successful two-binding retrieval
checkpoint. The $d=16$ semantic-continuation experiment is initialized from an
already trained $d=8$ semantic-continuation model.

These stages should not be interpreted as independent training runs from identical
initial conditions. Their optimization histories and initialization checkpoints
are part of the experimental protocol.

The trained distance $d_{\mathrm{train}}$ is also distinguished from the
evaluation distance $d$. During persistence fine-tuning, the model is optimized at
a specified recurrent horizon. During subsequent evaluation, the same learned
parameters are tested over a range of distances without additional gradient
updates.

This distinction makes it possible to examine whether the learned recurrent
dynamics supports retrieval beyond the distance explicitly encountered during
fine-tuning.

The implementations record training and validation statistics, including
answer-token cross-entropy, teacher-forced exact-match accuracy, optimizer step,
learning rate, and the learned feedback coefficient. The direct retrieval
implementation also saves condition-specific validation histories and periodic
checkpoints.

The present supplementary description specifies the data-generation rules,
optimization objectives, and evaluation protocols without reconstructing
unavailable learning curves. Any quantitative statement about convergence speed,
optimization plateaus, or the relative success of earlier training variants
requires the corresponding saved logs or checkpoints.

\vspace{6pt}
\paragraph{XI. Scope and interpretation.}
\label{sec:s3_scope}

The experiments are designed to isolate the ability of a recurrent hidden field to
encode and preserve episode-specific associations. The synthetic binding task
provides an explicit separation between the information introduced during WRITE
and the information available in the later QUERY. The CORRECT, WRONG, and OFF
conditions further distinguish retrieval supported by the appropriate recurrent
state from responses generated without that state.

The randomized-bridge experiments probe persistence under unrelated intervening
inputs, whereas the semantic-continuation experiments probe persistence under
controlled inputs that repeatedly refer to the same entities without revealing
their attribute values. These are complementary conditions, but they do not
establish general-purpose memory under arbitrary unconstrained input streams.

Successful retrieval at distances beyond the trained horizon demonstrates that
task-relevant information remains accessible through the recurrent field over
those tested distances. It does not, by itself, establish indefinite retention,
autonomous memory renewal, or a particular microscopic relaxation-spectrum
mechanism.

The principal methodological conclusion is that the formation and utilization of
recurrent memory are jointly shaped by the architecture, the input-generation
protocol, and the supervised fine-tuning objective. In particular, training both
independent queries from a shared WRITE field provides explicit supervision for
selective retrieval of multiple associations, while full BPTT through intervening
cycles provides a learning signal for their continued accessibility over time.

\end{document}